\documentclass{article}

\usepackage{times}
\usepackage{graphicx} 
\usepackage{subfigure} 

\usepackage{natbib}
\usepackage{times}
\usepackage{epsfig}
\usepackage{graphicx}
\usepackage{amsmath}
\usepackage{amsthm}
\usepackage{amssymb}
\usepackage{mathtools}
\usepackage{color}
\usepackage{multirow}
\usepackage{booktabs}
\usepackage{xcolor}

\usepackage[accepted]{icml2014}

\newcommand\x{\mathbf{x}}
\newcommand\w{\mathbf{w}}
\newcommand\uu{\mathbf{u}}
\newcommand\z{\mathbf{z}}
\newcommand\A{\mathbf{A}}
\newcommand\Z{\mathcal{Z}}
\newcommand\Real{\mathbb{R}}
\DeclareMathOperator{\sign}{sign}
\DeclareMathOperator{\cov}{cov}
\DeclareMathOperator*{\argmin}{arg\,min}

\newcommand{\Bs}{\mathcal{B}_I} 
\newcommand{\Gs}{\mathcal{G}} 
\newcommand{\Es}{\mathcal{E}}
\newcommand{\Vs}{\mathcal{V}}
\newcommand{\Us}{\mathcal{U}}
\newcommand{\Ps}{\mathcal{P}}
\newcommand{\inter}{\cap}
\newcommand{\union}{\cup}

\newtheorem{lemma}{Lemma}

\usepackage{amsfonts,amsmath,amssymb,amsthm}
\usepackage{amsthm}
\usepackage{color}

\icmltitlerunning{On learning to localize objects with minimal supervision}

\begin{document} 

\twocolumn[
\icmltitle{On learning to localize objects with minimal supervision}

\icmlauthor{Hyun Oh Song}{song@eecs.berkeley.edu}
\icmlauthor{Ross Girshick}{rbg@eecs.berkeley.edu}
\icmlauthor{Stefanie Jegelka}{stefje@eecs.berkeley.edu}
\icmlauthor{Julien Mairal}{julien.mairal@inria.fr}
\icmlauthor{Zaid Harchaoui}{zaid.harchaoui@inria.fr}
\icmlauthor{Trevor Darrell}{trevor@eecs.berkeley.edu}

\icmlkeywords{smooth optimization, machine learning, ICML}

\vskip 0.3in
]

\begin{abstract} 
Learning to localize objects with minimal supervision is an important problem in computer vision, since large fully annotated datasets are extremely costly to obtain. In this paper, we propose a new method that achieves this goal with only image-level labels of whether the objects are present or not. Our approach combines a discriminative submodular cover problem for automatically discovering a set of positive object windows with a smoothed latent SVM formulation.
The latter allows us to leverage efficient quasi-Newton optimization techniques. Our experiments demonstrate that the proposed approach provides a 50\% relative improvement in mean average precision over the current state-of-the-art on PASCAL VOC 2007 detection. 
\end{abstract} 

\section{Introduction}

The classical paradigm for learning object detection models starts by annotating each object instance, in all training images, with a bounding box. However, this exhaustive labeling approach is costly and error prone for large-scale datasets.
The massive amount of textually annotated visual data available online inspires a different, more challenging, research problem. Can weakly-labeled imagery, without bounding boxes, be used to reliably train object detectors? 

In this alternative paradigm, the goal is to learn to localize objects with minimal supervision \cite{perona1, perona2}. We focus on the case where the learner has access to binary image labels that encode whether an image contains the target object or not, without access to any instance level annotations (i.e., bounding boxes). 

Our approach starts by reducing the set of possible image locations that contain the object of interest from millions to thousands per image, using the selective search window proposal technique introduced by \citet{selectivesearch}. Then, we formulate a  discriminative submodular cover algorithm to discover an initial set of image windows that are likely to contain the target object. After training a detection model with this initial set, we refine the detector using a novel smoothed formulation of latent SVM \cite{misvm-nips,lsvm-pami}. We employ recently introduced object detection features, based on deep convolutional neural networks \cite{decafICML,girshick2014rcnn}, to represent the window proposals for clustering and detector training.

Compared to prior work on weakly-supervised detector training, we show substantial improvements on the standard evaluation metric (detection average precision on PASCAL VOC). Quantitatively, our approach achieves a 50\% relative improvement in mean average precision over the current state-of-the-art for weakly-supervised learning.

\section{Related work}

Our work is related to three active research areas: (1) weakly-supervised learning, (2) unsupervised discovery of mid-level visual elements, and (3) co-segmentation.

We build on a number of previous approaches for training object detectors from weakly-labeled data.
In nearly all cases, the task is formulated as a multiple instance learning (MIL) problem \cite{mil3}.
In this formulation, the learner has access to an image-level label indicating the presence or absence of the target class, but not its location (if it is present).
The challenge faced by the learner is to find the sliver of signal present in the positive images, but absent from the negative images.
The implicit assumption is that this signal will correspond to the positive class.

Although there have been recent works on convex relaxations \cite{li13, bach12}, most MIL algorithms start from an initialization and then perform some form of local optimization.
Early efforts, such as \cite{perona1,perona2,galleguillos2008weakly,fergus2007weakly,crandall2006weakly,chum2007exemplar, neil13}, focused on datasets with strong object-in-the-center biases (e.g. Caltech-101).
This simplified setting enabled clarity and focus on the MIL formulation, image features, and classifier design, but masked the vexing problem of finding a good initialization in data where such helpful biases are absent.

More recent work, such as \cite{siva1,siva2012defence}, attempts to learn detectors, or simply automatically generate bounding box annotations from much more challenging datasets such as PASCAL VOC \cite{PASCAL-ijcv}.
In this data regime, focusing on initialization is crucial and carefully designed heuristics, such as shrinking bounding boxes \cite{russakovsky}, are often employed.

Recent literature on unsupervised mid-level visual element discovery \cite{discovery1, discovery2, discovery4, discovery5, discovery6} uses weak labels to discover visual elements that occur commonly in positive images but not in negative images. Discovered visual element representation were shown to successfully provide discriminative information in classifying images into scene types. The most recent work \cite{discovery3} presents a discriminative mode seeking formulation and draws connections between discovery and mean-shift algorithms \cite{meanshift1}.



The problem of finding common structure is related to the challenging setting of co-segmentation \cite{rother06, joulin10, alexe10}, which is the unsupervised segmentation of an object that is present in multiple images. While in this paper we do not address pixel-level segmentation, we employ ideas from co-segmentation: the intuition behind our submodular cover framework in Section~\ref{sec:covering} is shared with CoSand \cite{kim11}. Finally, submodular covering ideas have recently been applied to (active) filtering of hypothesis after running a detector, and without the discriminative flavor we propose \cite{barinova12,chen14}.









\section{Problem formulation}
Our goal is to learn a detector for a visual category from a set of images, each with a binary label.
We model an image as a set of overlapping rectangular windows and follow a standard approach to detection: reduce the problem of detection to the problem of binary classification of image windows.
However, at training time we are only given image-level labels, which leads to a classic multiple instance learning (MIL) problem.
We can think of each image as a ``bag'' of instances (rectangular windows) and
the binary image label $y = 1$ specifies that the bag contains at least one instance of the target category.
The label $y = -1$ specifies that the image contains no instances of the category.
During training, no instance labels are available.

MIL problems are typically solved (locally) by finding a local minimum of a non-convex objective function, such as MI-SVM \cite{misvm-nips}.
In practice, the quality of the local solution depends heavily on the quality of the initialization.
We therefore focus extensively on finding a good initialization.
In Section \ref{sec:covering}, we develop an initialization method by formulating a discriminative set multicover problem that can be solved approximately with a greedy algorithm.
This initialization, without further MIL refinement, already produces good object detectors, validating our approach.
However, we can further improve these detectors by optimizing the MIL objective.
We explore two alternative MIL objectives in Section \ref{sec:slsvm}.
The first is the standard Latent SVM (equivalently MI-SVM) objective function, which can be optimized by coordinate descent on an auxiliary objective that upper-bounds the LSVM objective.
The second method is a novel technique that smoothes the Latent SVM objective and can be solved more directly with unconstrained smooth optimization techniques, such as L-BFGS \cite{lbfgs}.
Our experimental results show modest improvements from our smoothed LSVM formulation on a variety of MIL datasets.

\section{Finding objects via submodular cover}
\label{sec:covering}

Learning with LSVM is a chicken and egg problem:
The model weights are needed to infer latent annotations, but the latent annotations are needed to estimate the model weights.
To initialize this process, we approximately identifying jointly present objects in a weakly supervised manner. The experiments show a significant effect from this initialization.
%
Our procedure implements two essential assumptions: (i) the correct boxes are similar, in an appropriate feature space, across positive images (or there are few modes), and (ii) the correct boxes do not occur in the negative images. In short, in the similarity graph of all boxes we seek dense subgraphs that only span the positive images. Finding such subgraphs is a nontrivial combinatorial optimization problem.

The problem of finding and encoding a jointly present signal in images is an old one, and has been addressed by clustering, minimum description length priors, and the concept of exemplar \cite{darrell90,leibe04,schiele06,kim11}.
These approaches share the idea that a small number of exemplars or clusters should well encode the shared information we are interested in. We formalize this intuition as a flexible \emph{submodular cover} problem. However, we also have label information at hand that can help identify correct boxes. We therefore integrate into our covering framework the relevance for positively versus negatively labeled images, generalizing ideas from \cite{discovery1}. This combination allows us to find multiple modes of the object appearance distribution.

Let $\Ps$ be the set of all positive images. Each image contains a set $\Bs = \{b_1, \ldots, b_m\}$ of candidate bounding boxes generated from selective search region proposals \cite{selectivesearch}. In practice, there are about $2000$ region proposal boxes per image and about $5000$ training images in the PASCAL VOC dataset. 
Ultimately, we will define a function $F(S)$ on sets $S$ of boxes that measures how well the set $S$ represents $\Ps$.
For each box $b$, we find its nearest neighbor box in each (positive \emph{and negative}) image. We sort the set $\mathcal{N}(b)$ of all such neighbors of $b$ in increasing order by their distance to $b$. This can be done in parallel.
We will define a graph using these nearest neighbors that allows us to optimize for a small set of boxes $S$ that are (i) \emph{relevant} (occur in many positive images); (ii) \emph{discriminative} (dissimilar to the boxes in the negative images); and (iii) \emph{complementary} (capture multiple modes).

\begin{figure}
  \centering
  \includegraphics[width=0.48\textwidth]{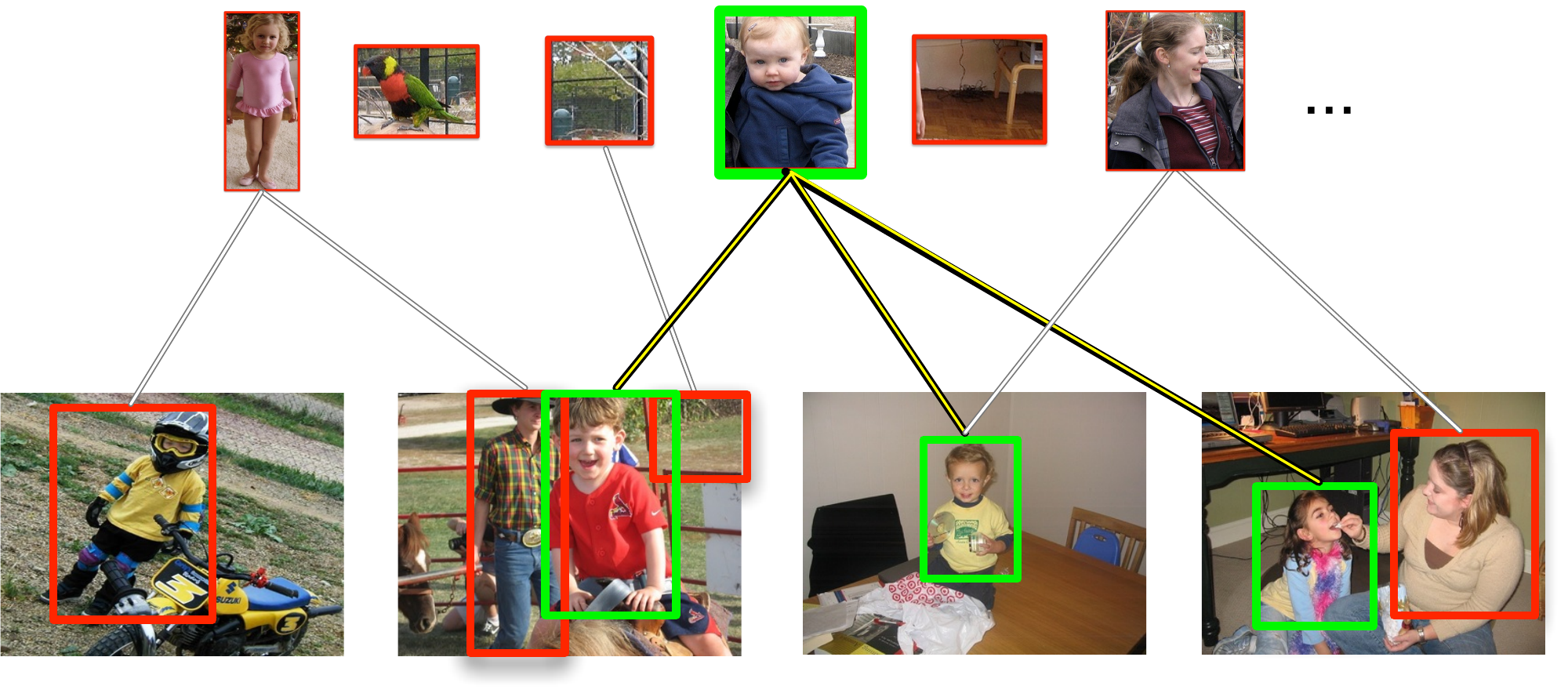} 
  \vspace{-0.8cm}
  \caption{Illustration of the graph $\Gs$ with $\Vs$ (top row) and $\Us$ (bottom row). Each box $b \in \Vs$ is connected to its closest neighbors from positive images (one from each image). Non-discriminative boxes occur in all images equally, and may not even have any boxes from positive images among their closest neighbors -- and consequently no connections to $\Us$. Picking the green-framed box $v$ in $\Vs$ ``covers'' its (green) highlighted neighbors $\Gamma(b)$.}
  \label{fig:graph}
\end{figure}

We construct a bipartite graph $\Gs = (\Vs, \Us, \Es)$ whose nodes $\Vs$ and $\Us$ are all boxes occurring in $\Ps$ (each $b$ occurs once in $\Vs$ and once in $\Us$). The nodes in $\Us$ are partitioned into groups $\Bs$: $\Bs$ contains all boxes from image $I \in \Ps$. The edges $\Es$ are formed by connecting each node (box) $b \in \Vs$ to its top $k$ neighbors in $\mathcal{N}(b) \subseteq \Us$ from positive images. 
Figure~\ref{fig:graph} illustrates the graph.
Connecting only to the top $k$ neighbors (instead of all) implements discriminativeness: the neighbors must compete. If $b$ occurs in positively and negatively labeled images equally, then many top-$k$ closest neighbors in $\mathcal{N}(b)$ stem from negative images. Consequently, $b$ will not be connected to many nodes (boxes from $\Ps$) in $\Gs$. 
%
We denote the neighborhood of a set of nodes $S \subseteq \Vs$ by
  $\Gamma(S) = \{ b \in \Us \mid \exists (v,b) \in \Es \text{ with } v \in S\}$.

Let $S \subseteq \Vs$ denote a set of selected boxes. We define a \emph{covering score} $\cov_{I,t}(S)$ for each $I$ that is determined by a \emph{covering threshold} $t$ and a scalar, nondecreasing concave function $g: \mathbb{R}_+ \to \mathbb{R}_+$:
\begin{equation}
  \label{eq:1}
 \vspace{-4pt}
  \cov_{I,t}(S) = g( \min\{t, \; |\Gamma(S) \inter \Bs|\}).
\end{equation}
This score measures how many boxes in $\Bs$ are neighbors of $S$ and thus ``covered''. We gain from covering up to $t$ boxes from $\Bs$ -- anything beyond that is considered redundant.
The \emph{total covering score} of a set $S \subseteq \Vs$ is then 
\begin{equation}
 \vspace{-4pt}
  \label{eq:totalcover}
  F(S) = \sum\nolimits_{I \in \Ps} \cov_{I,t}(S).
\end{equation}
The threshold $t$ balances relevance and complementarity: let, for simplicity, $g = \mathrm{id}$. If $t=1$, then a set that maximizes $\cov_{I,t}(S)$ contains boxes from many different images, and few from a single image. The selected neighborhoods are very complementary, but some of them may not be very relevant and cover outliers. If $t$ is large, then any additionally covered box yields a gain, and the best boxes $b \in \Vs$ are those with the largest degree. A box has large degree if many of its closest neighbors in $\mathcal{N}(b)$ are from positive images. This also means $b$ is discriminative and relevant for $\Ps$.
\begin{lemma}
  The function $F: 2^\Vs \to \mathbb{R}_+$ defined in Equation~\eqref{eq:totalcover} is nondecreasing and submodular.
\end{lemma}
A set function is \emph{submodular} if it satisfies \emph{diminishing marginal returns}: for all $v$ and $S \subseteq T \subseteq \Vs \setminus \{v\}$, it holds that $F(S \union \{v\}) - F(S) \geq F(T \union \{v\}) - F(T)$.
\vspace{-5pt}
\begin{proof}
  First, the function $S \mapsto |\Gamma(S) \inter \Bs|$ is a covering function and thus submodular: let $S \subset T \subseteq \Vs \setminus b$. Then $\Gamma(S) \subseteq \Gamma(T)$ and therefore
  \begin{align}
    \vspace{-4pt}
    |\Gamma(T \union \{b\})| - |\Gamma(T)| &= |\Gamma(b) \setminus \Gamma(T)|\\
    &\leq |\Gamma(b) \setminus \Gamma(S)|\\
    &= |\Gamma(S \union \{b\})| - |\Gamma(S)|.
  \end{align}
  The same holds when intersecting with $\Bs$.
  Thus, $\cov_{t,I}(S)$ is a nondecreasing concave function of a submodular function and therefore submodular. Finally, $F$ is a sum of submodular functions and hence also submodular. Monotonicity is obvious.
 %
\end{proof}

We aim to select a representative subset $S \subseteq \Vs$ with minimum cardinality:
\begin{align}
 \vspace{-4pt}
  \label{eq:submodcover}
  \min_{S \subseteq \Vs} |S| \quad \text{s.t.} \;\; F(S) \geq \alpha F(\Vs)
\end{align}
for $\alpha \in (0,1]$. We optimize this via a greedy algorithm: let $S_0 = \emptyset$ and, in each step $\tau$, add the node $v$ that maximizes the marginal gain $F(S_{\tau} \union \{v\}) - F(S_{\tau})$.
\begin{lemma}
  \label{lem:bound}
  The greedy algorithm solves Problem~\eqref{eq:submodcover} within an approximation factor of 
  $1 + \log\left(\frac{k g(1)}{g(t)-g(t-1)}\right) = O(\log k)$.
\end{lemma}
Lemma~\ref{lem:bound} says that the algorithm returns a set $\widehat{S}$ with $F(\widehat{S}) \geq \alpha F(\Vs)$ and $|\widehat{S}| \leq O(\log k) |S^*|$, where $S^*$ is an optimal solution.
This result follows from the analysis by \citet{wolsey82} (Thm.~1) adapted to our setting.
To get a better intuition for the formulation~\eqref{eq:submodcover} we list some special cases:\\
\textbf{Min-cost cover.} With $t=1$ and $g(a) = a$ being the identity, Problem~\ref{eq:submodcover} becomes a min-cost cover problem. Such straightforward covering formulations have been used for filtering after running a detector \cite{barinova12}.\\
\textbf{Maximum relevance.} A minimum-cost cover merely focuses on complementarity of the selected nodes $S$, which may include rare outliers. At the other extreme ($t$ large), we would merely select by the number of neighbors (\citet{discovery1} choose one single $\mathcal{N}(b)$ that way).\\
%
%
\textbf{Multi-cover.} To smoothly move between the two extremes, one may choose $t > 1$ and $g$ to be sub-linear. This trades off representation, relevance, and discriminativeness.

\begin{figure*}[htbp]
\centering
\includegraphics[width=0.1\textwidth, height=1cm]{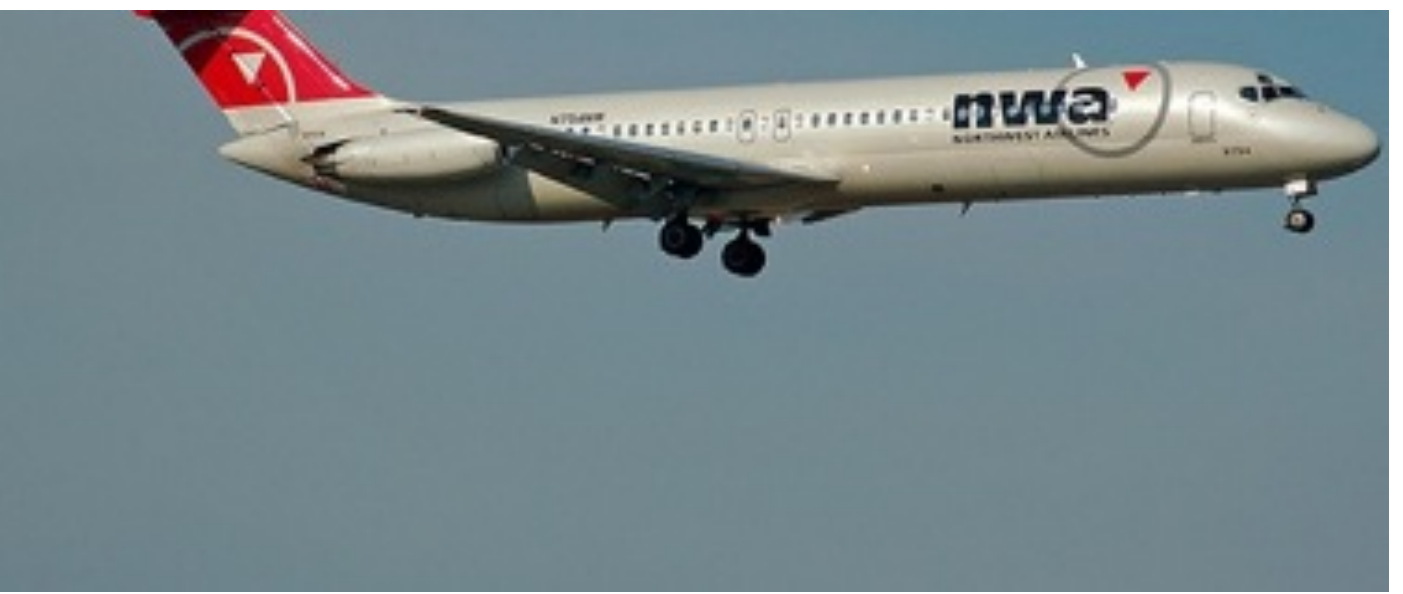}\hspace{-0.12cm}
\includegraphics[width=0.1\textwidth, height=1cm]{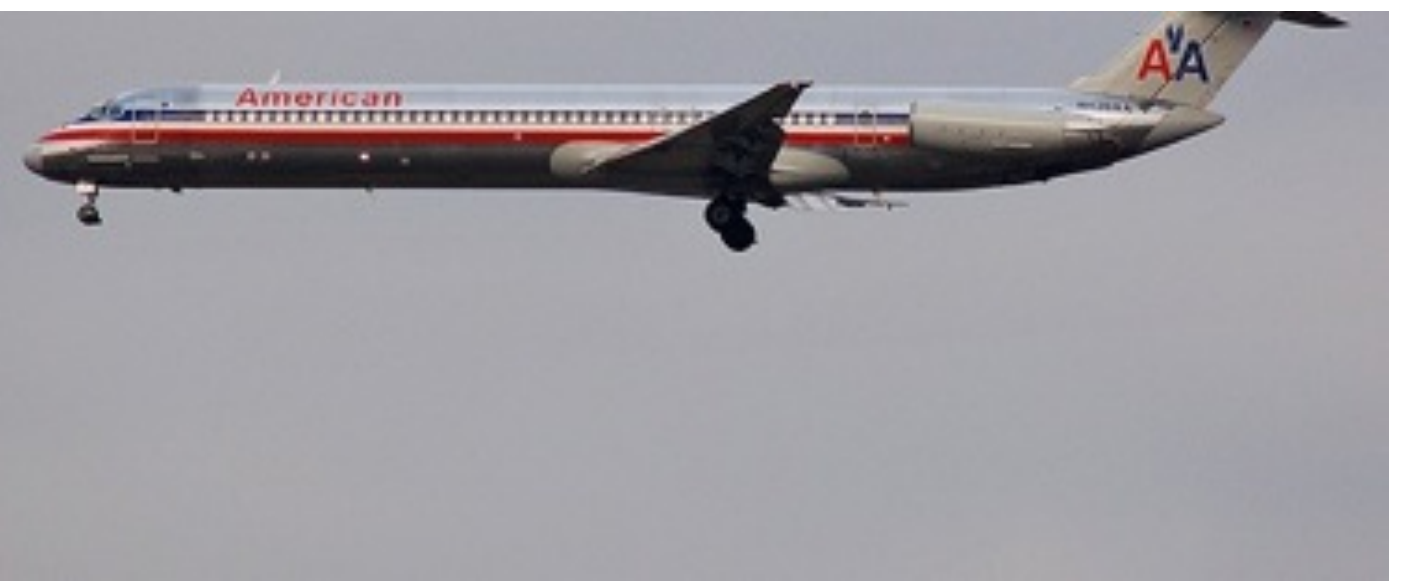}\hspace{-0.12cm}
\includegraphics[width=0.1\textwidth, height=1cm]{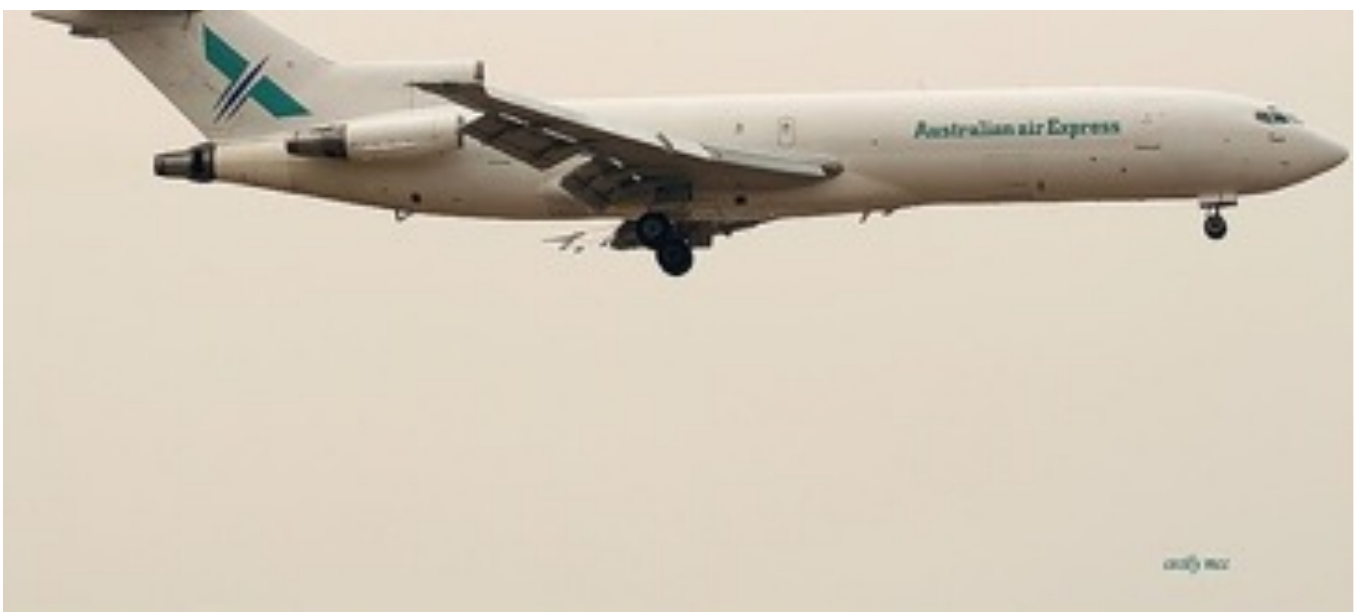}\hspace{-0.12cm}
\includegraphics[width=0.1\textwidth, height=1cm]{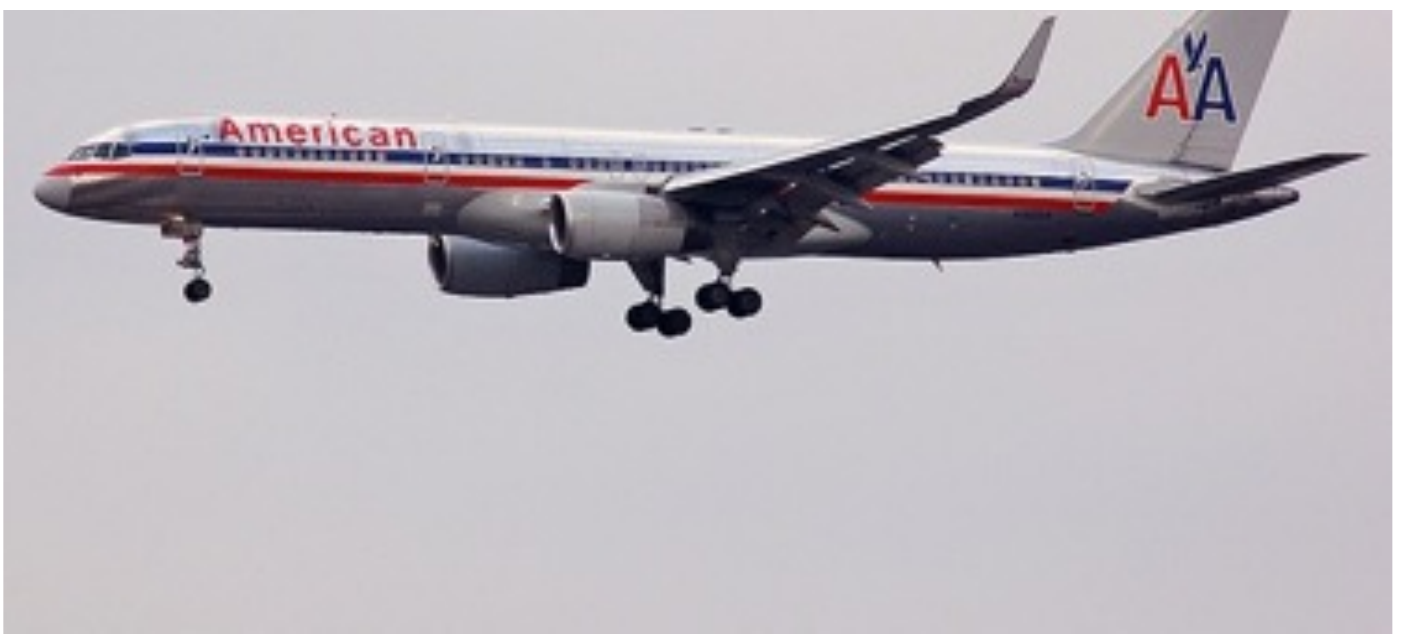}\hspace{-0.12cm}
\includegraphics[width=0.1\textwidth, height=1cm]{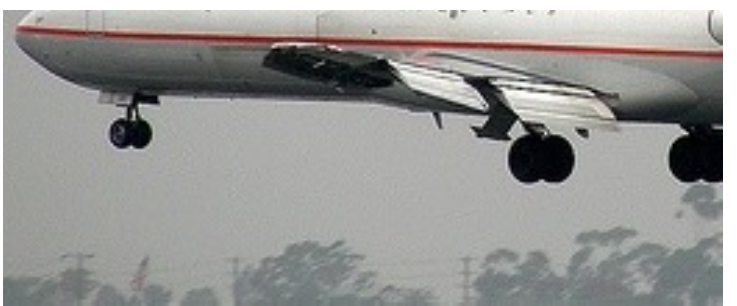}\hspace{0.2cm}
\includegraphics[width=0.1\textwidth, height=1cm]{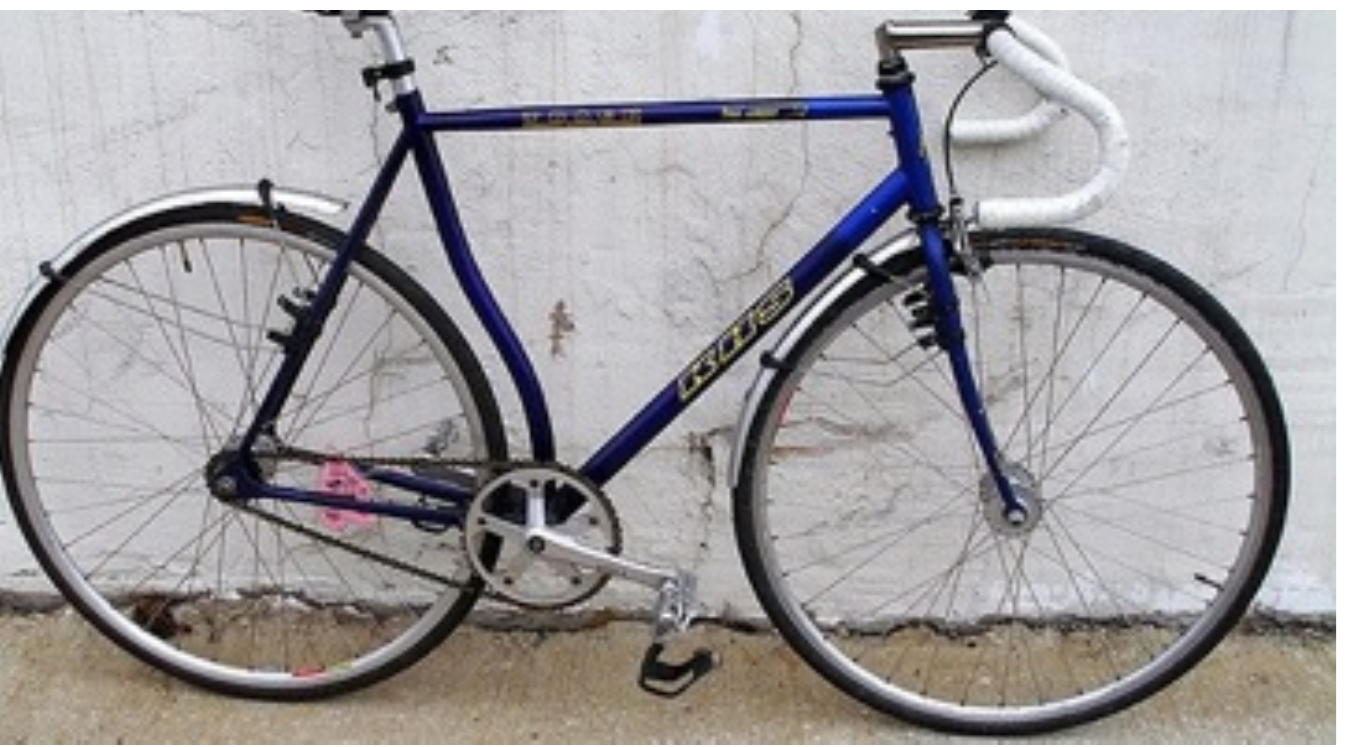}\hspace{-0.12cm}
\includegraphics[width=0.1\textwidth, height=1cm]{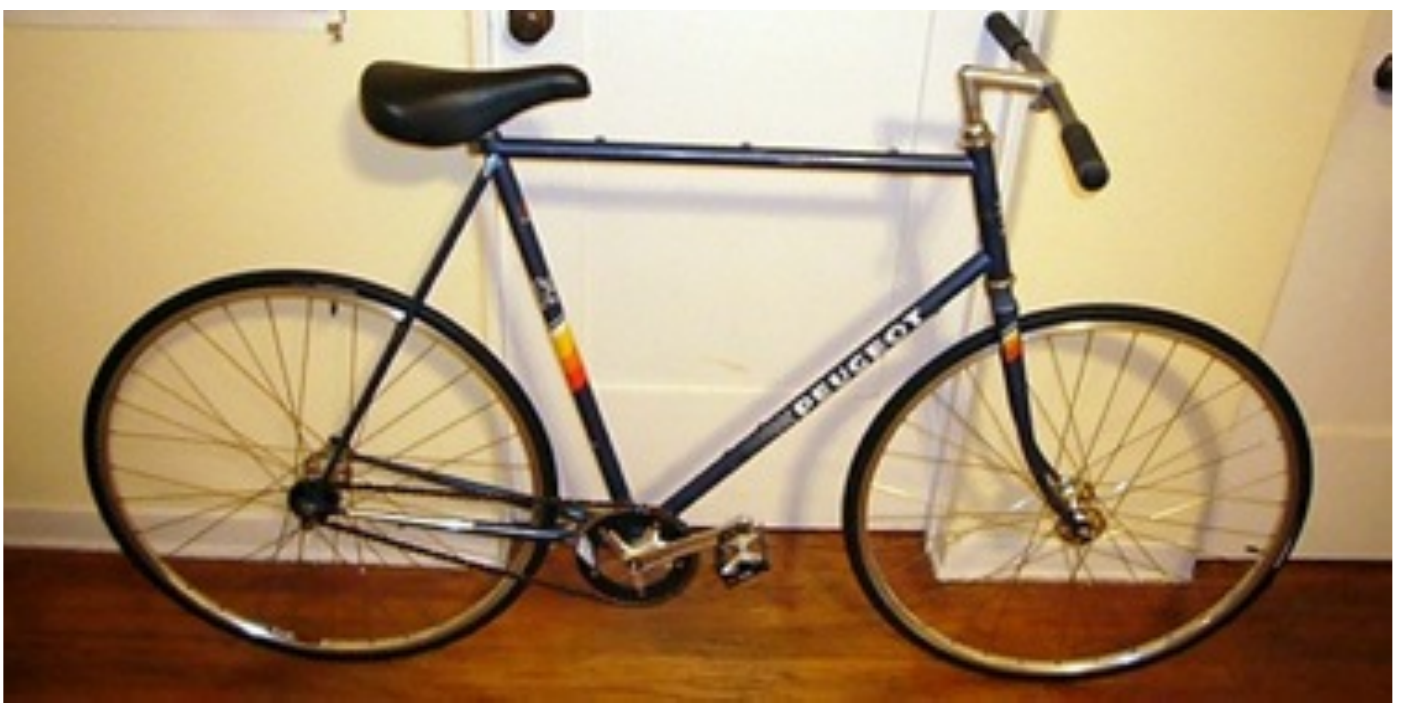}\hspace{-0.12cm}
\includegraphics[width=0.1\textwidth, height=1cm]{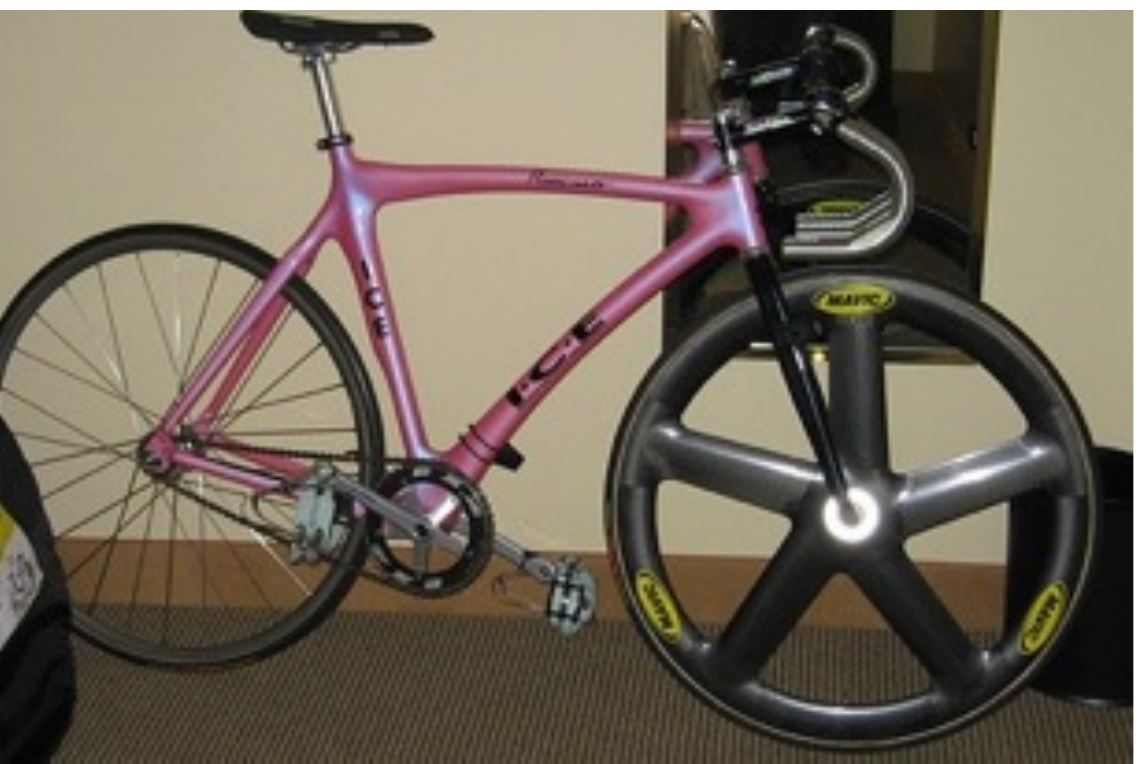}\hspace{-0.12cm}
\includegraphics[width=0.1\textwidth, height=1cm]{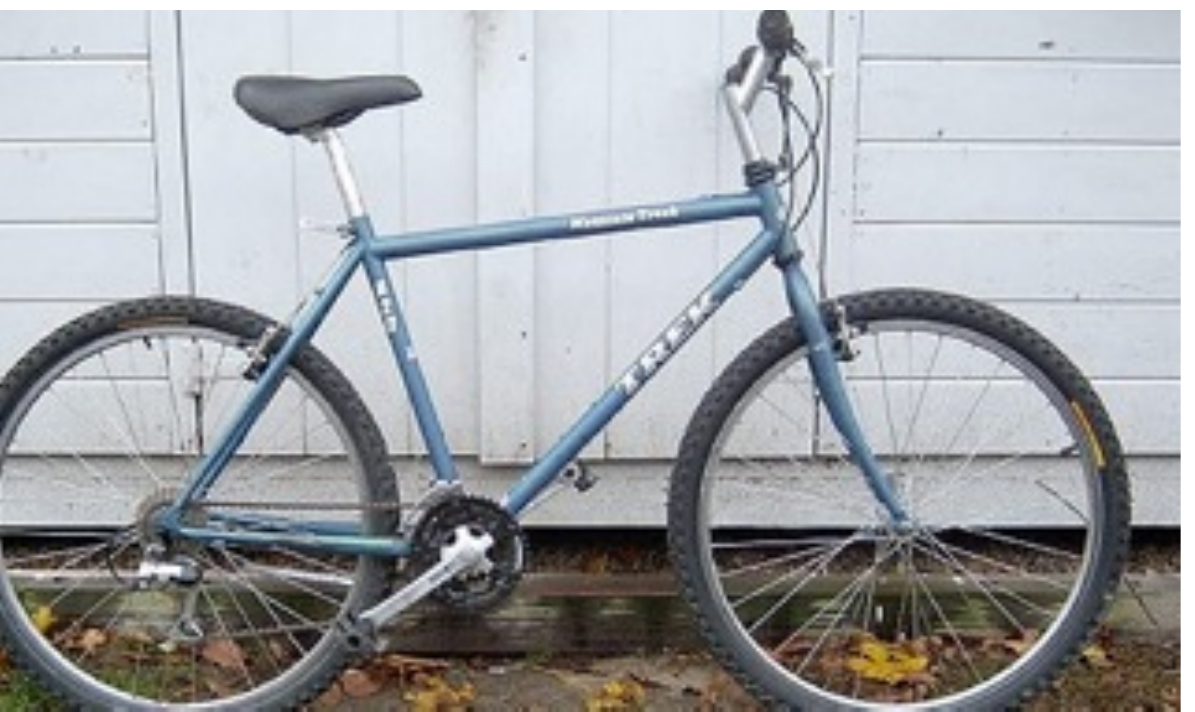}\hspace{-0.12cm}
\includegraphics[width=0.1\textwidth, height=1cm]{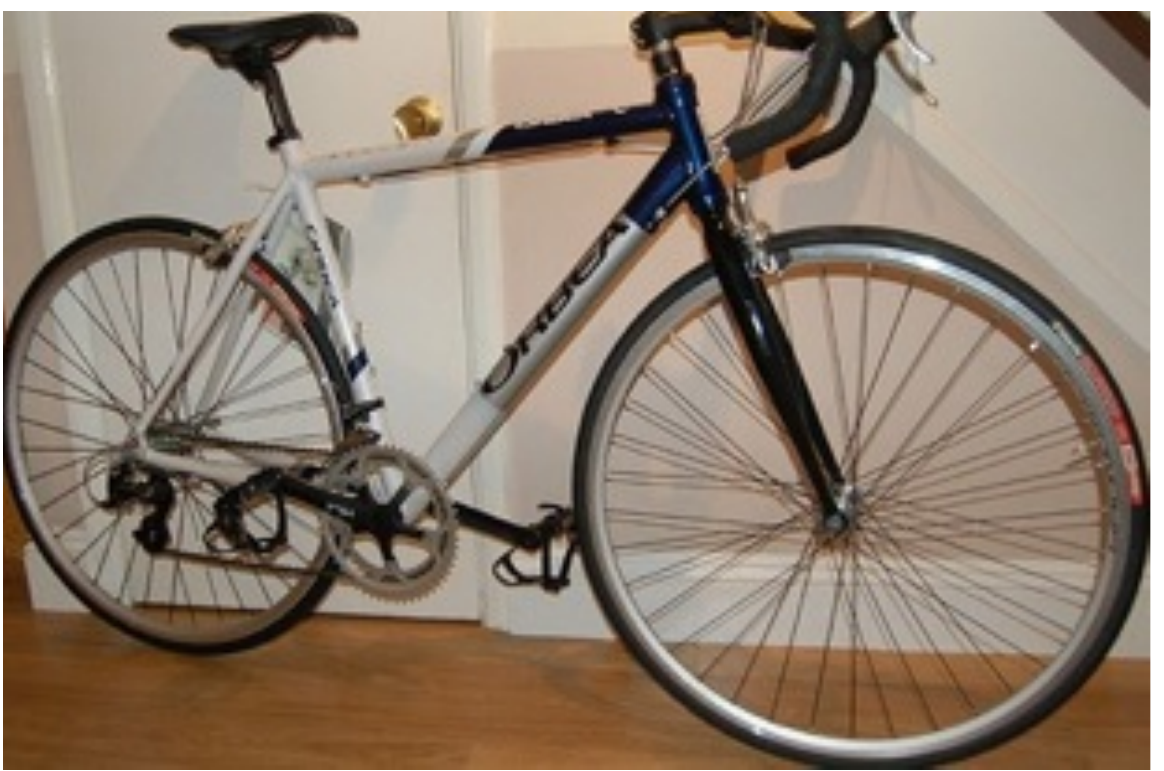}\vspace{0.2cm}\\
\includegraphics[width=0.1\textwidth, height=1cm]{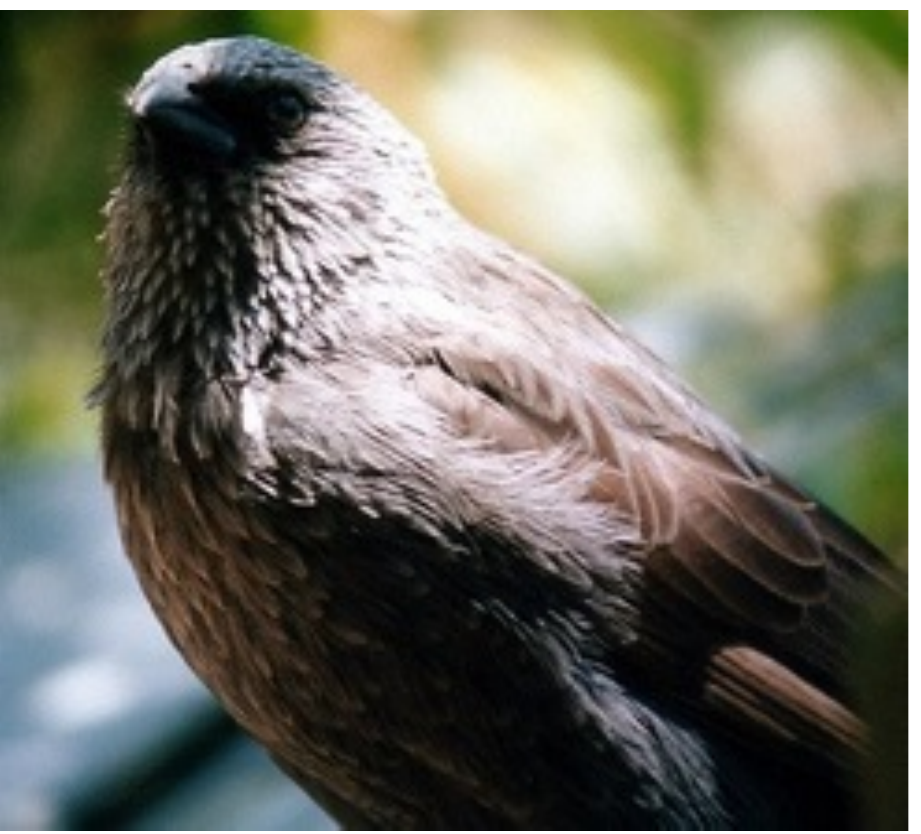}\hspace{-0.12cm}
\includegraphics[width=0.1\textwidth, height=1cm]{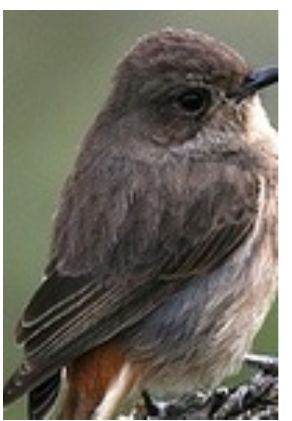}\hspace{-0.12cm}
\includegraphics[width=0.1\textwidth, height=1cm]{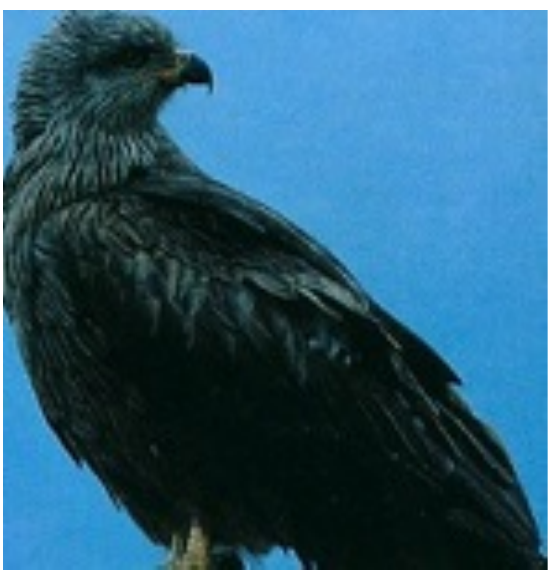}\hspace{-0.12cm}
\includegraphics[width=0.1\textwidth, height=1cm]{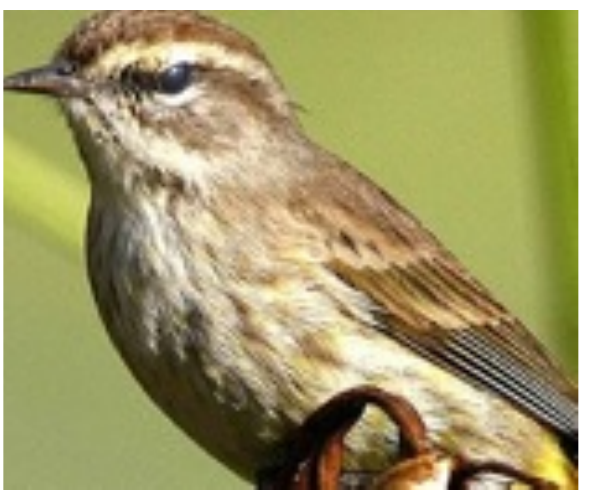}\hspace{-0.12cm}
\includegraphics[width=0.1\textwidth, height=1cm]{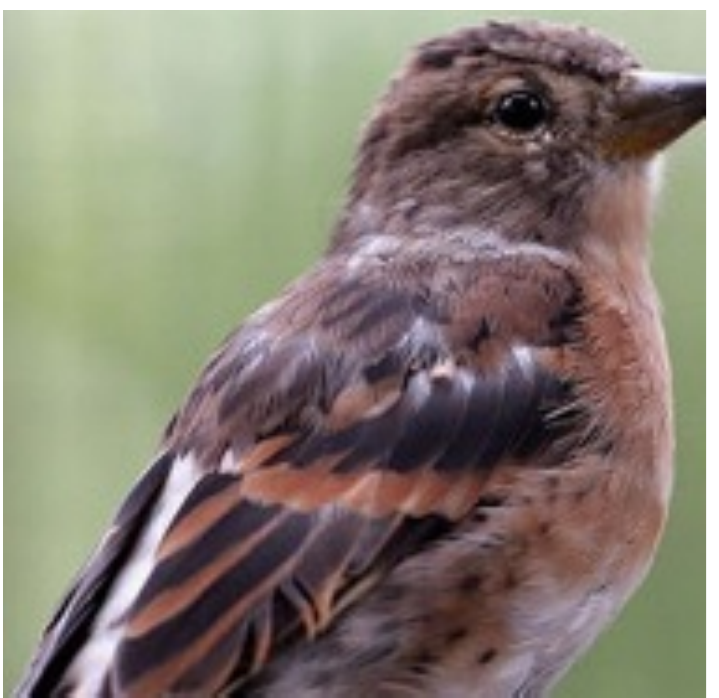}\hspace{0.2cm}
\includegraphics[width=0.1\textwidth, height=1cm]{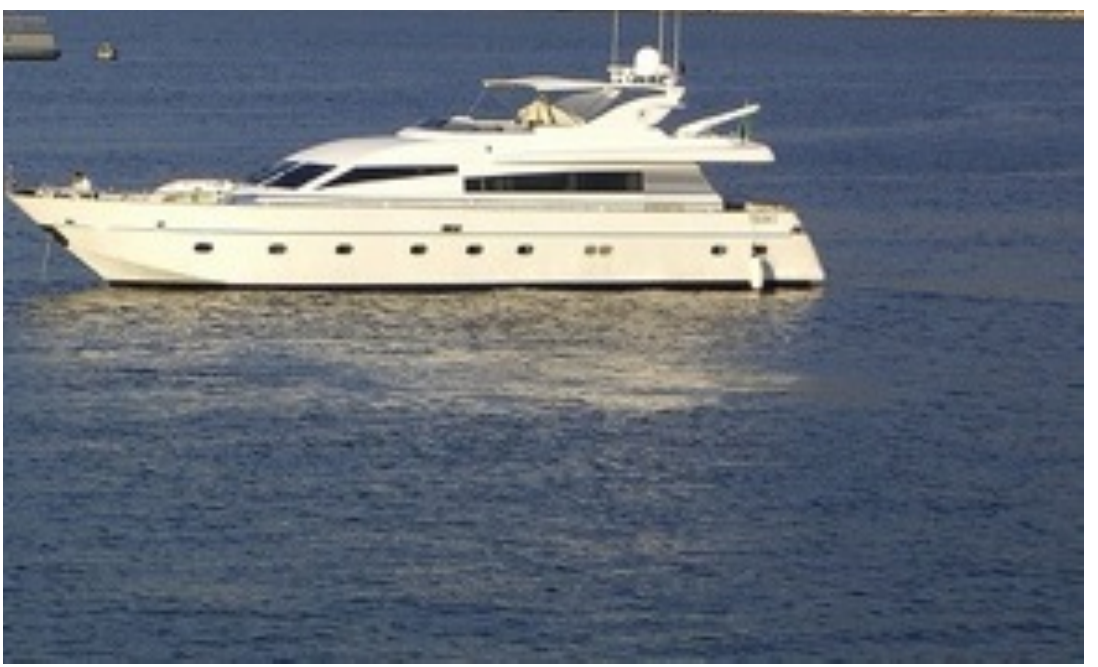}\hspace{-0.12cm}
\includegraphics[width=0.1\textwidth, height=1cm]{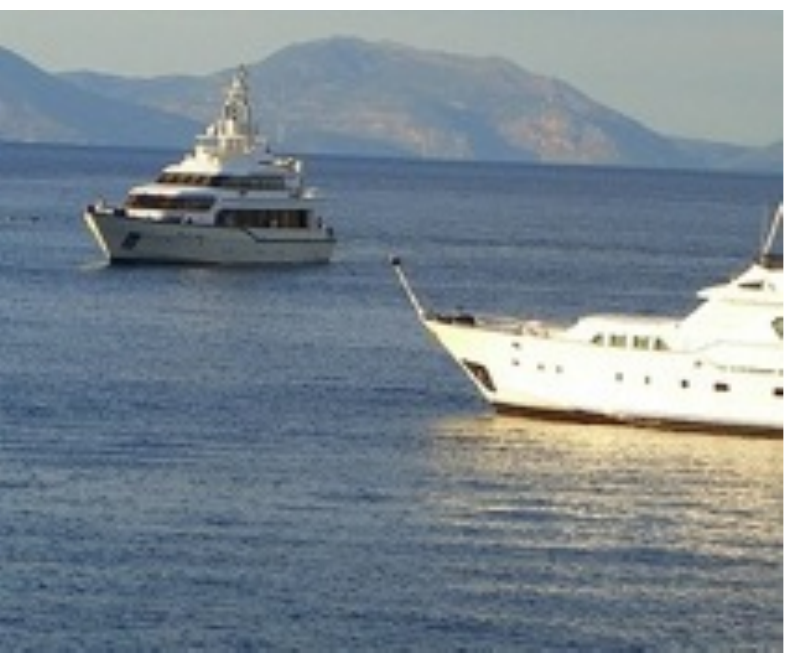}\hspace{-0.12cm}
\includegraphics[width=0.1\textwidth, height=1cm]{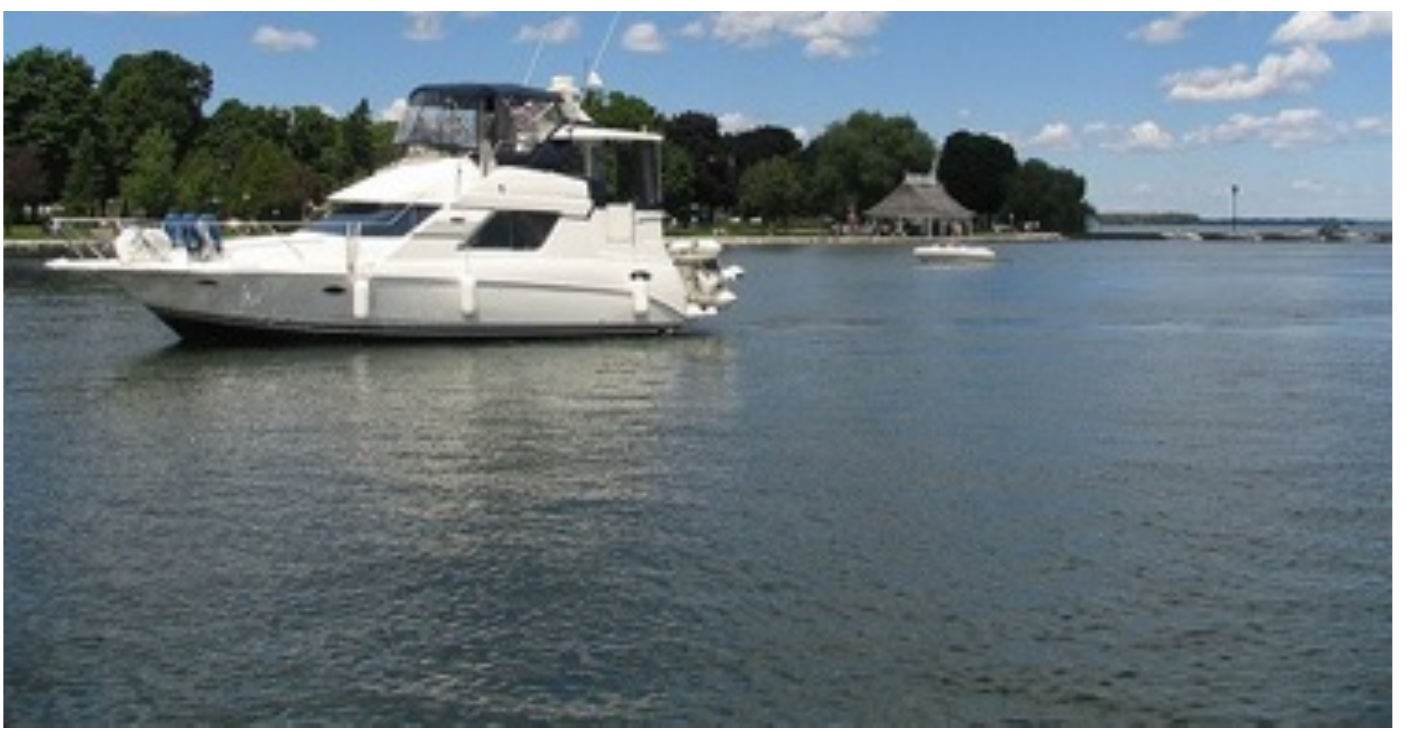}\hspace{-0.12cm}
\includegraphics[width=0.1\textwidth, height=1cm]{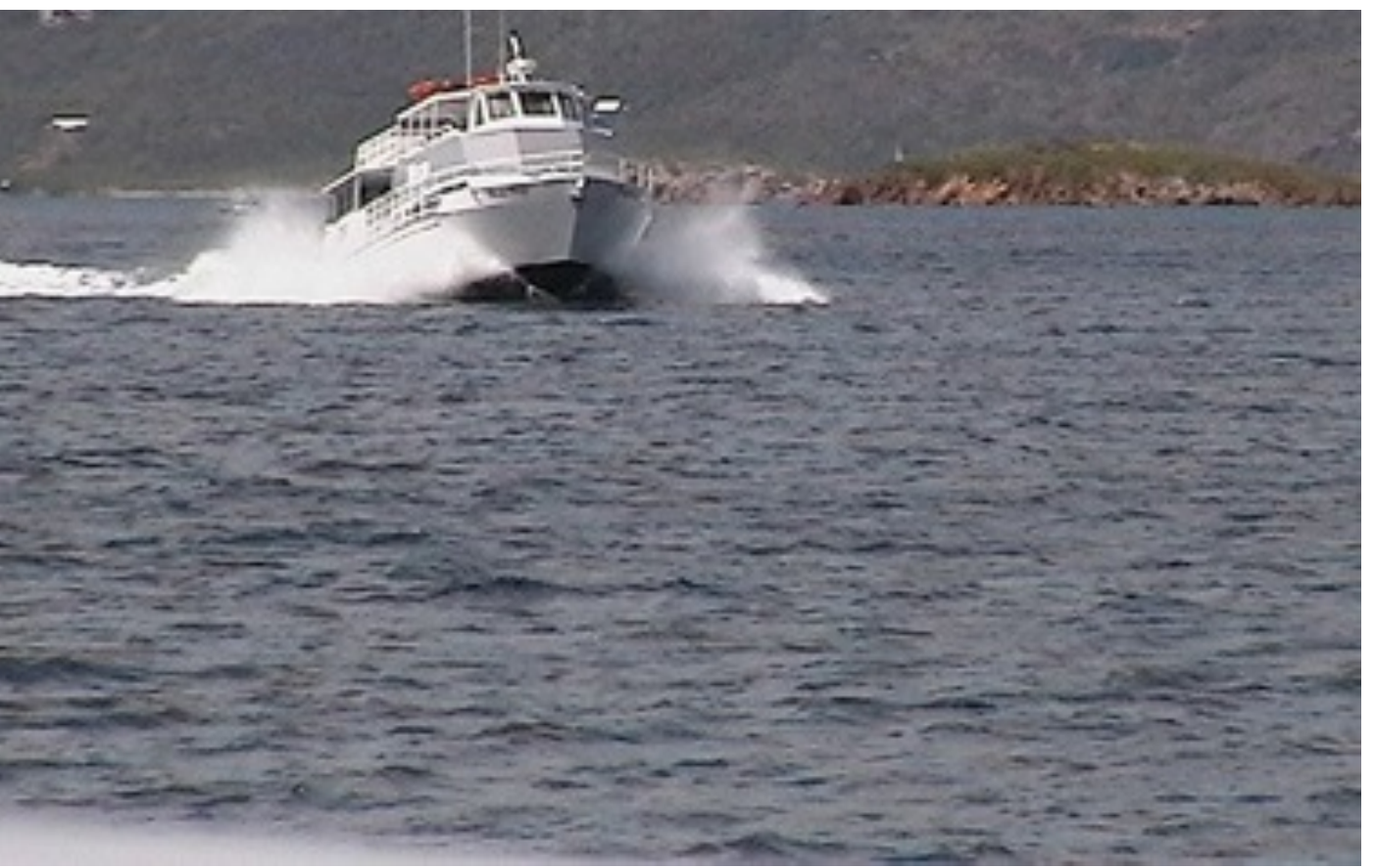}\hspace{-0.12cm}
\includegraphics[width=0.1\textwidth, height=1cm]{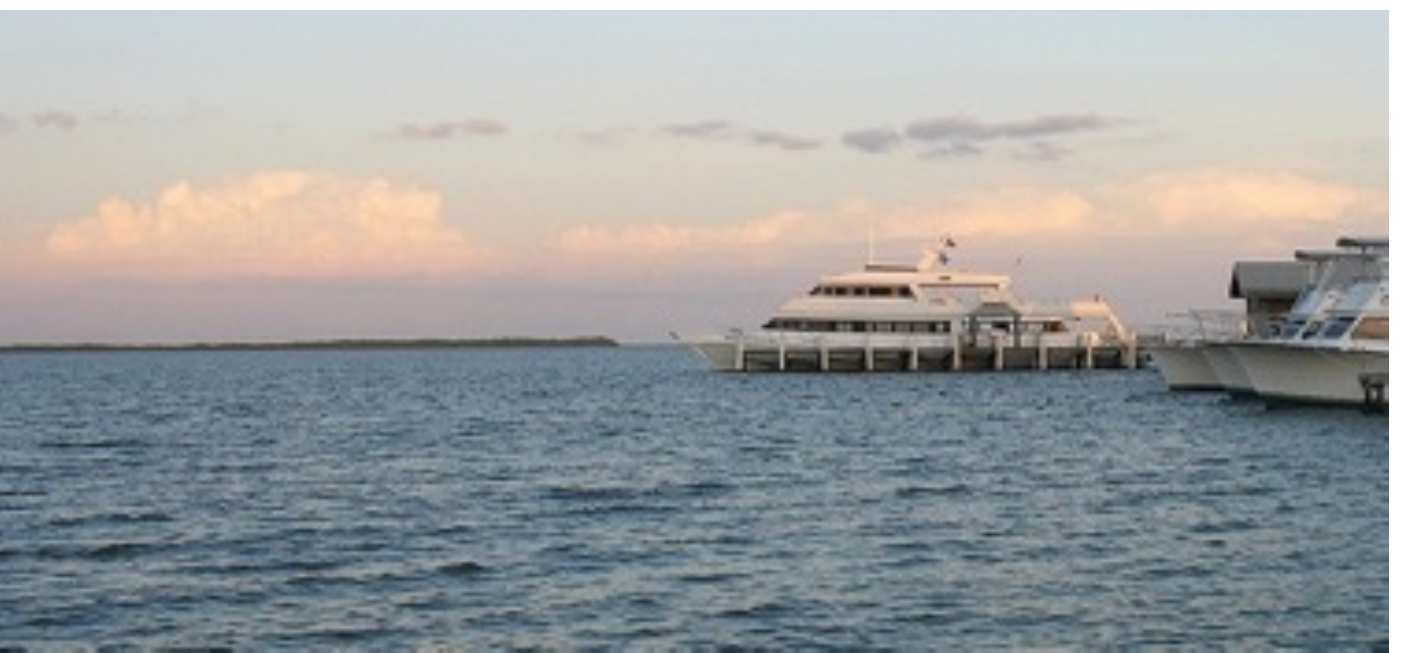}\vspace{0.2cm}\\
\includegraphics[width=0.1\textwidth, height=1cm]{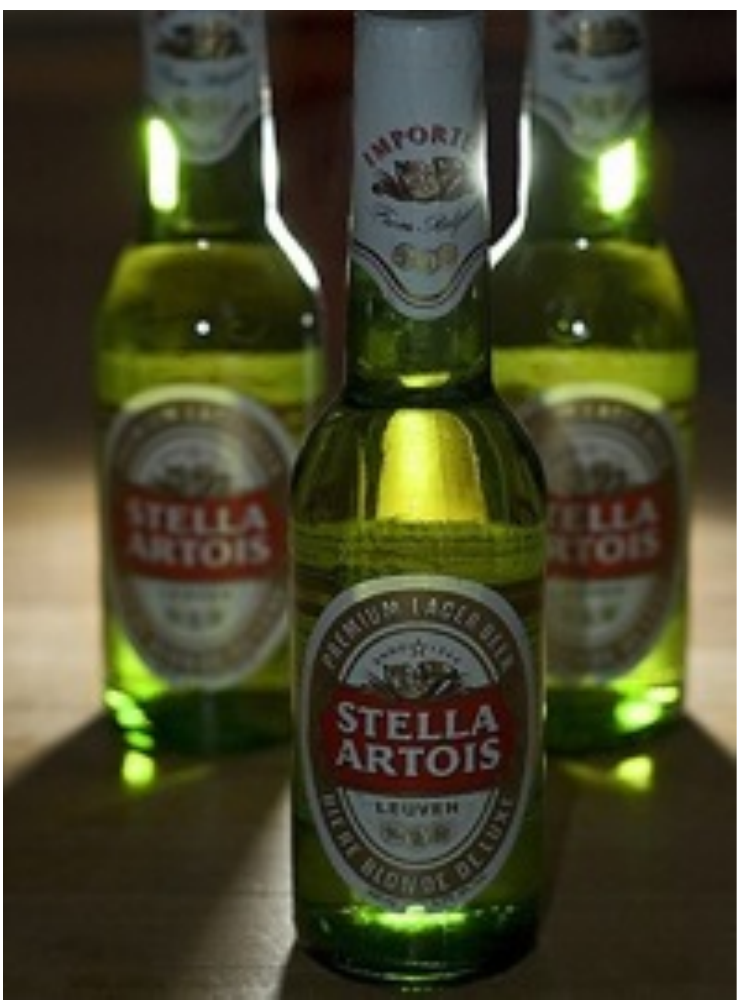}\hspace{-0.12cm}
\includegraphics[width=0.1\textwidth, height=1cm]{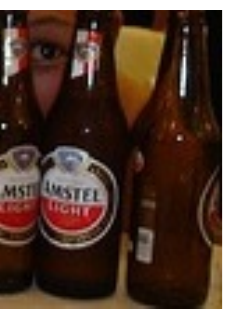}\hspace{-0.12cm}
\includegraphics[width=0.1\textwidth, height=1cm]{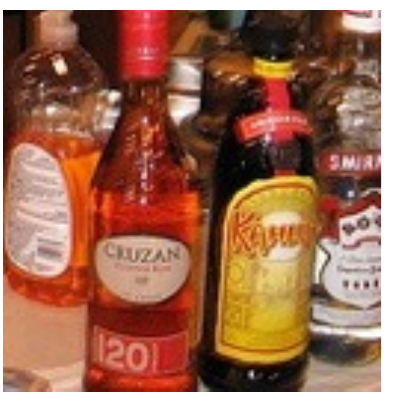}\hspace{-0.12cm}
\includegraphics[width=0.1\textwidth, height=1cm]{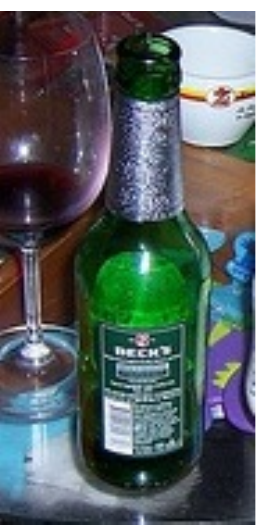}\hspace{-0.12cm}
\includegraphics[width=0.1\textwidth, height=1cm]{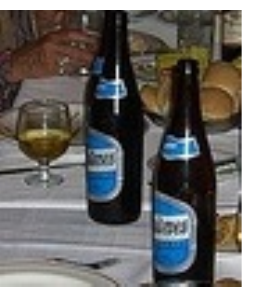}\hspace{0.2cm}
\includegraphics[width=0.1\textwidth, height=1cm]{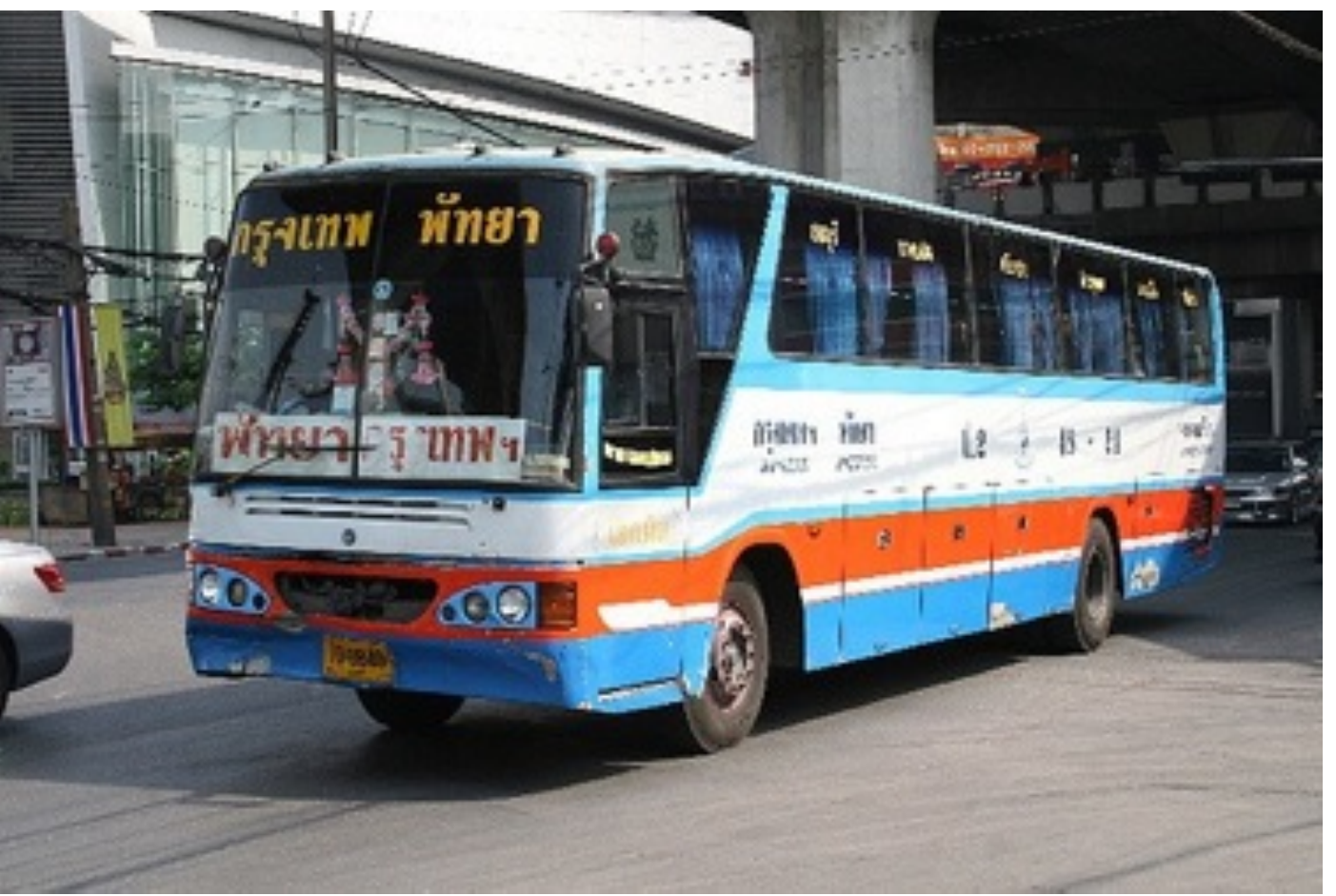}\hspace{-0.12cm}
\includegraphics[width=0.1\textwidth, height=1cm]{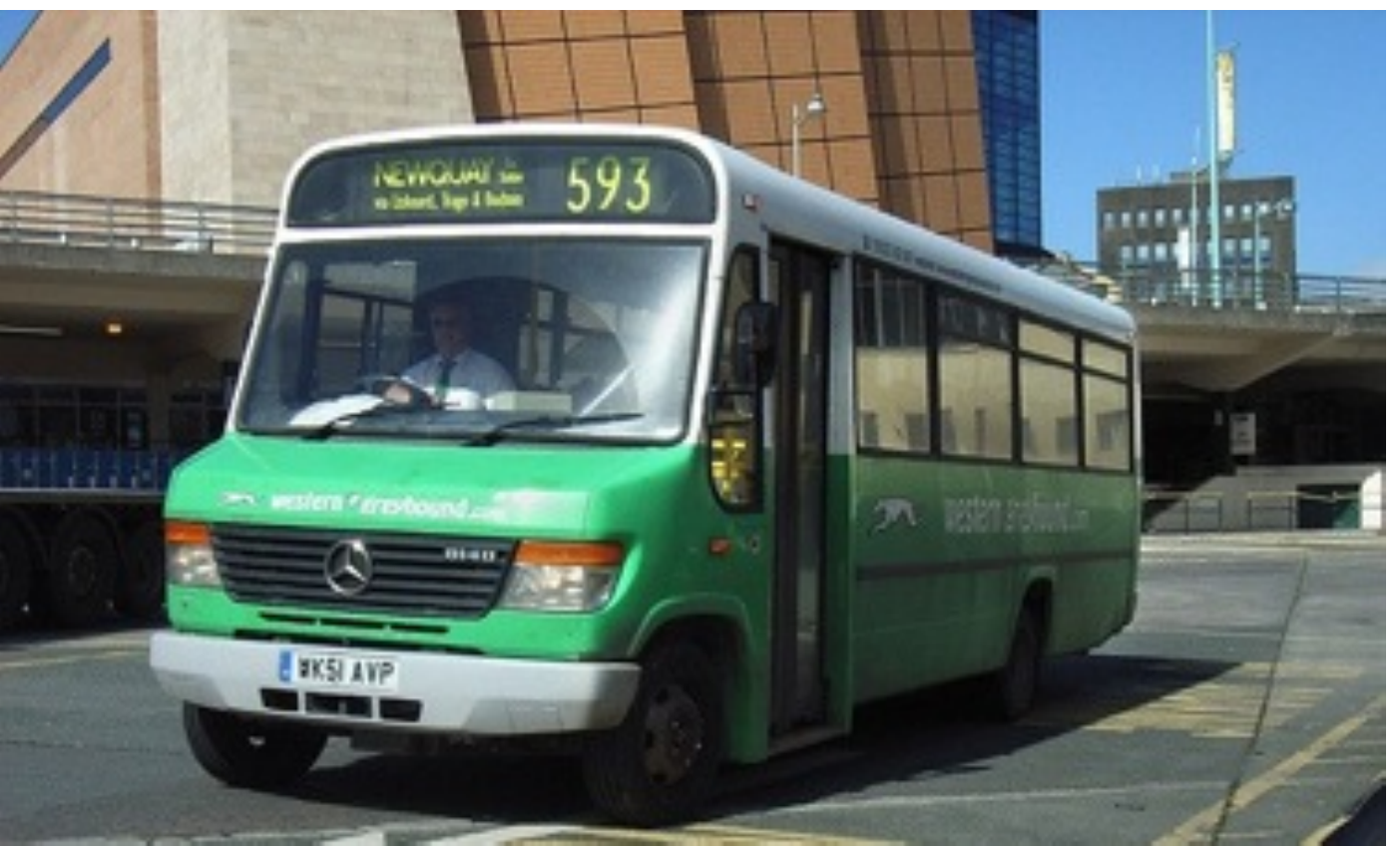}\hspace{-0.12cm}
\includegraphics[width=0.1\textwidth, height=1cm]{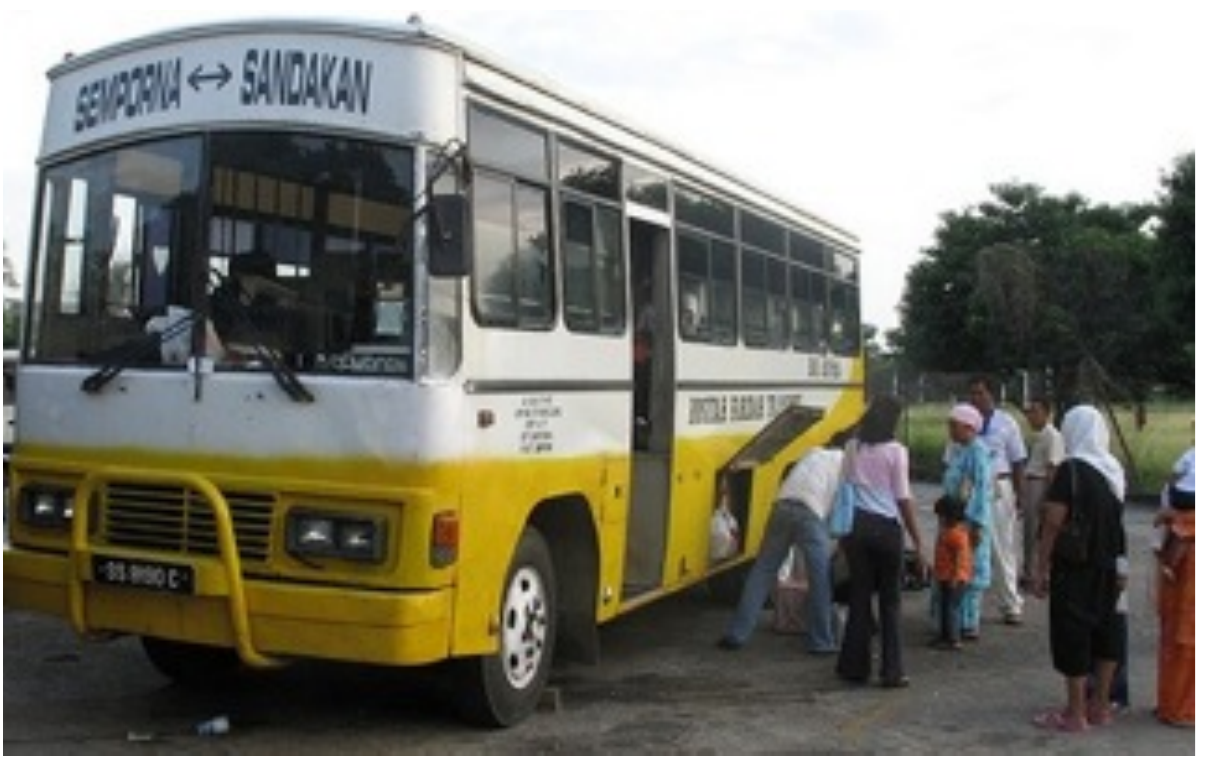}\hspace{-0.12cm}
\includegraphics[width=0.1\textwidth, height=1cm]{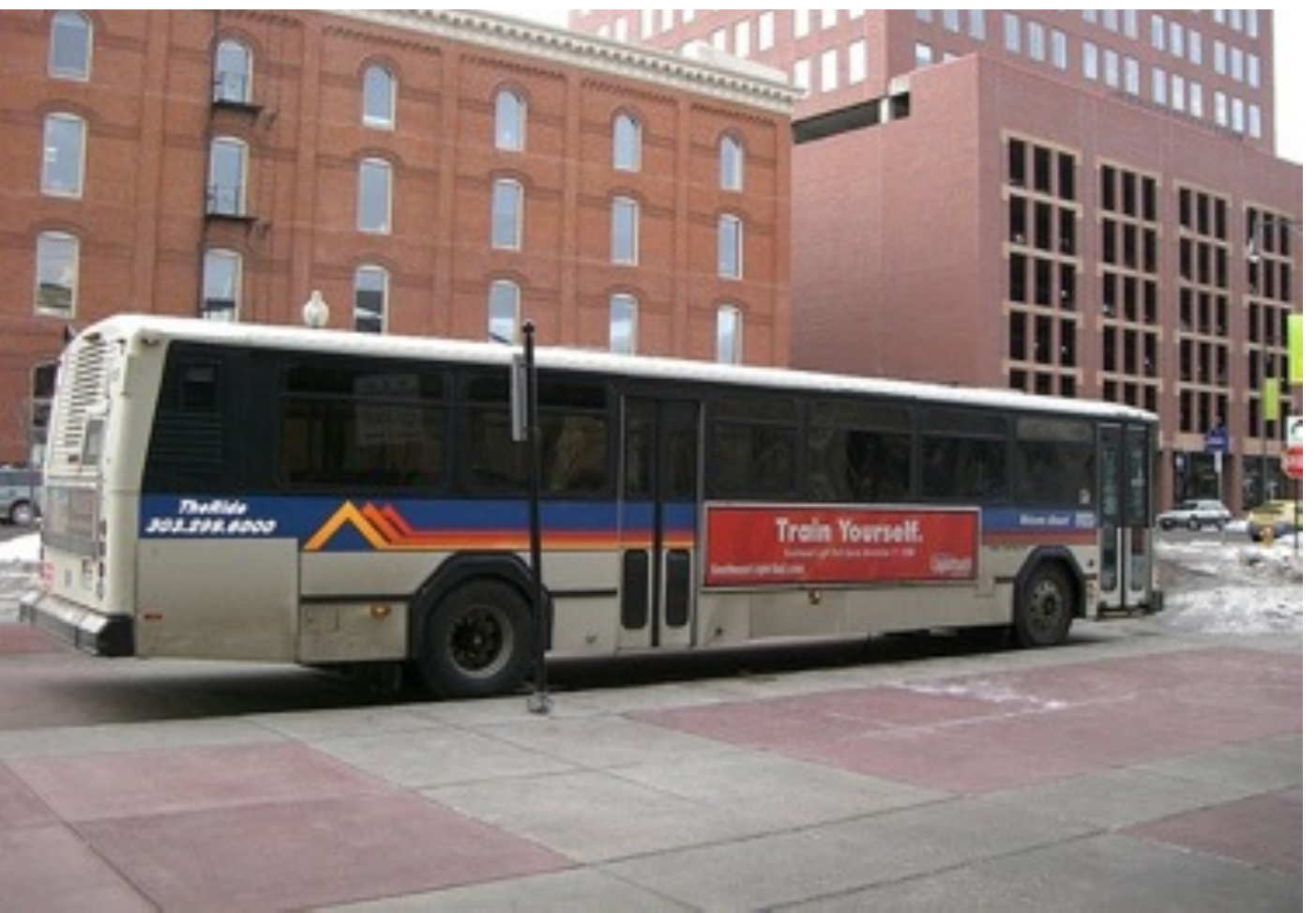}\hspace{-0.12cm}
\includegraphics[width=0.1\textwidth, height=1cm]{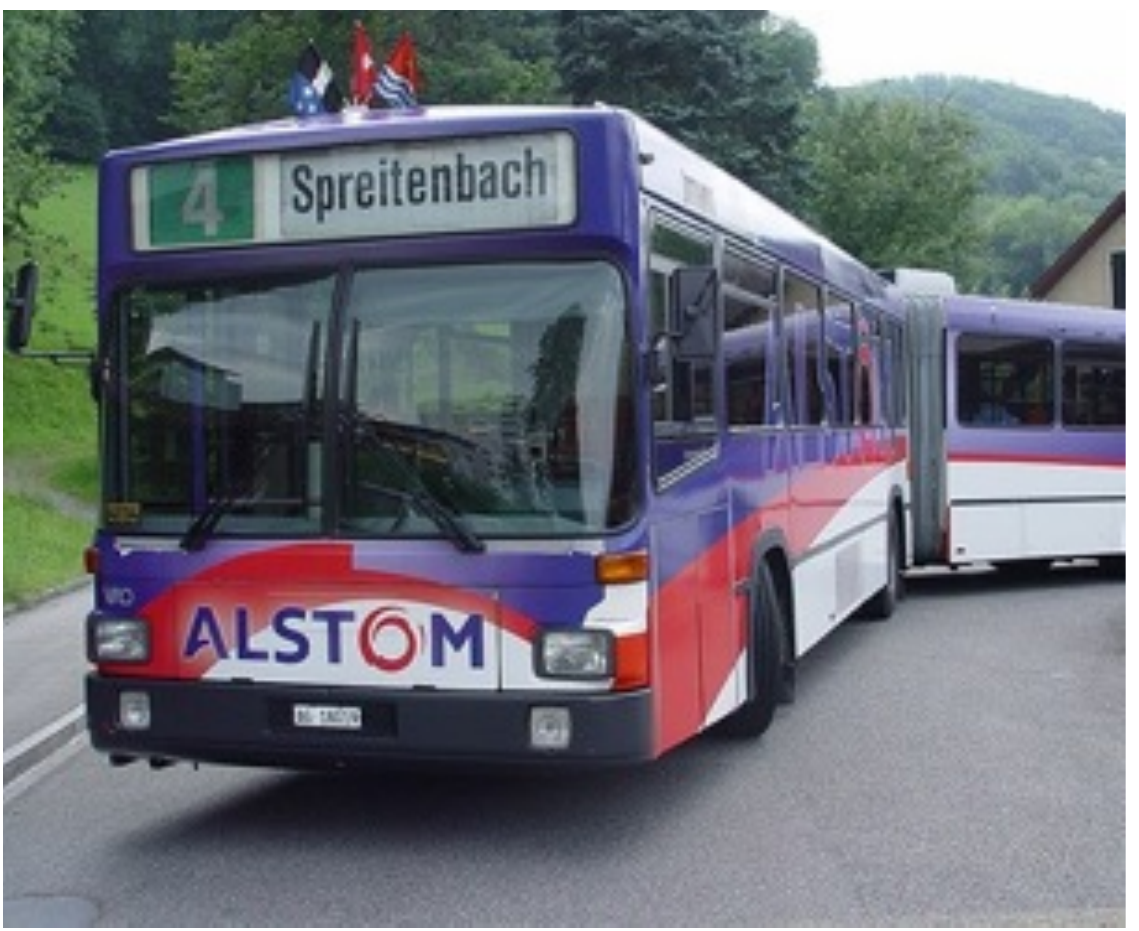}\vspace{0.2cm}\\
\includegraphics[width=0.1\textwidth, height=1cm]{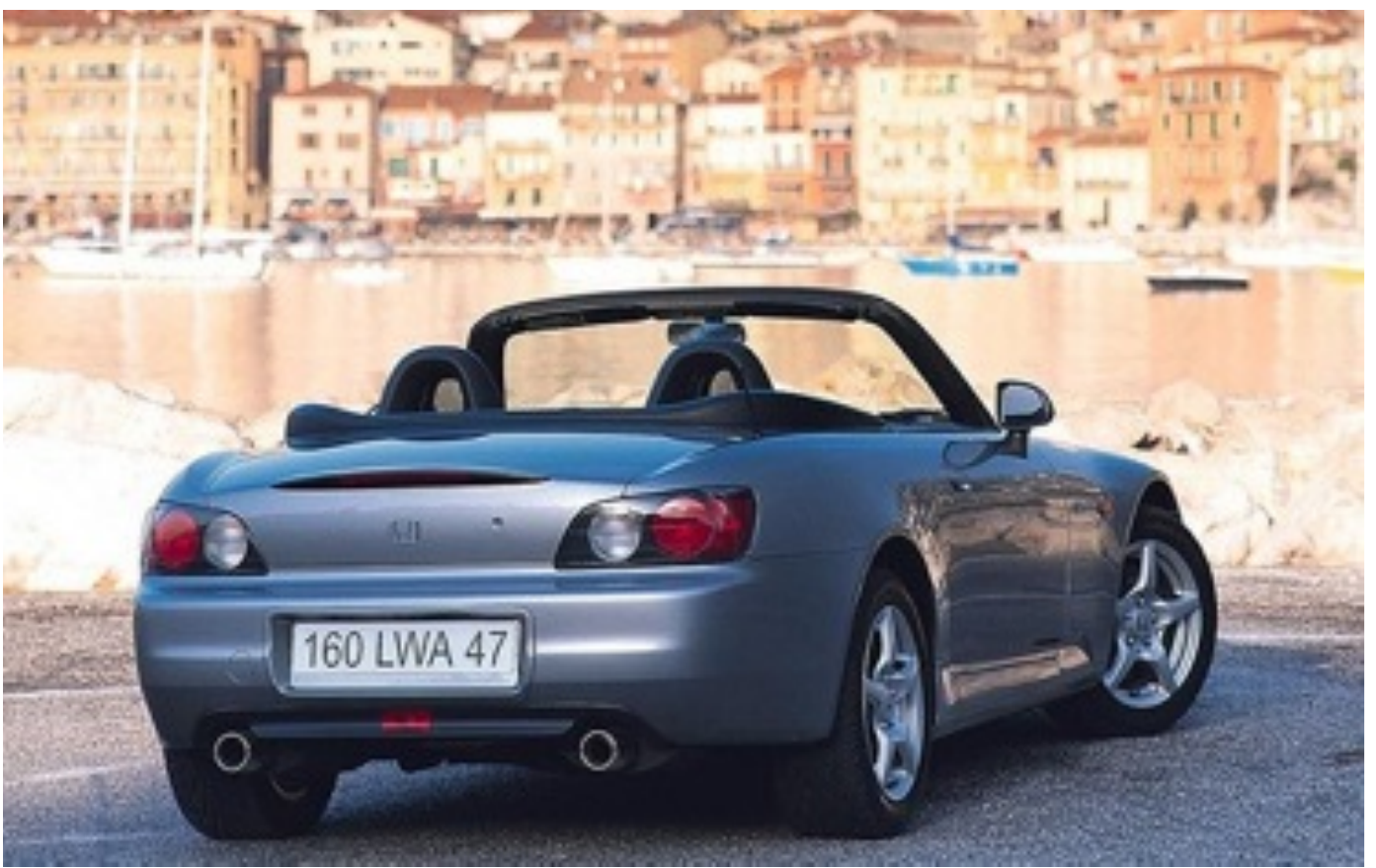}\hspace{-0.12cm}
\includegraphics[width=0.1\textwidth, height=1cm]{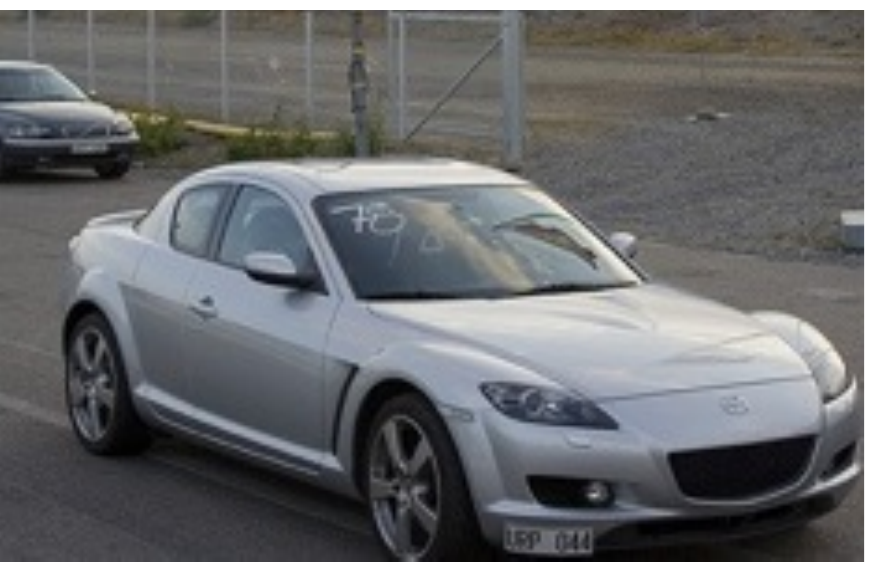}\hspace{-0.12cm}
\includegraphics[width=0.1\textwidth, height=1cm]{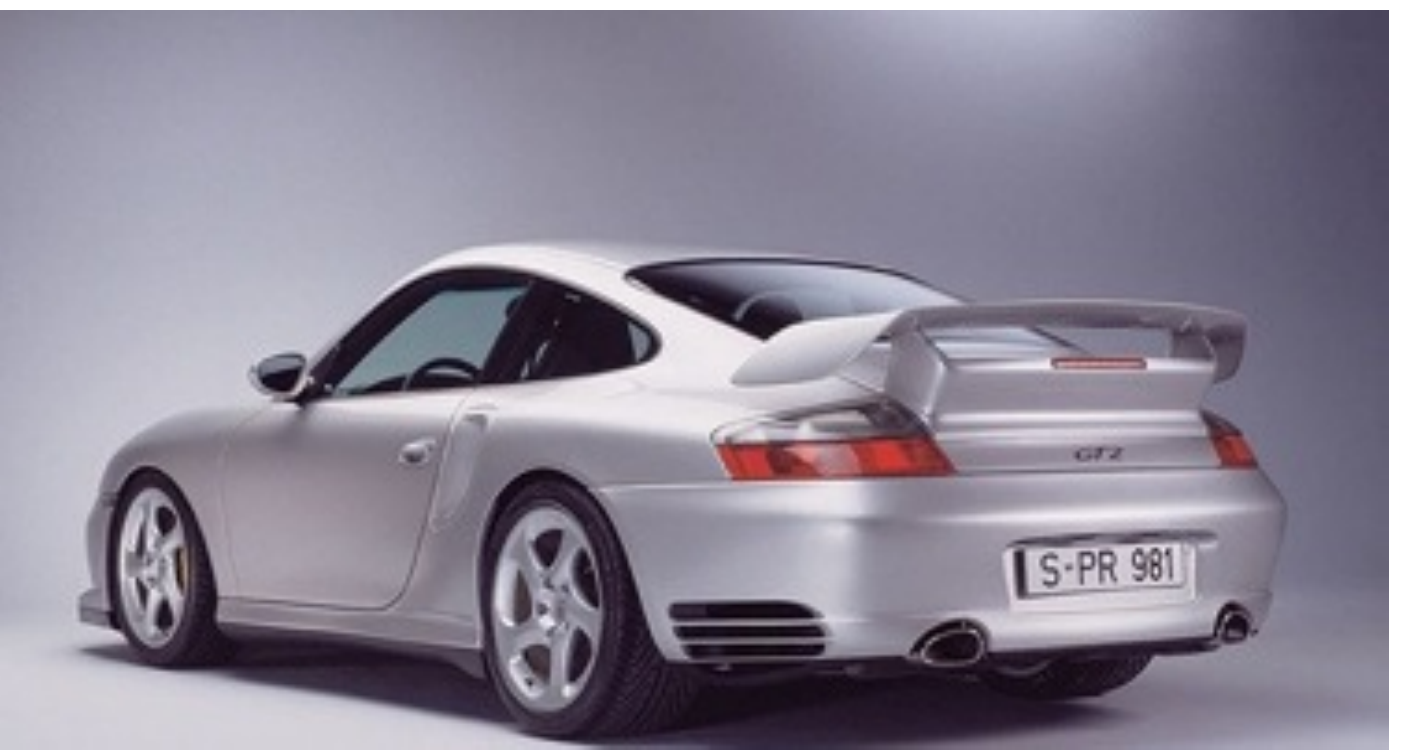}\hspace{-0.12cm}
\includegraphics[width=0.1\textwidth, height=1cm]{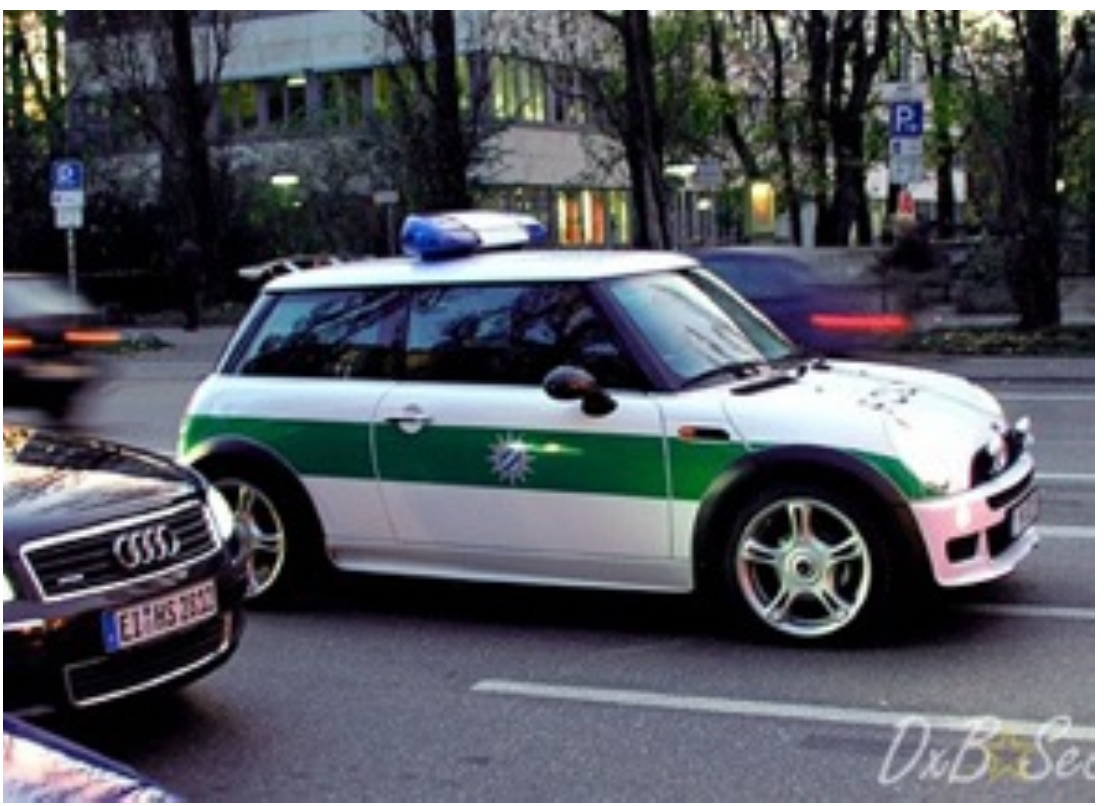}\hspace{-0.12cm}
\includegraphics[width=0.1\textwidth, height=1cm]{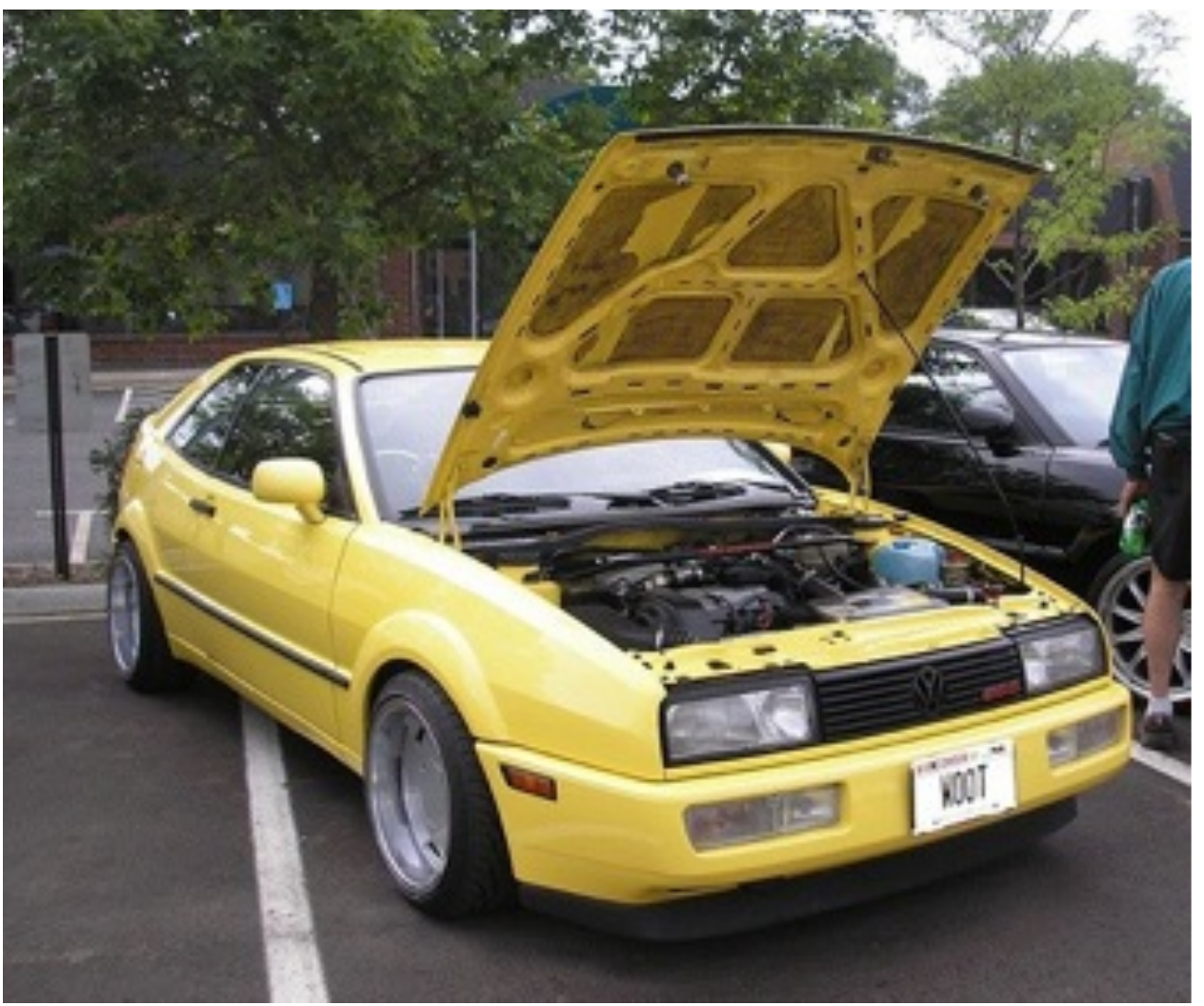}\hspace{0.2cm}
\includegraphics[width=0.1\textwidth, height=1cm]{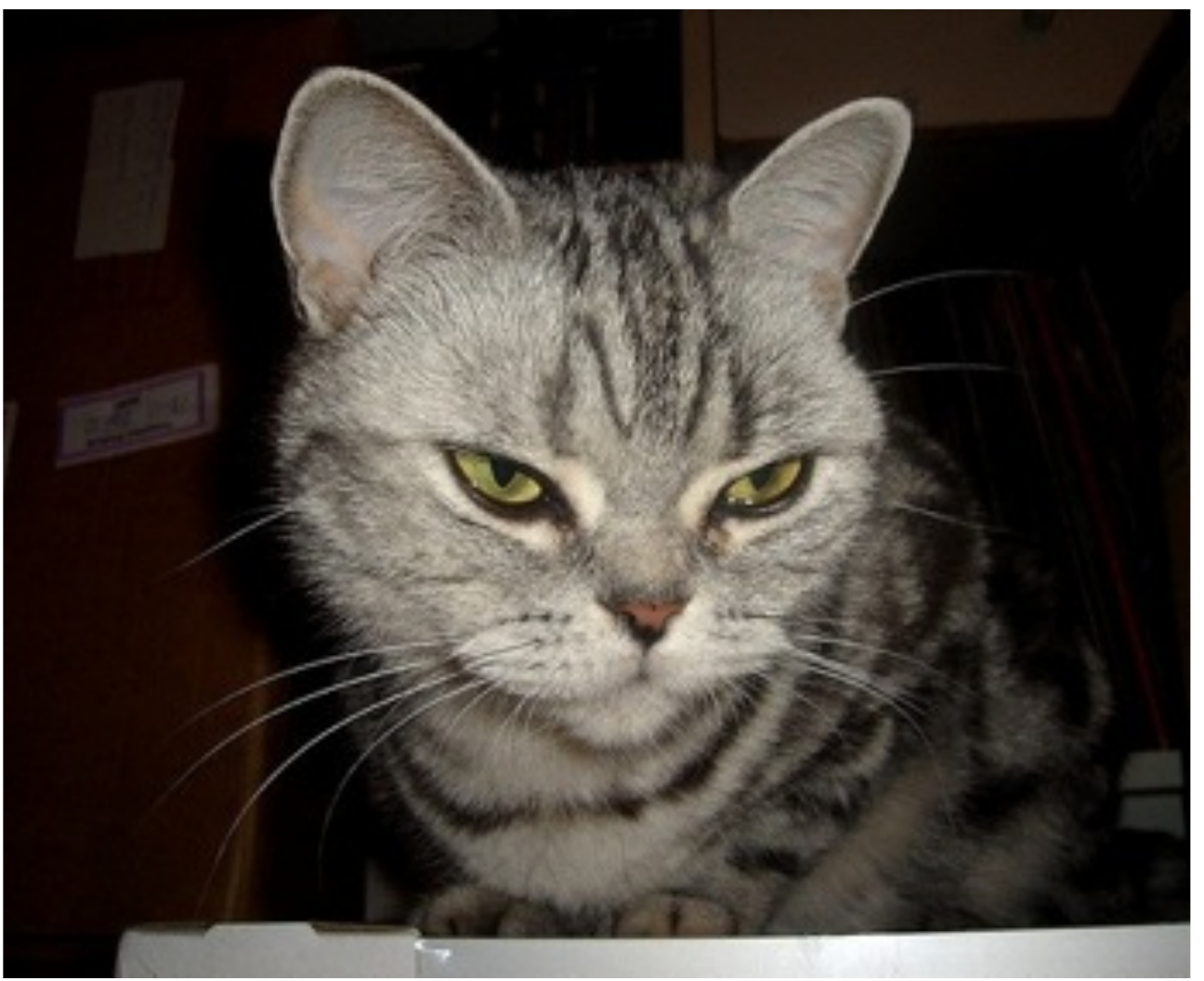}\hspace{-0.12cm}
\includegraphics[width=0.1\textwidth, height=1cm]{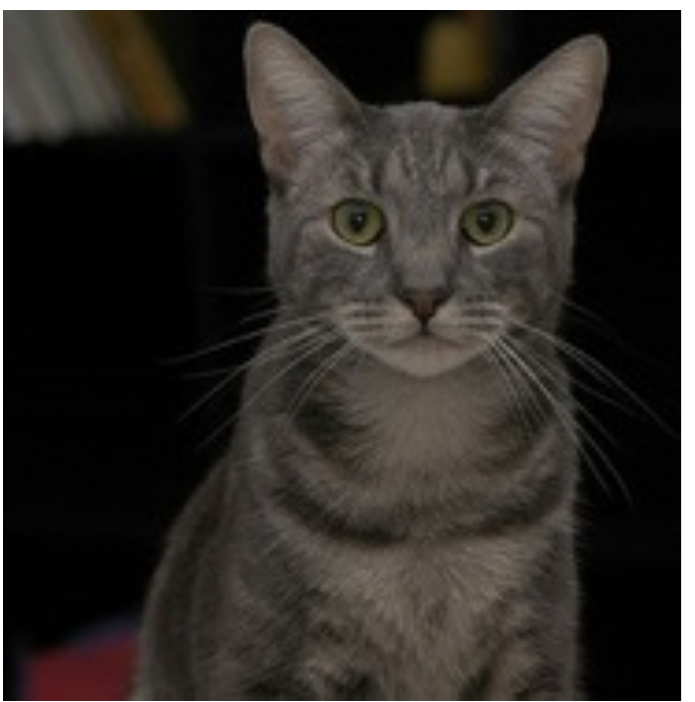}\hspace{-0.12cm}
\includegraphics[width=0.1\textwidth, height=1cm]{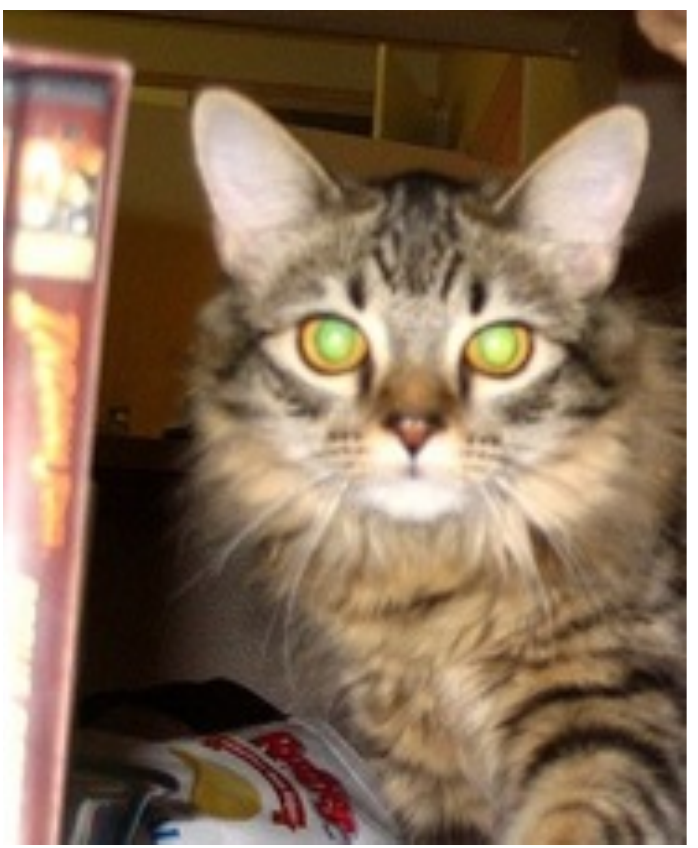}\hspace{-0.12cm}
\includegraphics[width=0.1\textwidth, height=1cm]{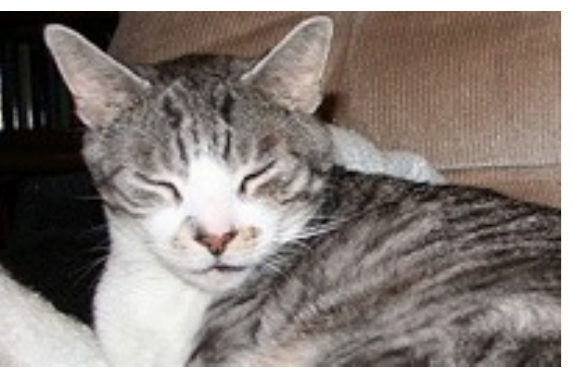}\hspace{-0.12cm}
\includegraphics[width=0.1\textwidth, height=1cm]{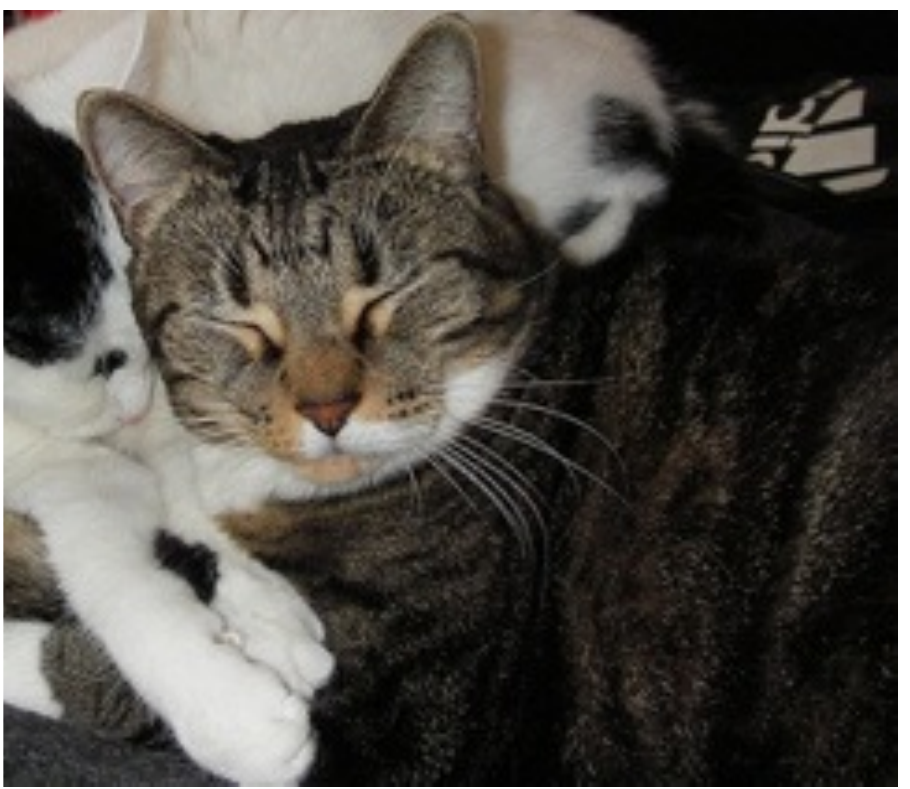}\vspace{0.2cm}\\
\includegraphics[width=0.1\textwidth, height=1cm]{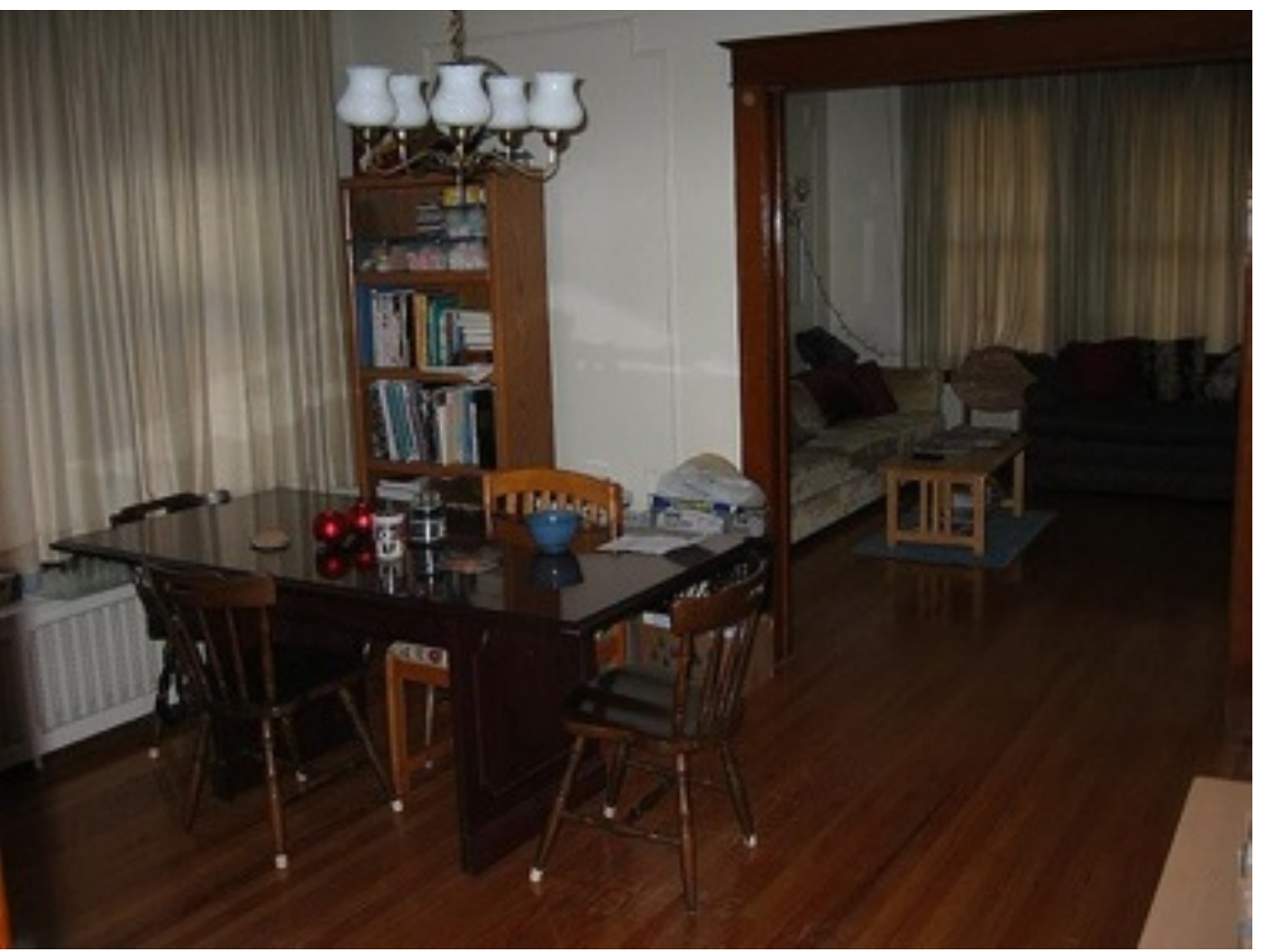}\hspace{-0.12cm}
\includegraphics[width=0.1\textwidth, height=1cm]{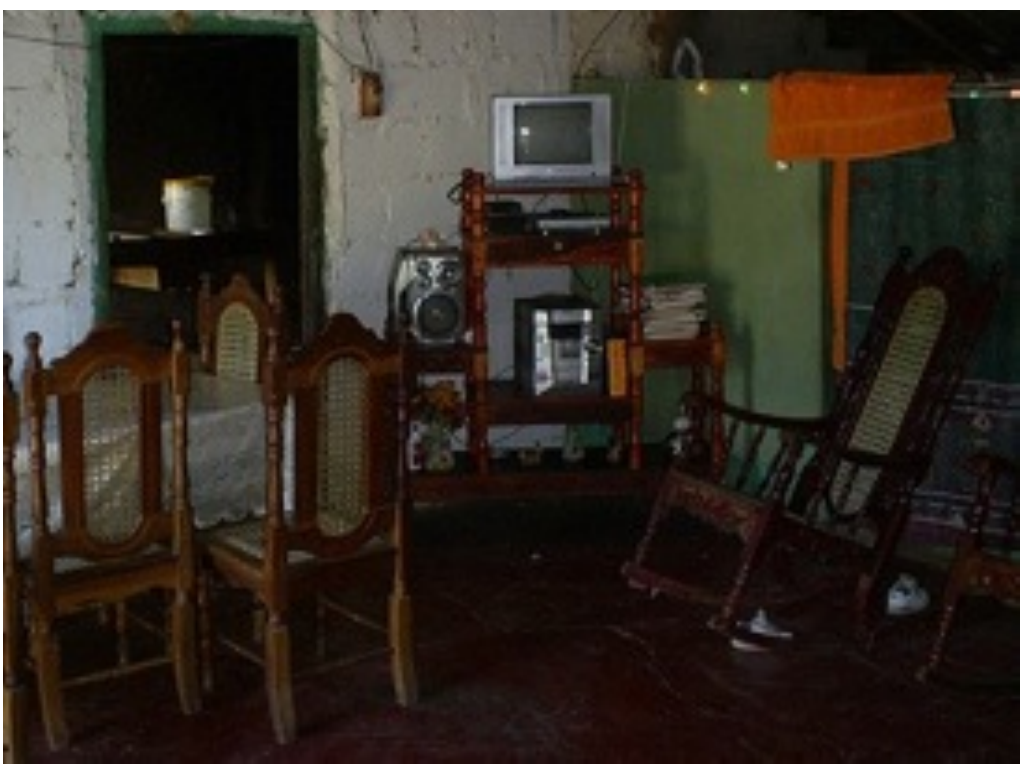}\hspace{-0.12cm}
\includegraphics[width=0.1\textwidth, height=1cm]{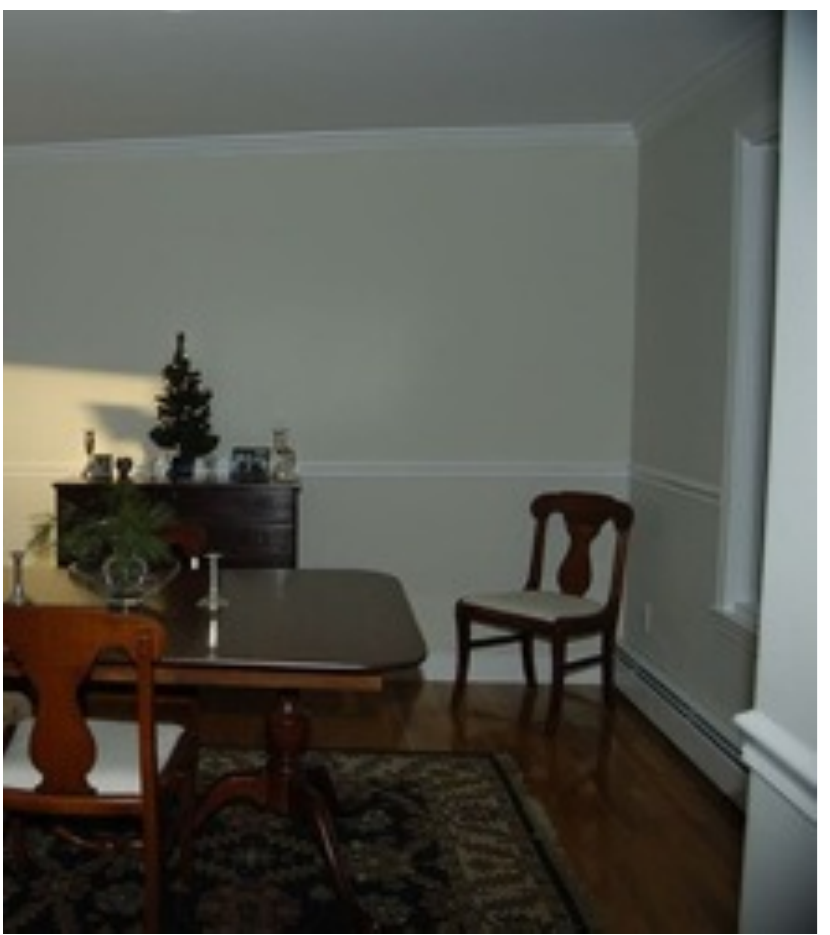}\hspace{-0.12cm}
\includegraphics[width=0.1\textwidth, height=1cm]{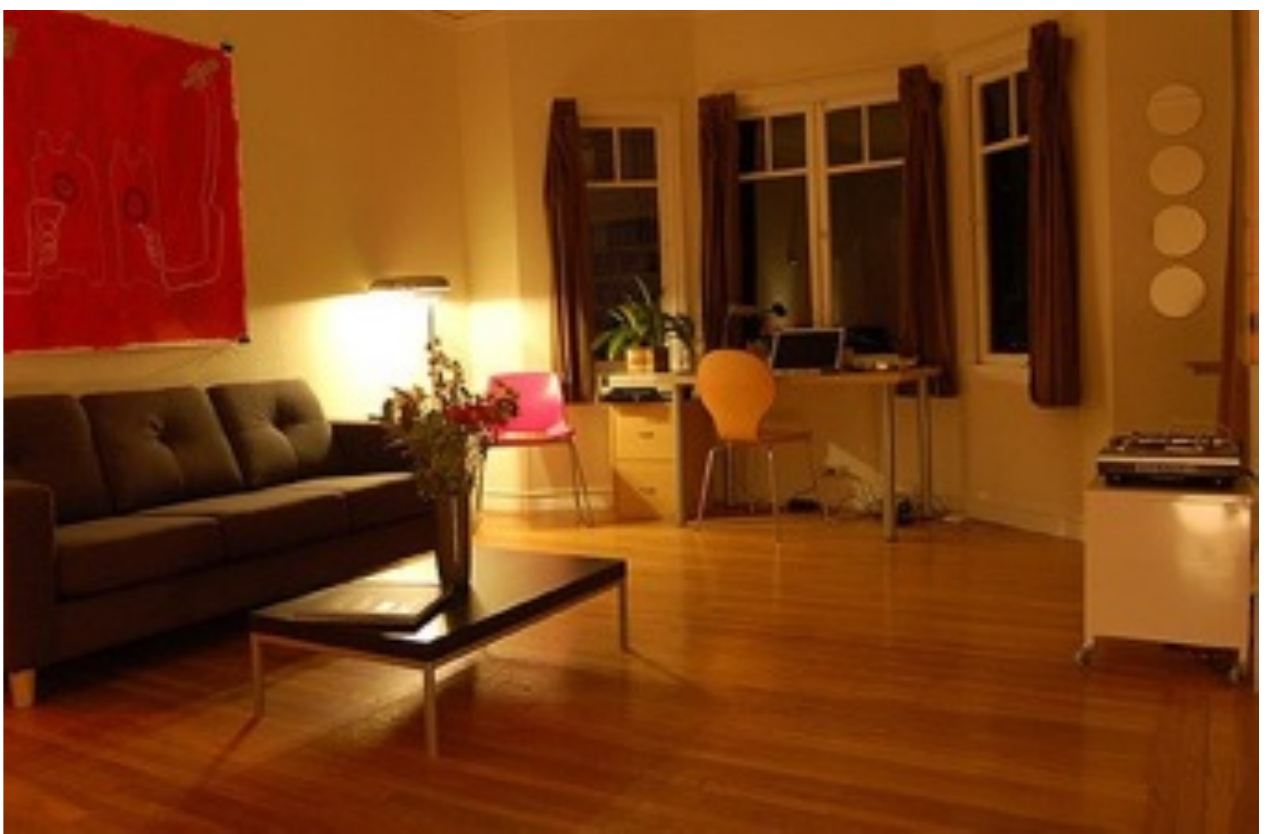}\hspace{-0.12cm}
\includegraphics[width=0.1\textwidth, height=1cm]{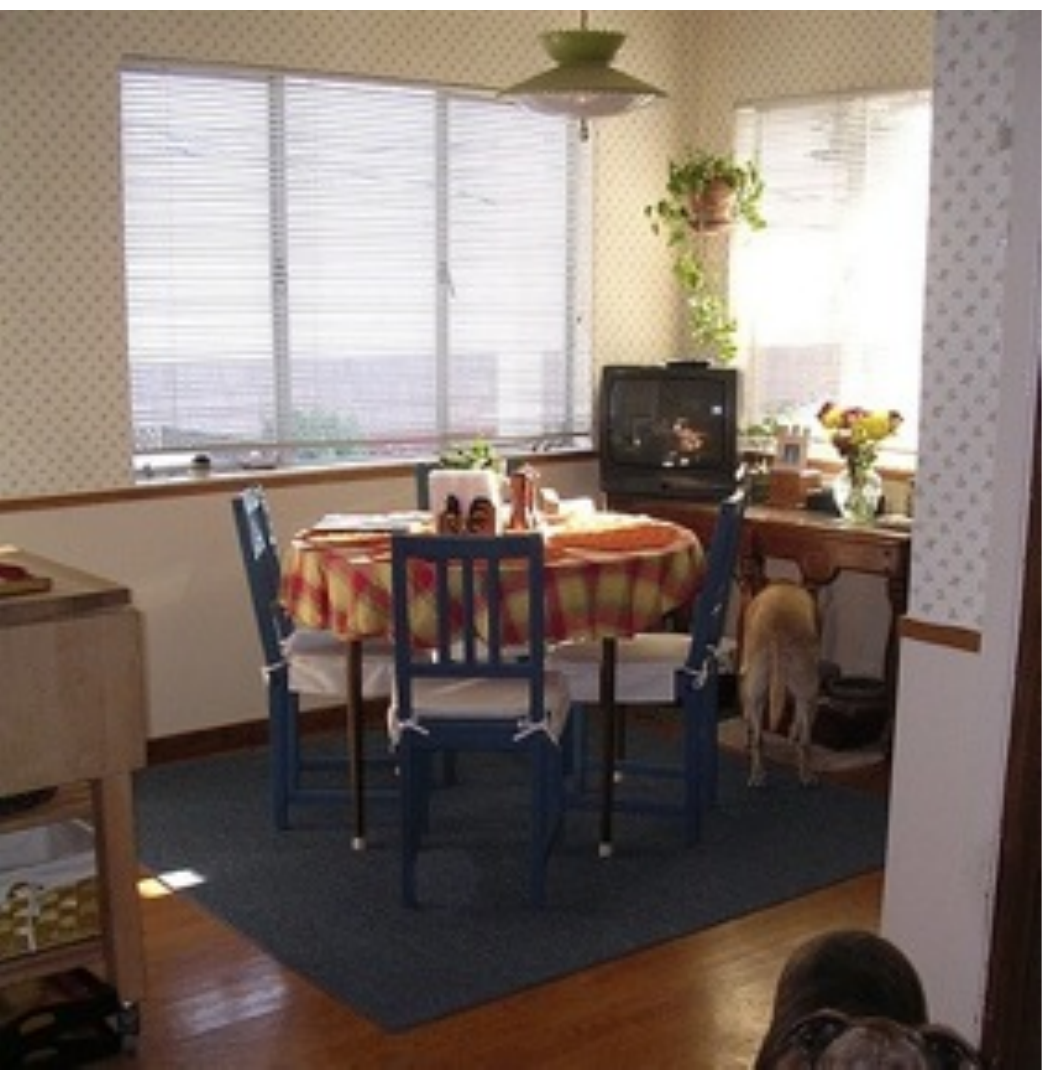}\hspace{0.2cm}
\includegraphics[width=0.1\textwidth, height=1cm]{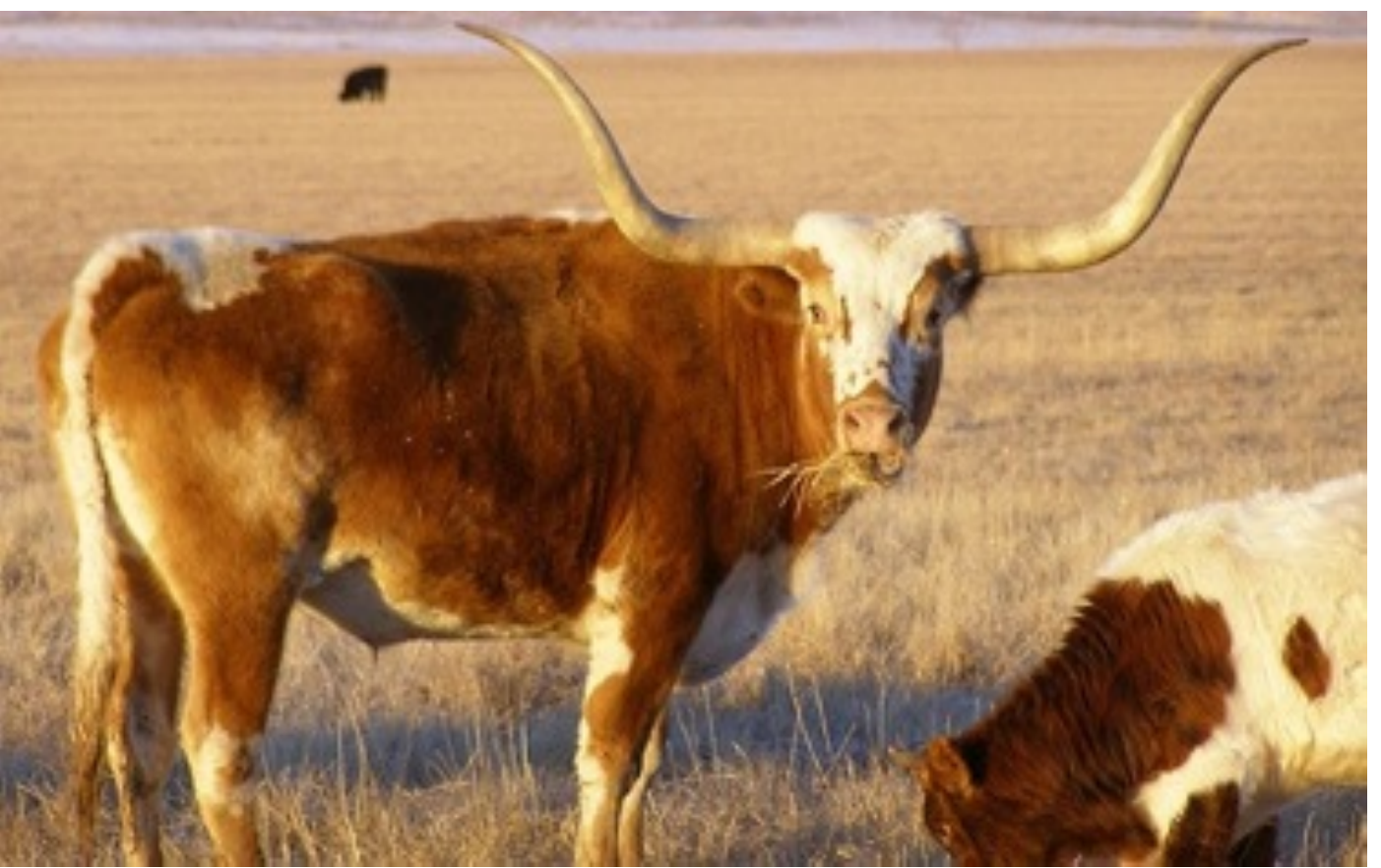}\hspace{-0.12cm}
\includegraphics[width=0.1\textwidth, height=1cm]{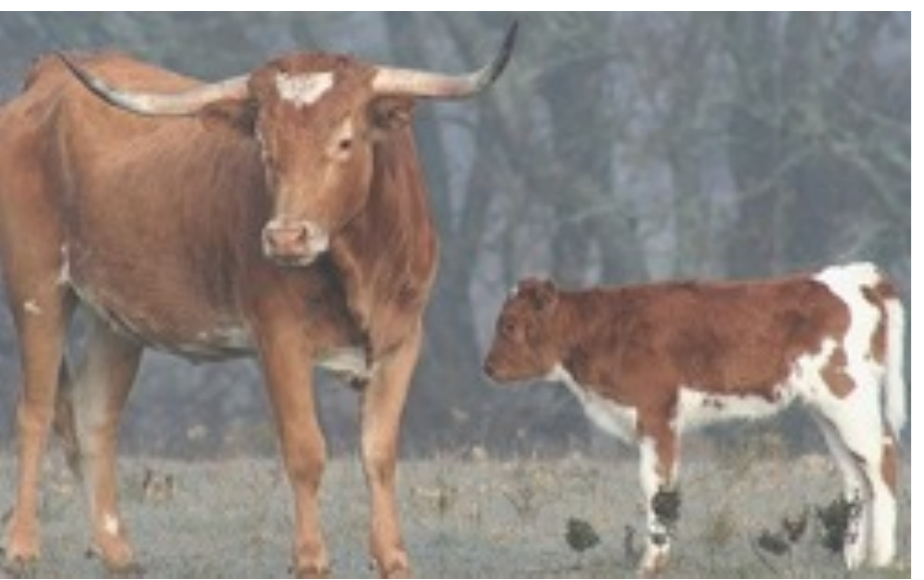}\hspace{-0.12cm}
\includegraphics[width=0.1\textwidth, height=1cm]{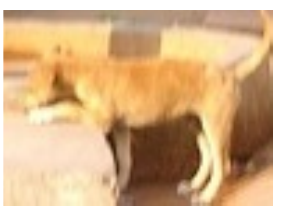}\hspace{-0.12cm}
\includegraphics[width=0.1\textwidth, height=1cm]{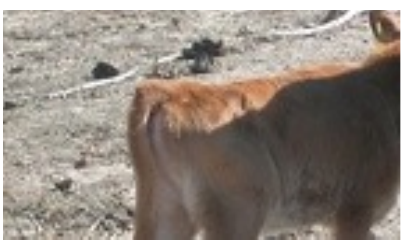}\hspace{-0.12cm}
\includegraphics[width=0.1\textwidth, height=1cm]{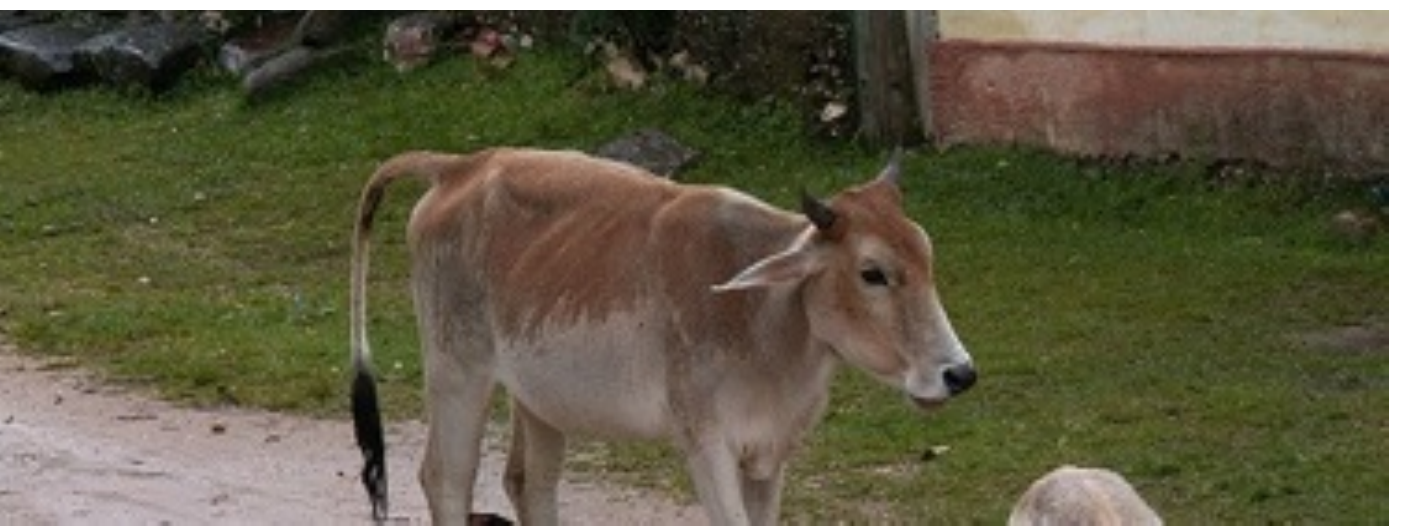}\vspace{0.2cm}\\
\includegraphics[width=0.1\textwidth, height=1cm]{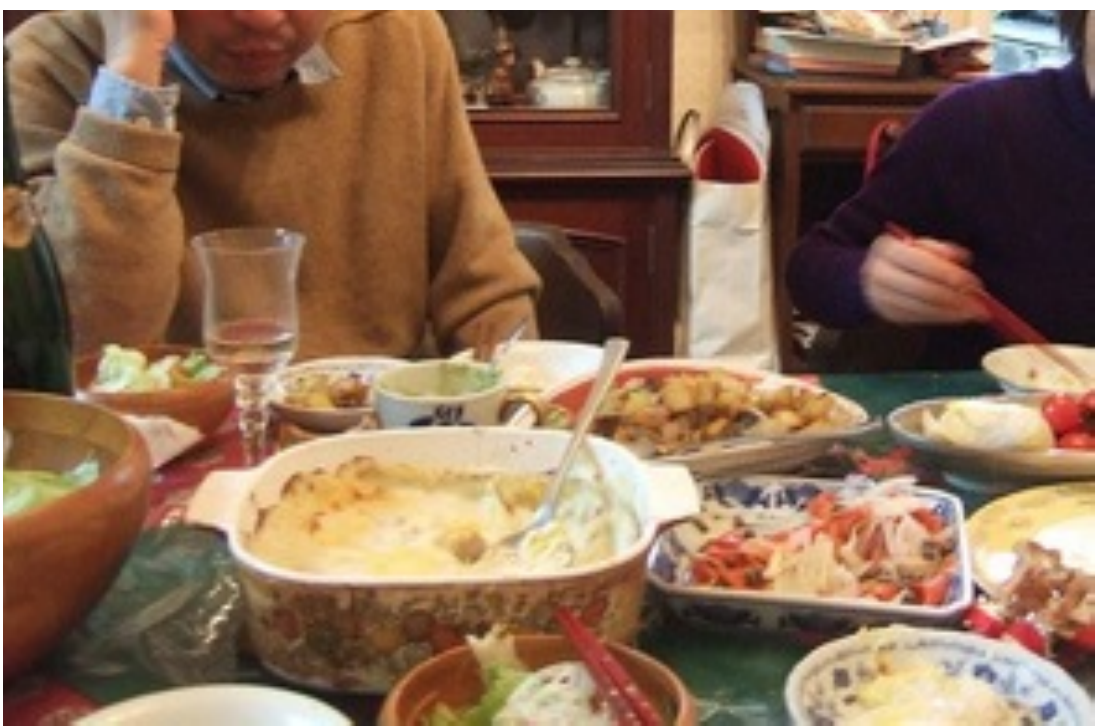}\hspace{-0.12cm}
\includegraphics[width=0.1\textwidth, height=1cm]{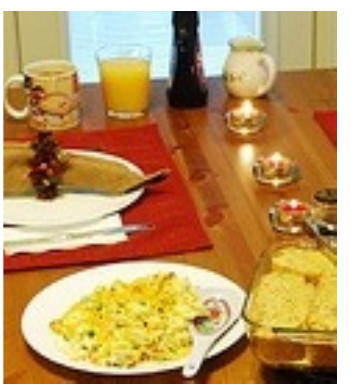}\hspace{-0.12cm}
\includegraphics[width=0.1\textwidth, height=1cm]{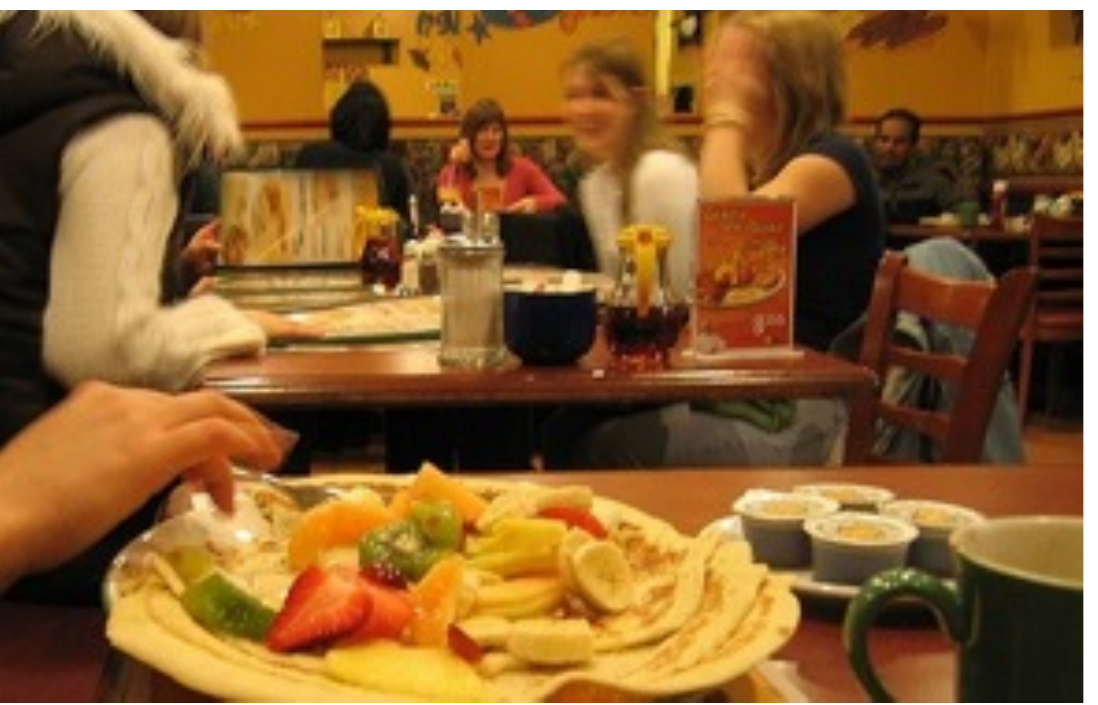}\hspace{-0.12cm}
\includegraphics[width=0.1\textwidth, height=1cm]{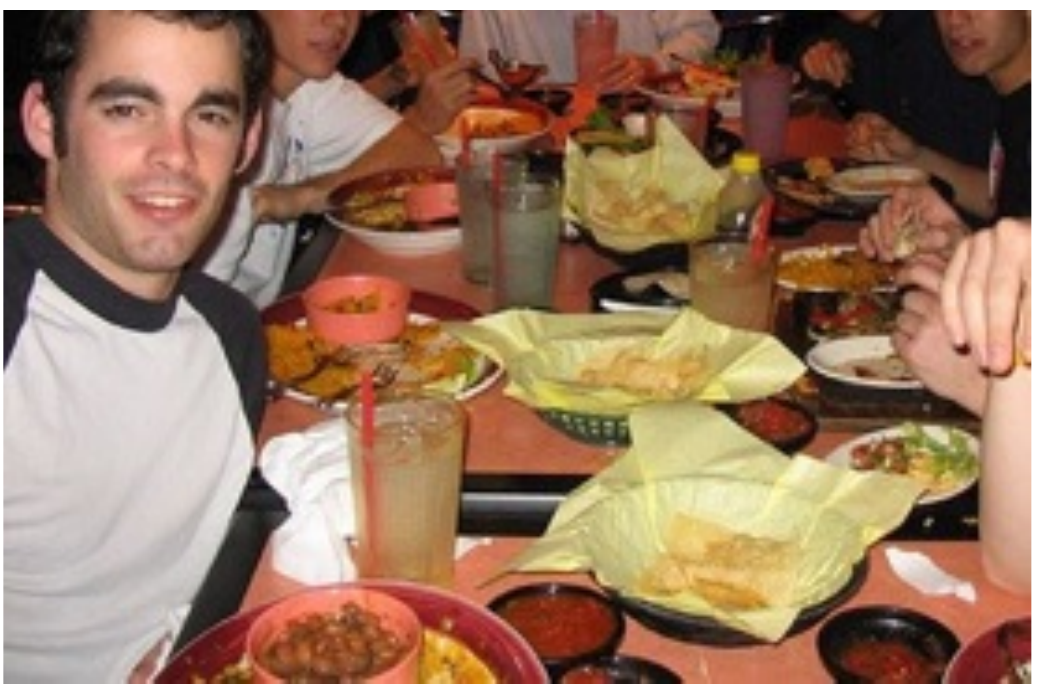}\hspace{-0.12cm}
\includegraphics[width=0.1\textwidth, height=1cm]{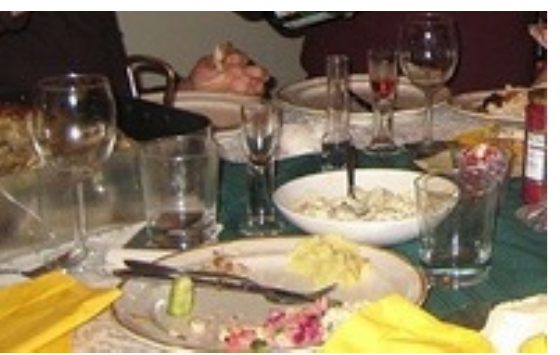}\hspace{0.2cm}
\includegraphics[width=0.1\textwidth, height=1cm]{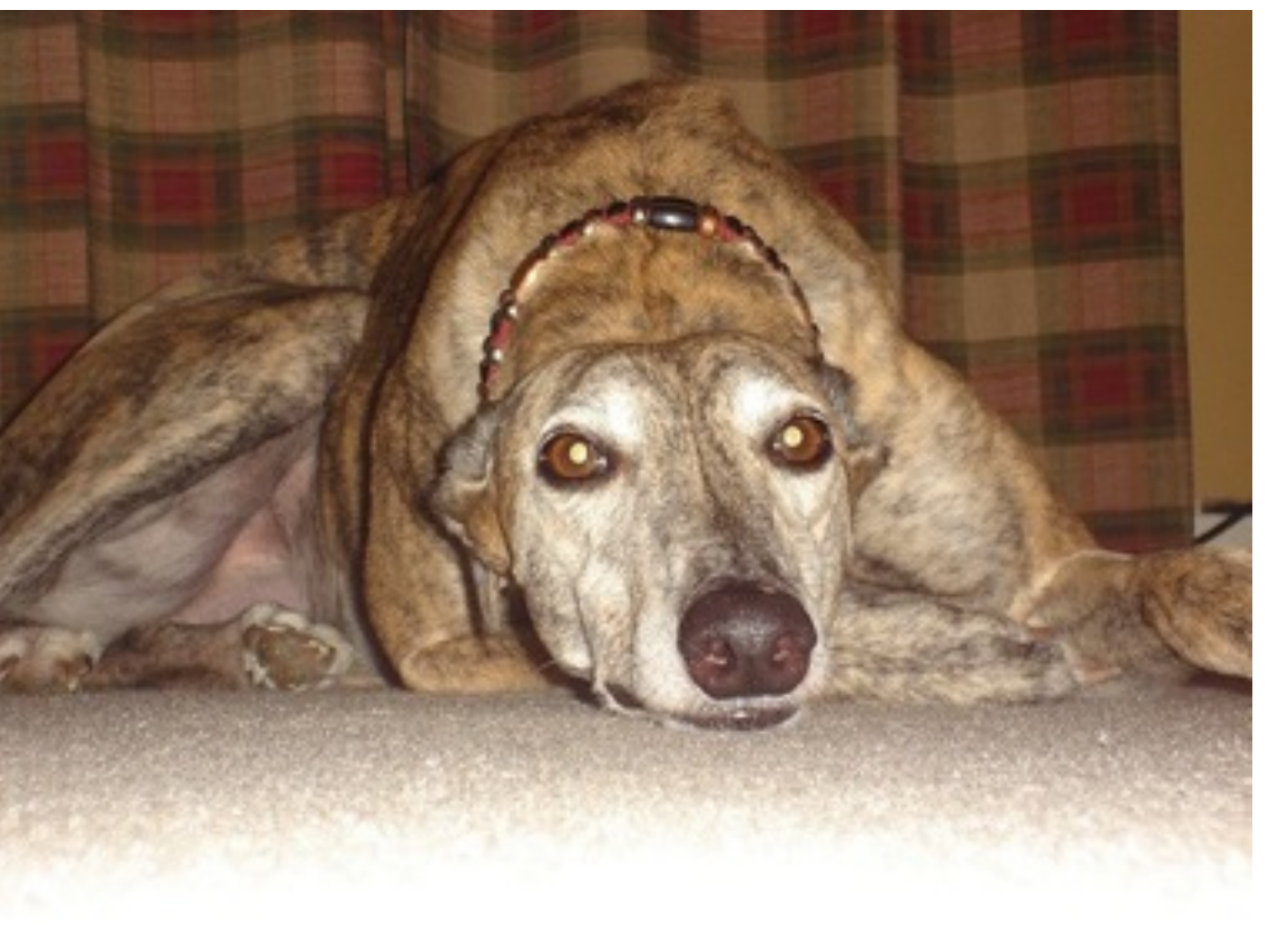}\hspace{-0.12cm}
\includegraphics[width=0.1\textwidth, height=1cm]{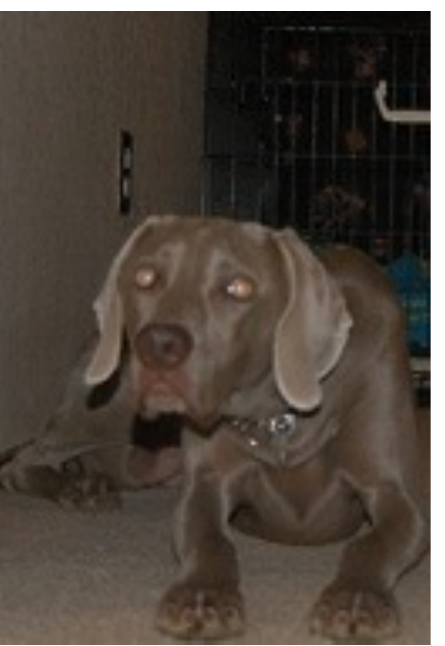}\hspace{-0.12cm}
\includegraphics[width=0.1\textwidth, height=1cm]{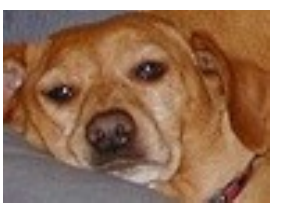}\hspace{-0.12cm}
\includegraphics[width=0.1\textwidth, height=1cm]{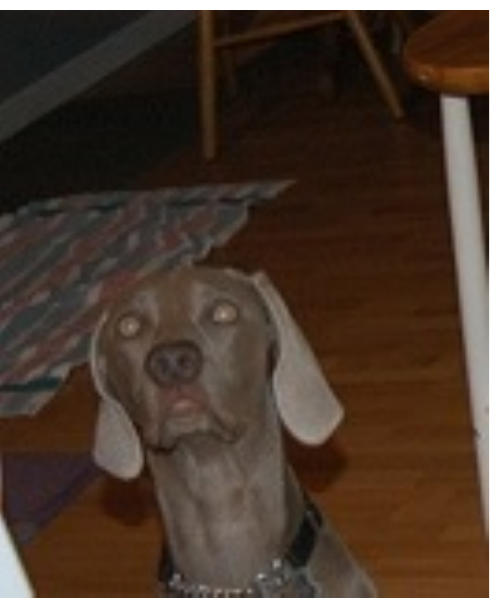}\hspace{-0.12cm}
\includegraphics[width=0.1\textwidth, height=1cm]{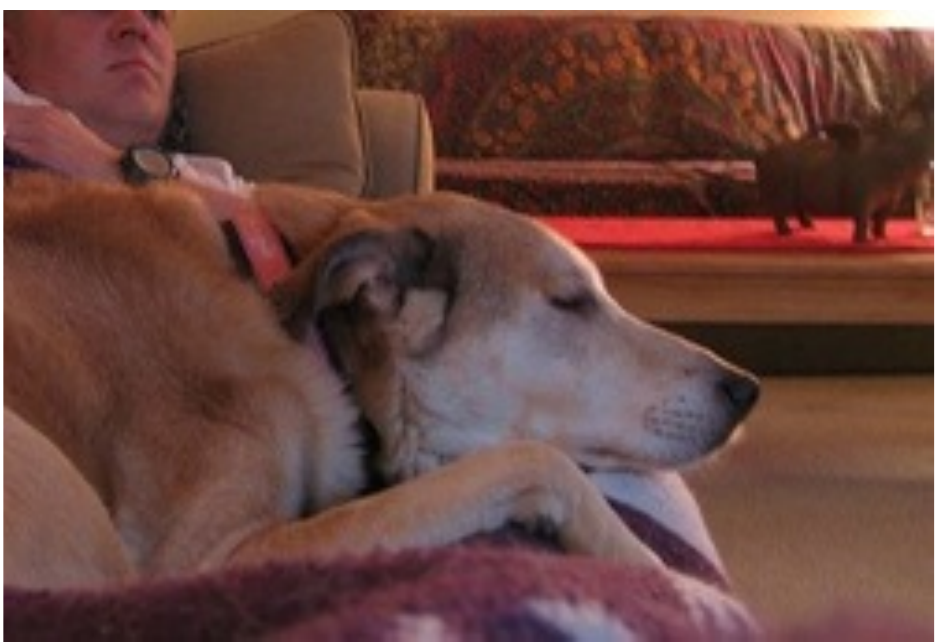}\vspace{0.2cm}\\
\includegraphics[width=0.1\textwidth, height=1cm]{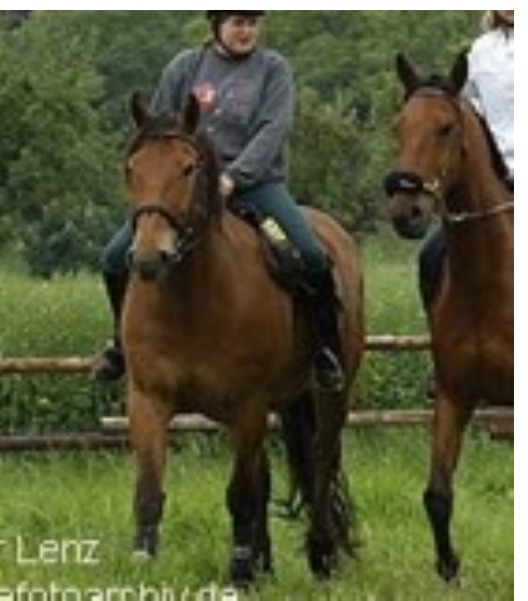}\hspace{-0.12cm}
\includegraphics[width=0.1\textwidth, height=1cm]{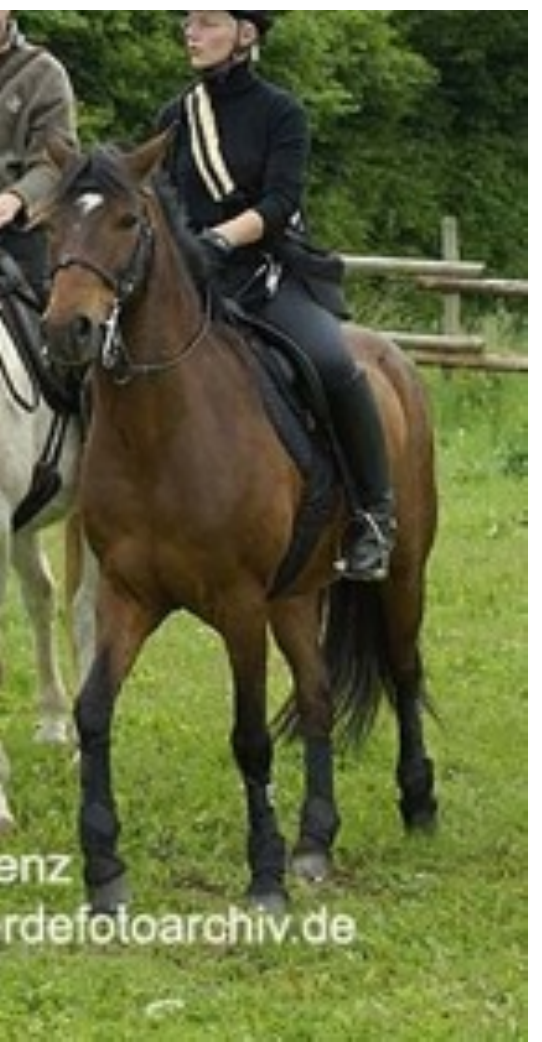}\hspace{-0.12cm}
\includegraphics[width=0.1\textwidth, height=1cm]{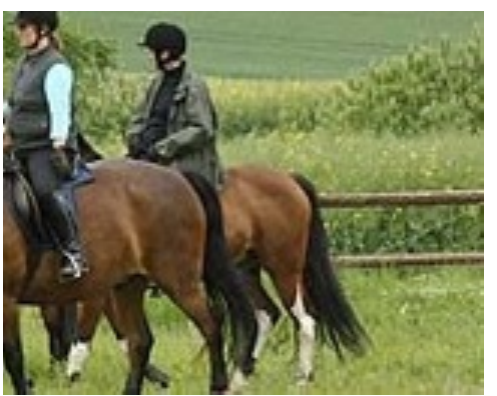}\hspace{-0.12cm}
\includegraphics[width=0.1\textwidth, height=1cm]{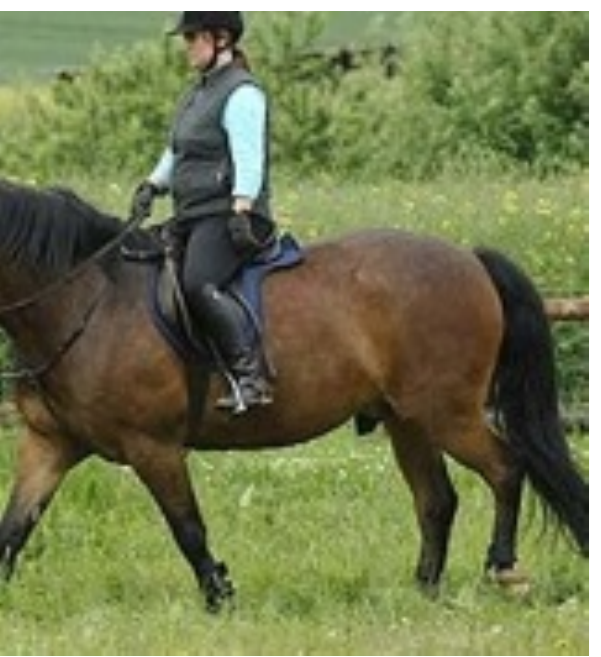}\hspace{-0.12cm}
\includegraphics[width=0.1\textwidth, height=1cm]{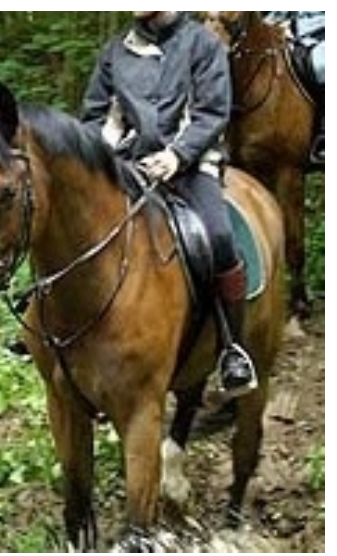}\hspace{0.2cm}
\includegraphics[width=0.1\textwidth, height=1cm]{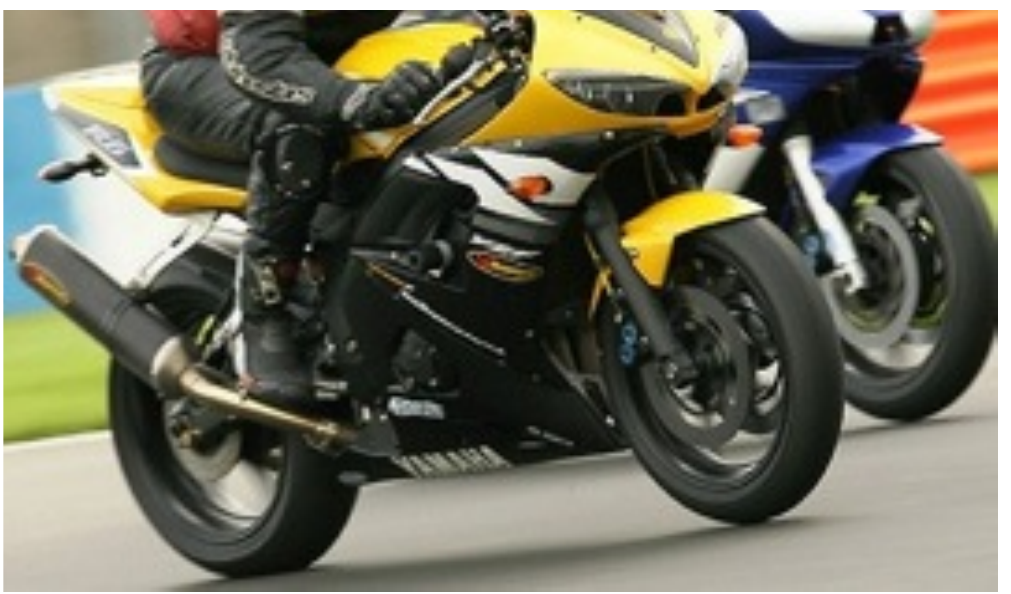}\hspace{-0.12cm}
\includegraphics[width=0.1\textwidth, height=1cm]{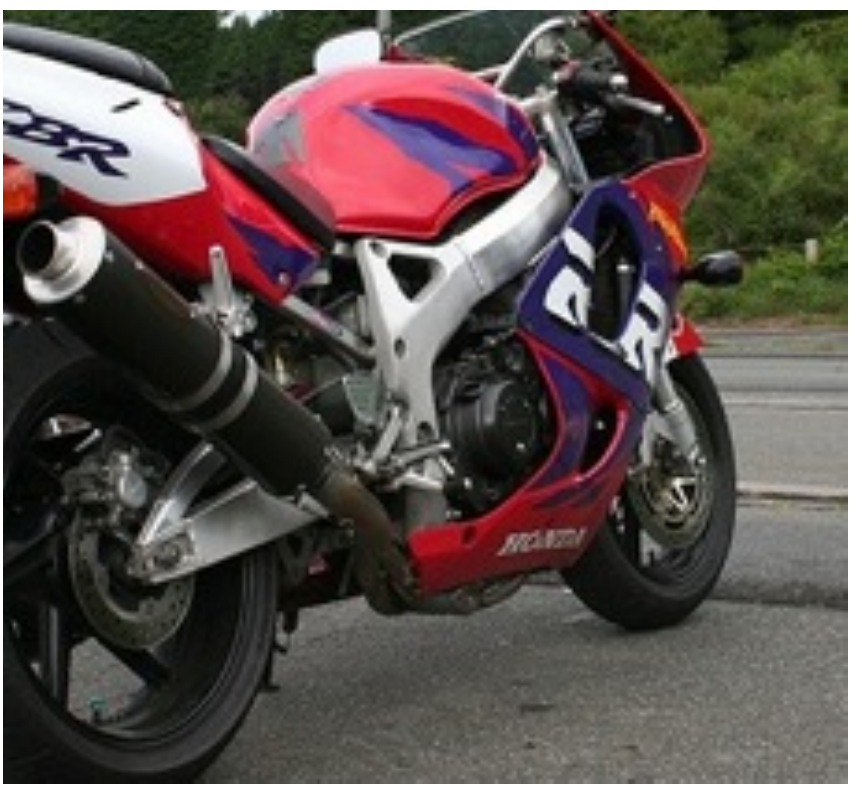}\hspace{-0.12cm}
\includegraphics[width=0.1\textwidth, height=1cm]{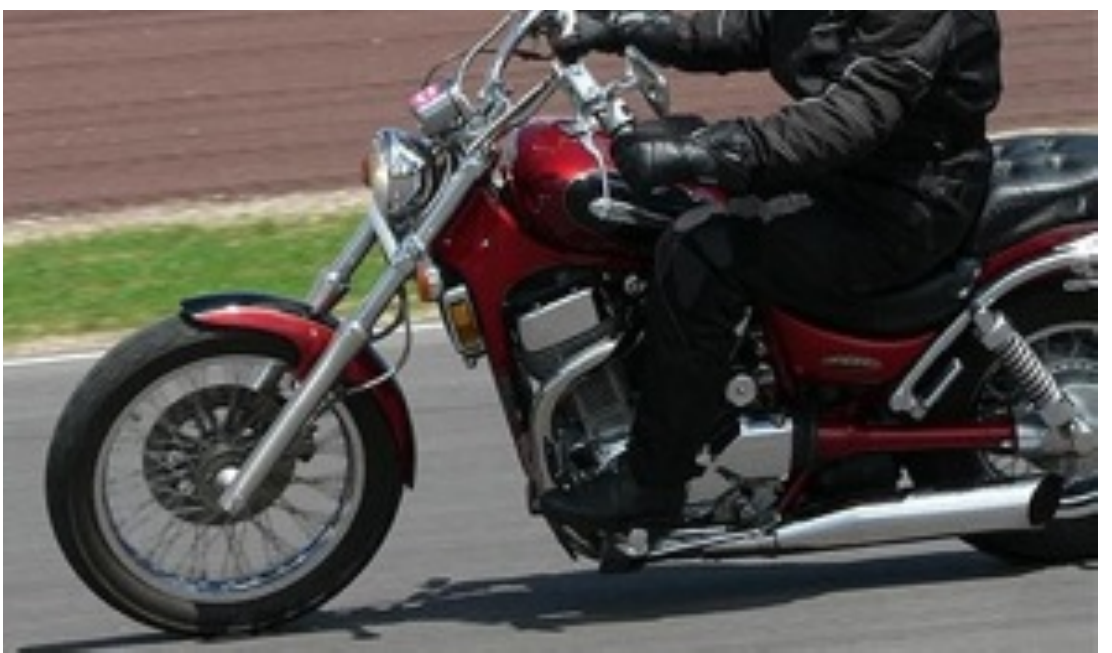}\hspace{-0.12cm}
\includegraphics[width=0.1\textwidth, height=1cm]{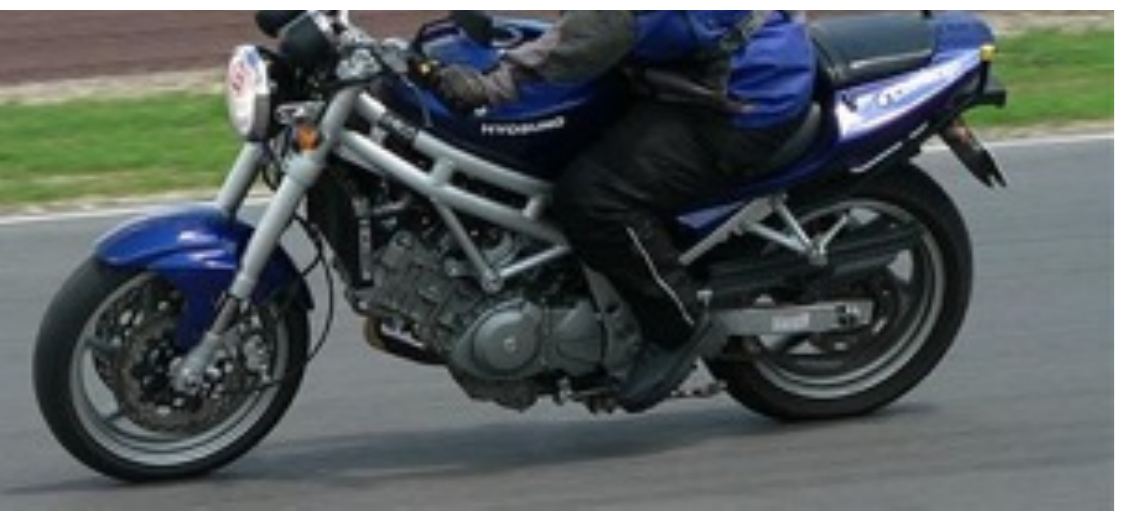}\hspace{-0.12cm}
\includegraphics[width=0.1\textwidth, height=1cm]{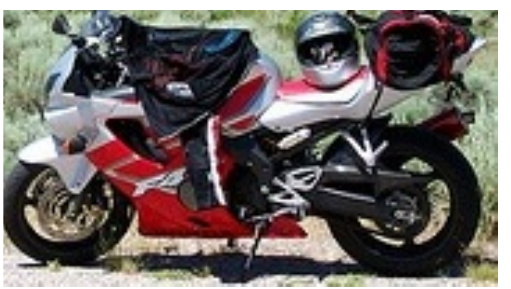}\vspace{0.2cm}\\
\includegraphics[width=0.1\textwidth, height=1cm]{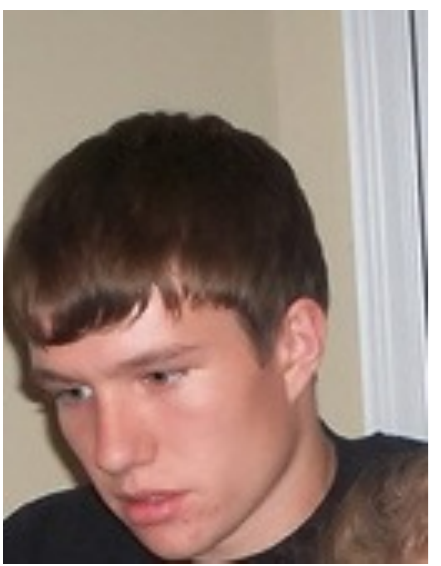}\hspace{-0.12cm}
\includegraphics[width=0.1\textwidth, height=1cm]{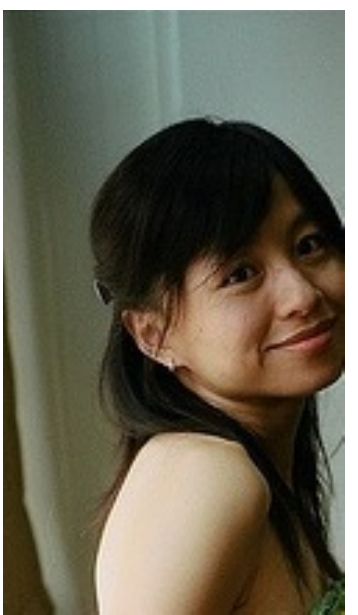}\hspace{-0.12cm}
\includegraphics[width=0.1\textwidth, height=1cm]{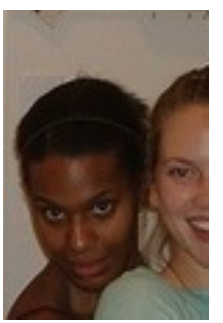}\hspace{-0.12cm}
\includegraphics[width=0.1\textwidth, height=1cm]{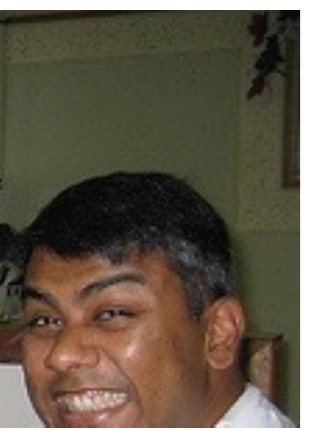}\hspace{-0.12cm}
\includegraphics[width=0.1\textwidth, height=1cm]{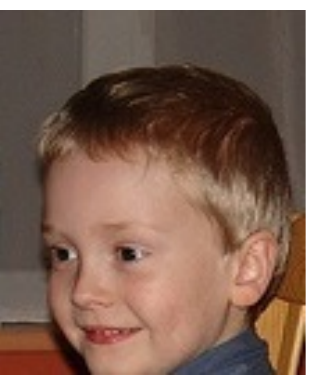}\hspace{0.2cm}
\includegraphics[width=0.1\textwidth, height=1cm]{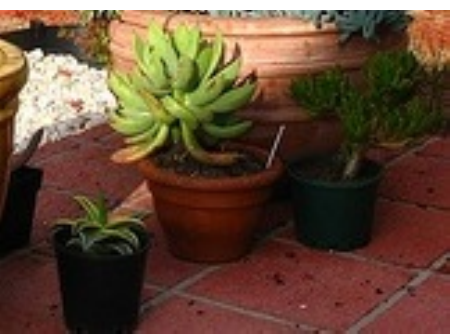}\hspace{-0.12cm}
\includegraphics[width=0.1\textwidth, height=1cm]{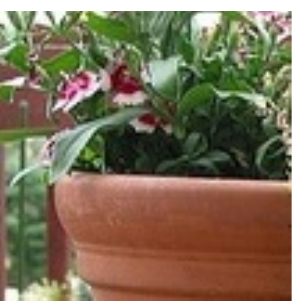}\hspace{-0.12cm}
\includegraphics[width=0.1\textwidth, height=1cm]{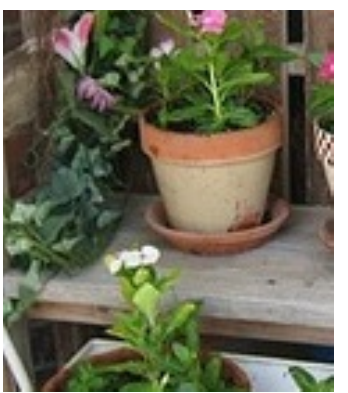}\hspace{-0.12cm}
\includegraphics[width=0.1\textwidth, height=1cm]{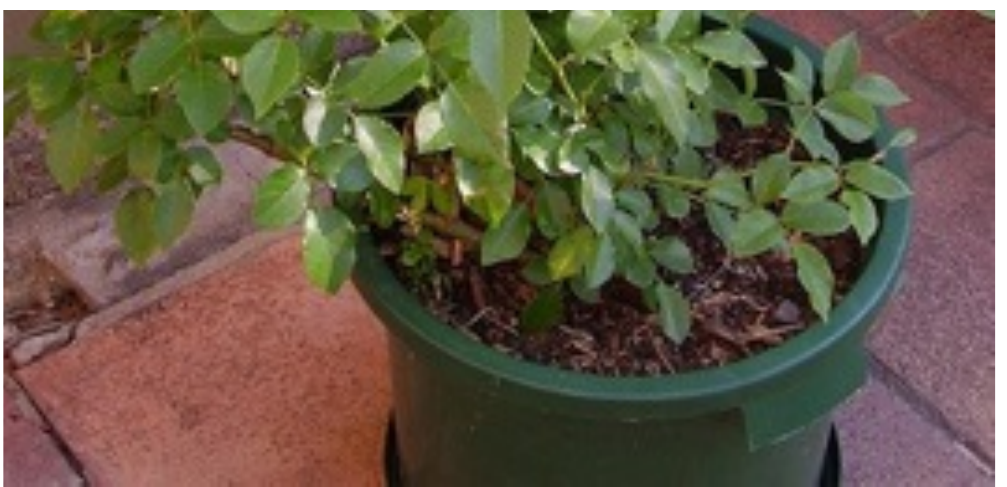}\hspace{-0.12cm}
\includegraphics[width=0.1\textwidth, height=1cm]{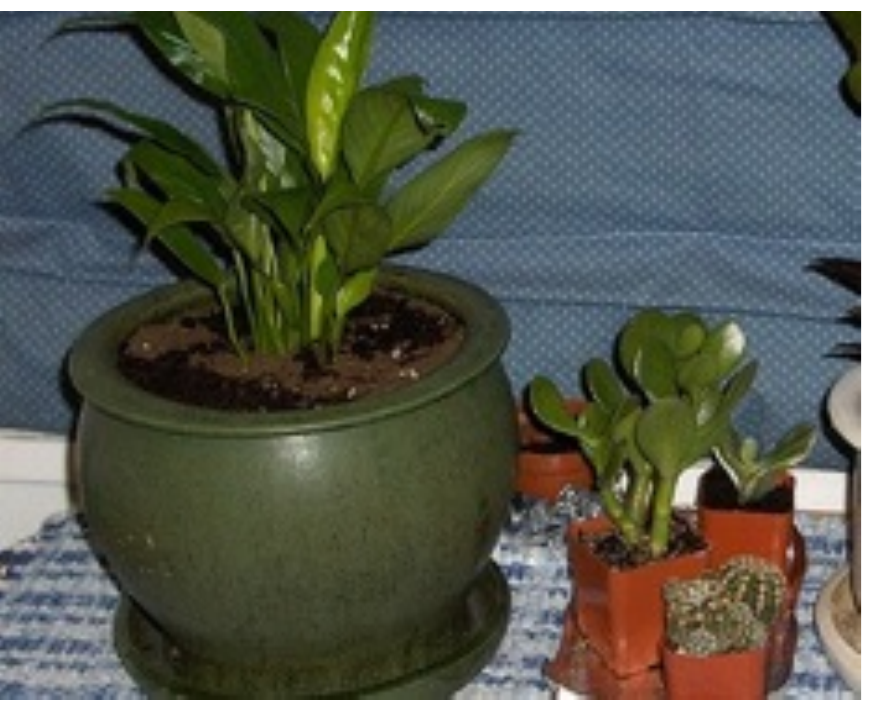}\vspace{0.2cm}\\
\includegraphics[width=0.1\textwidth, height=1cm]{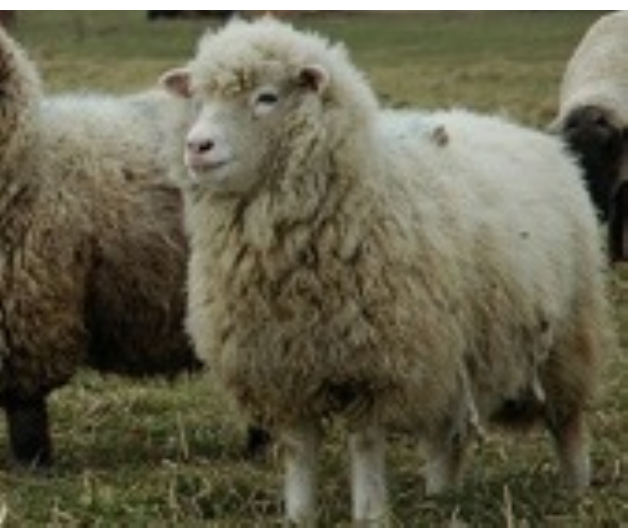}\hspace{-0.12cm}
\includegraphics[width=0.1\textwidth, height=1cm]{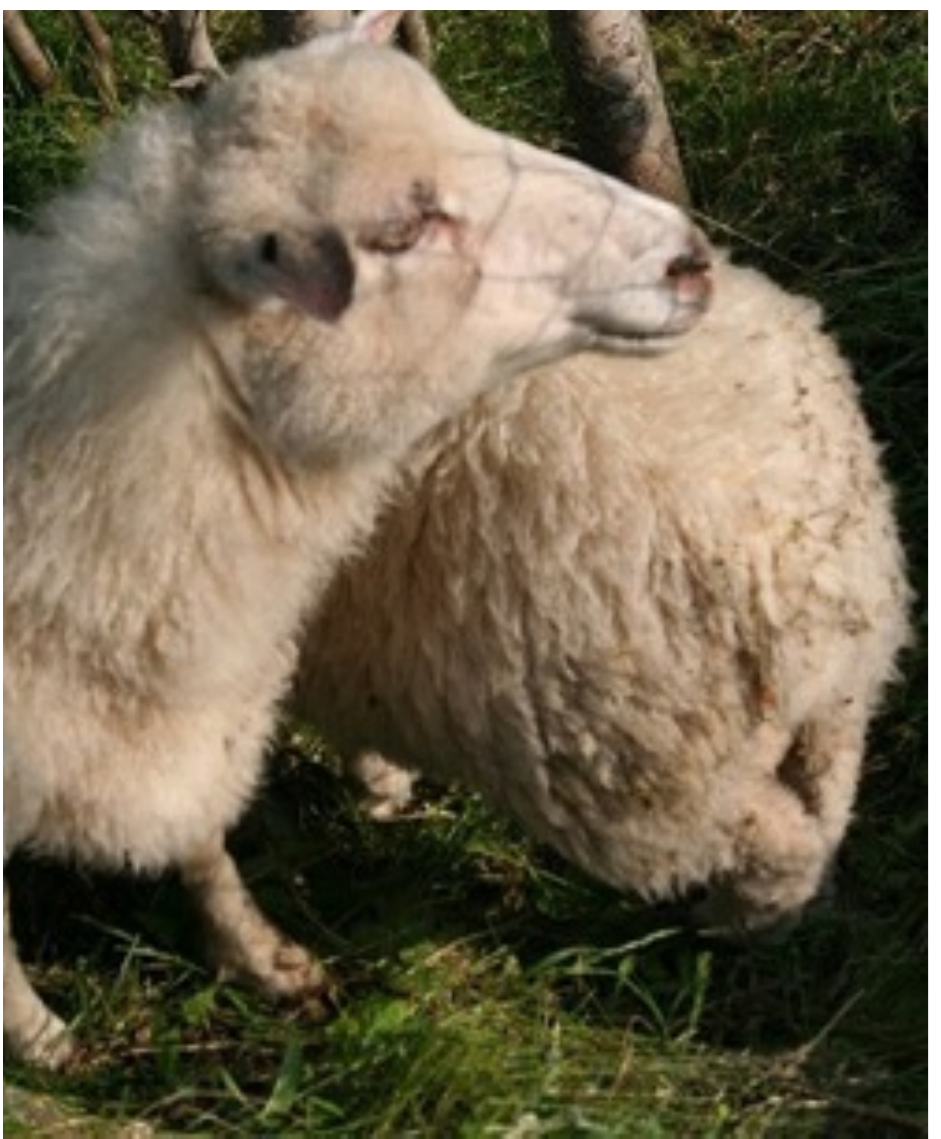}\hspace{-0.12cm}
\includegraphics[width=0.1\textwidth, height=1cm]{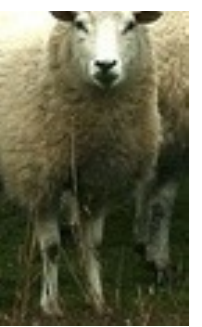}\hspace{-0.12cm}
\includegraphics[width=0.1\textwidth, height=1cm]{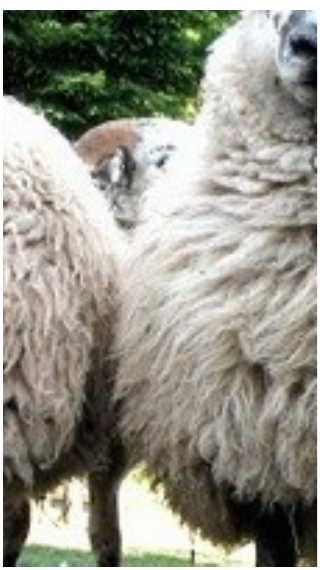}\hspace{-0.12cm}
\includegraphics[width=0.1\textwidth, height=1cm]{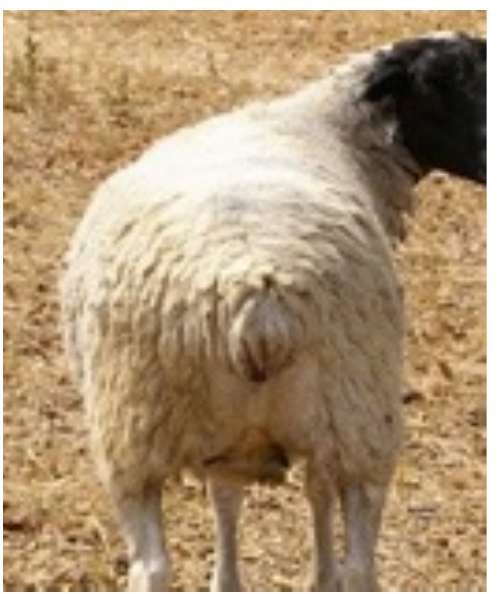}\hspace{0.2cm}
\includegraphics[width=0.1\textwidth, height=1cm]{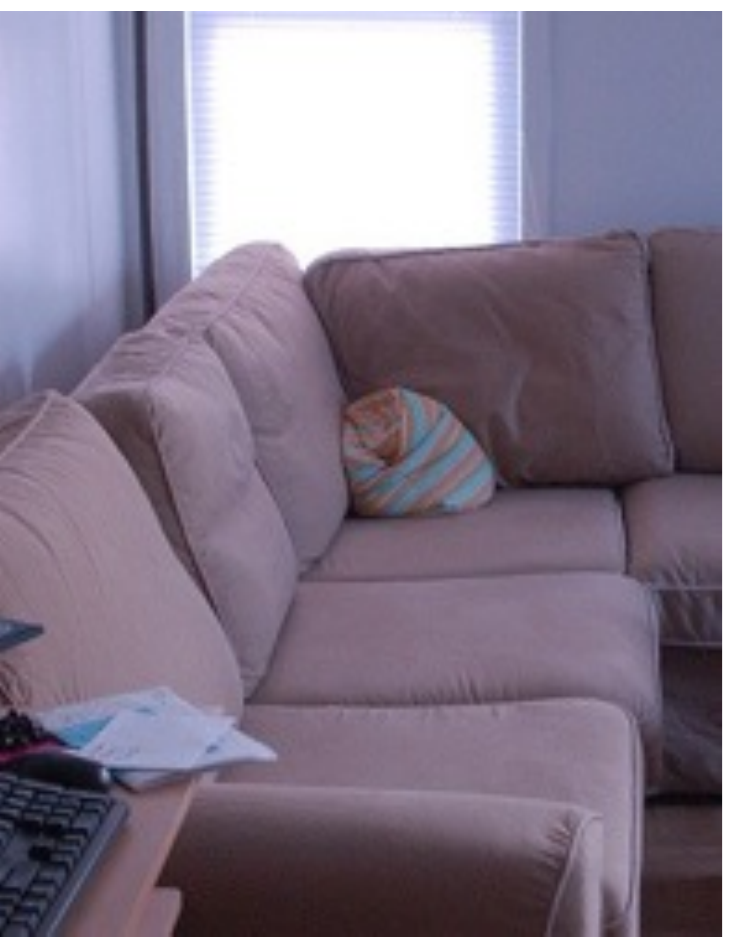}\hspace{-0.12cm}
\includegraphics[width=0.1\textwidth, height=1cm]{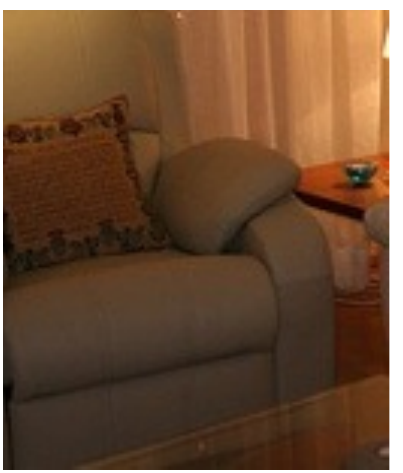}\hspace{-0.12cm}
\includegraphics[width=0.1\textwidth, height=1cm]{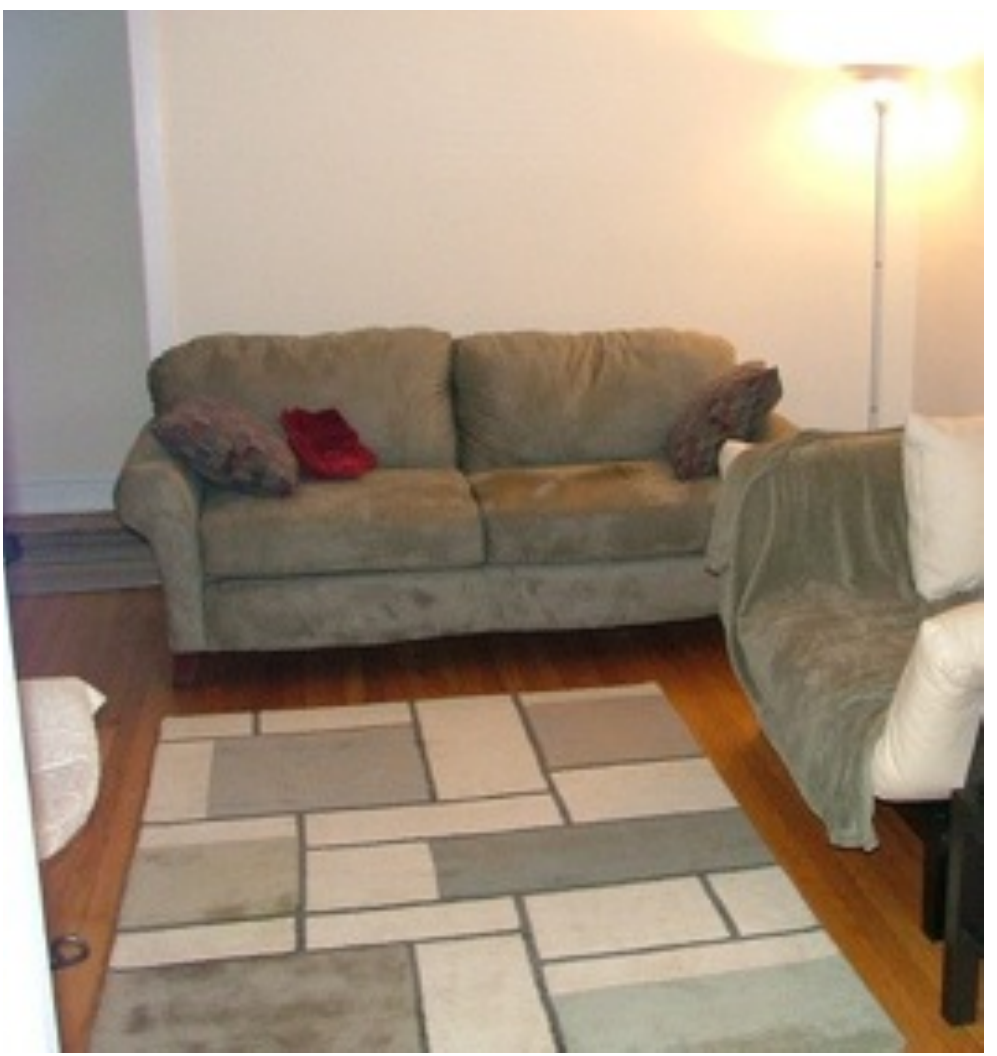}\hspace{-0.12cm}
\includegraphics[width=0.1\textwidth, height=1cm]{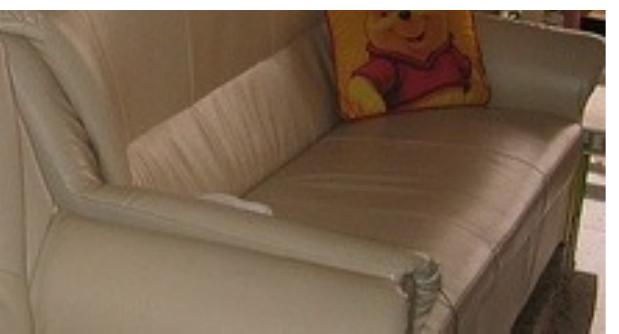}\hspace{-0.12cm}
\includegraphics[width=0.1\textwidth, height=1cm]{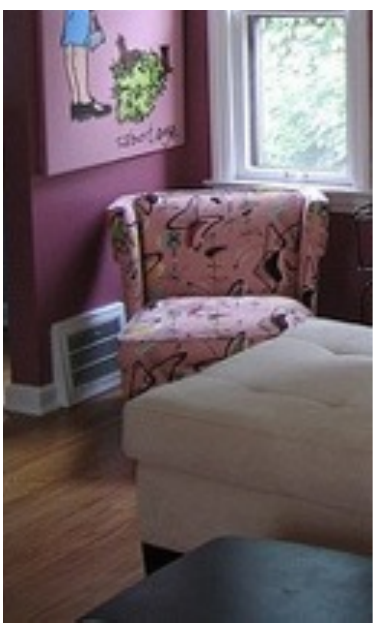}\vspace{0.2cm}\\
\includegraphics[width=0.1\textwidth, height=1cm]{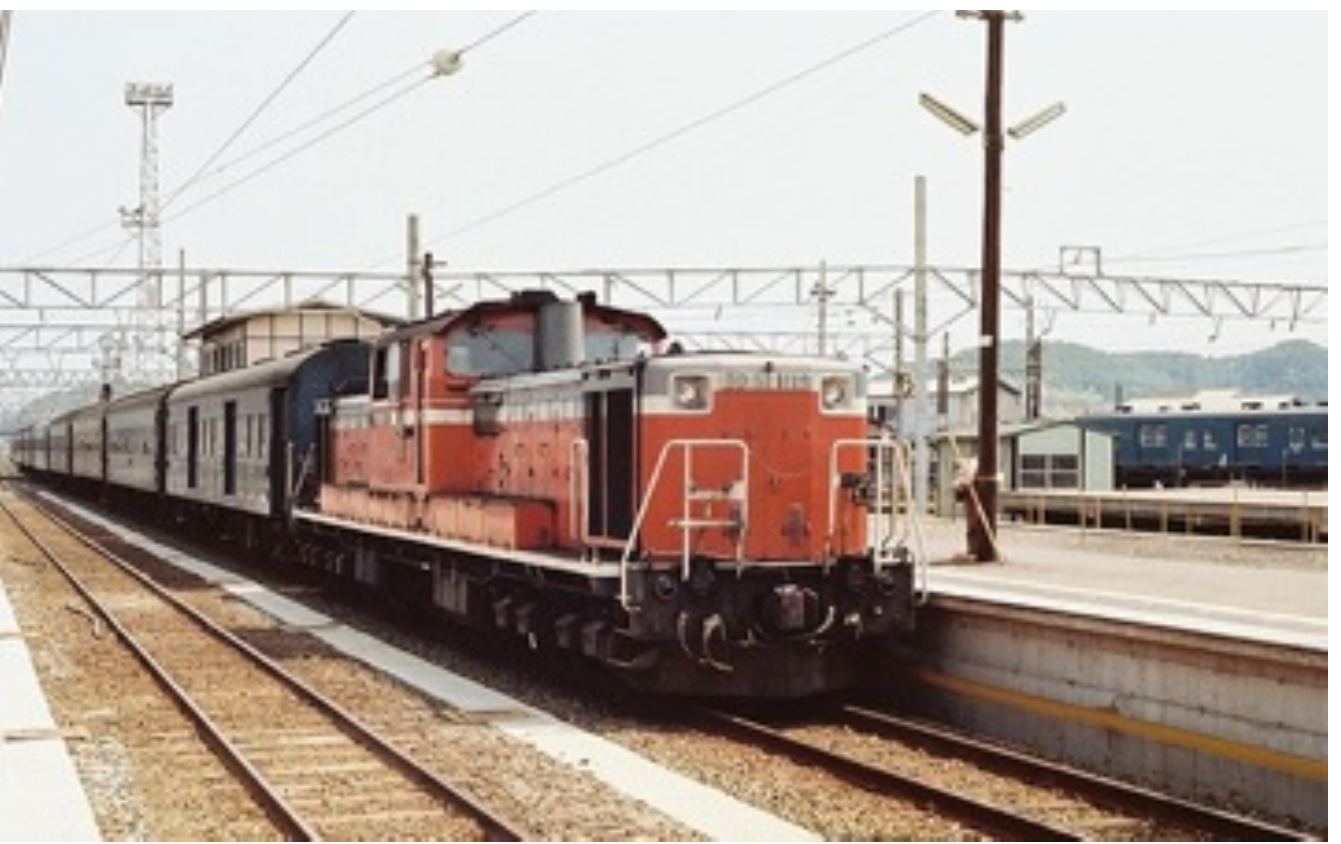}\hspace{-0.12cm}
\includegraphics[width=0.1\textwidth, height=1cm]{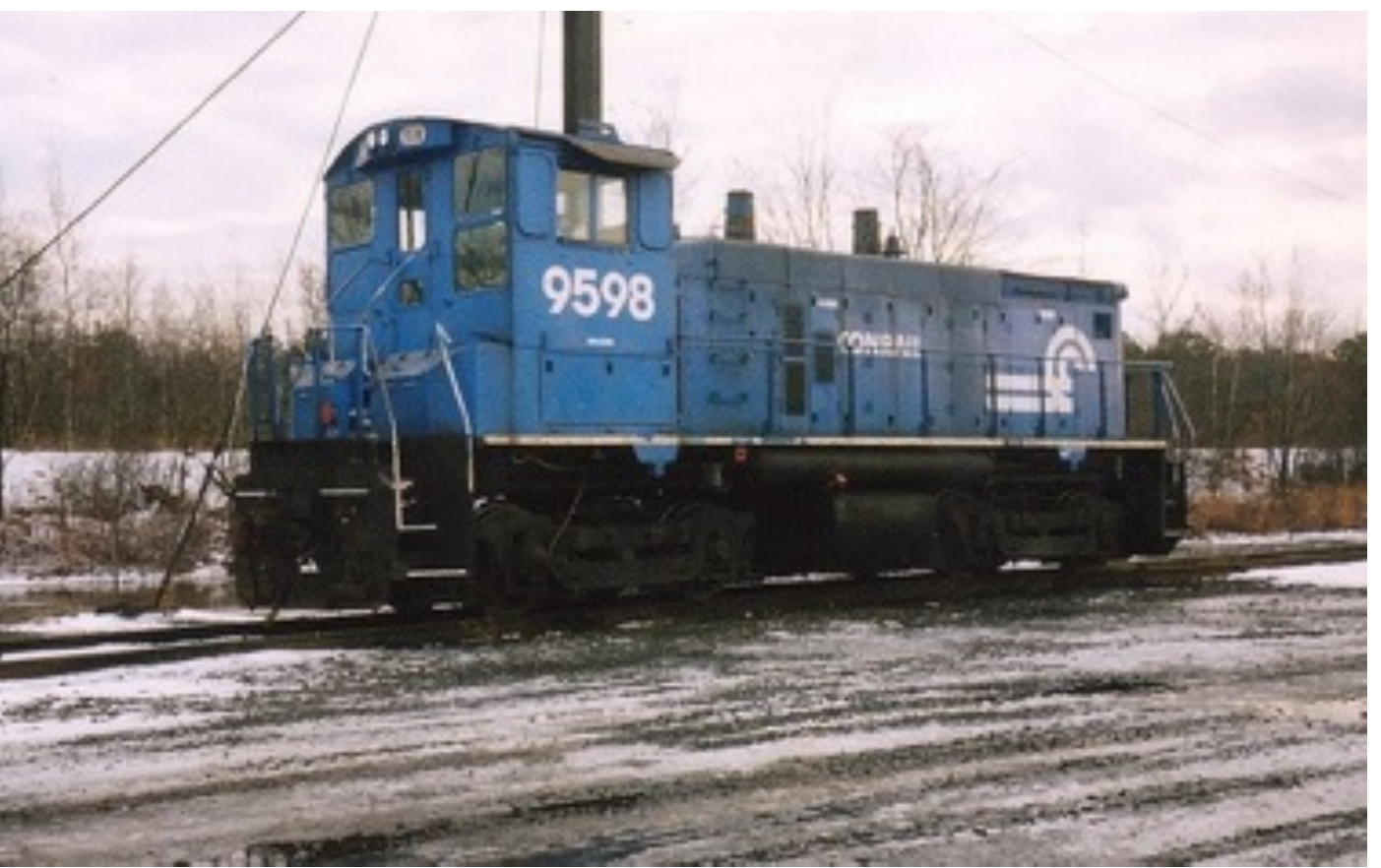}\hspace{-0.12cm}
\includegraphics[width=0.1\textwidth, height=1cm]{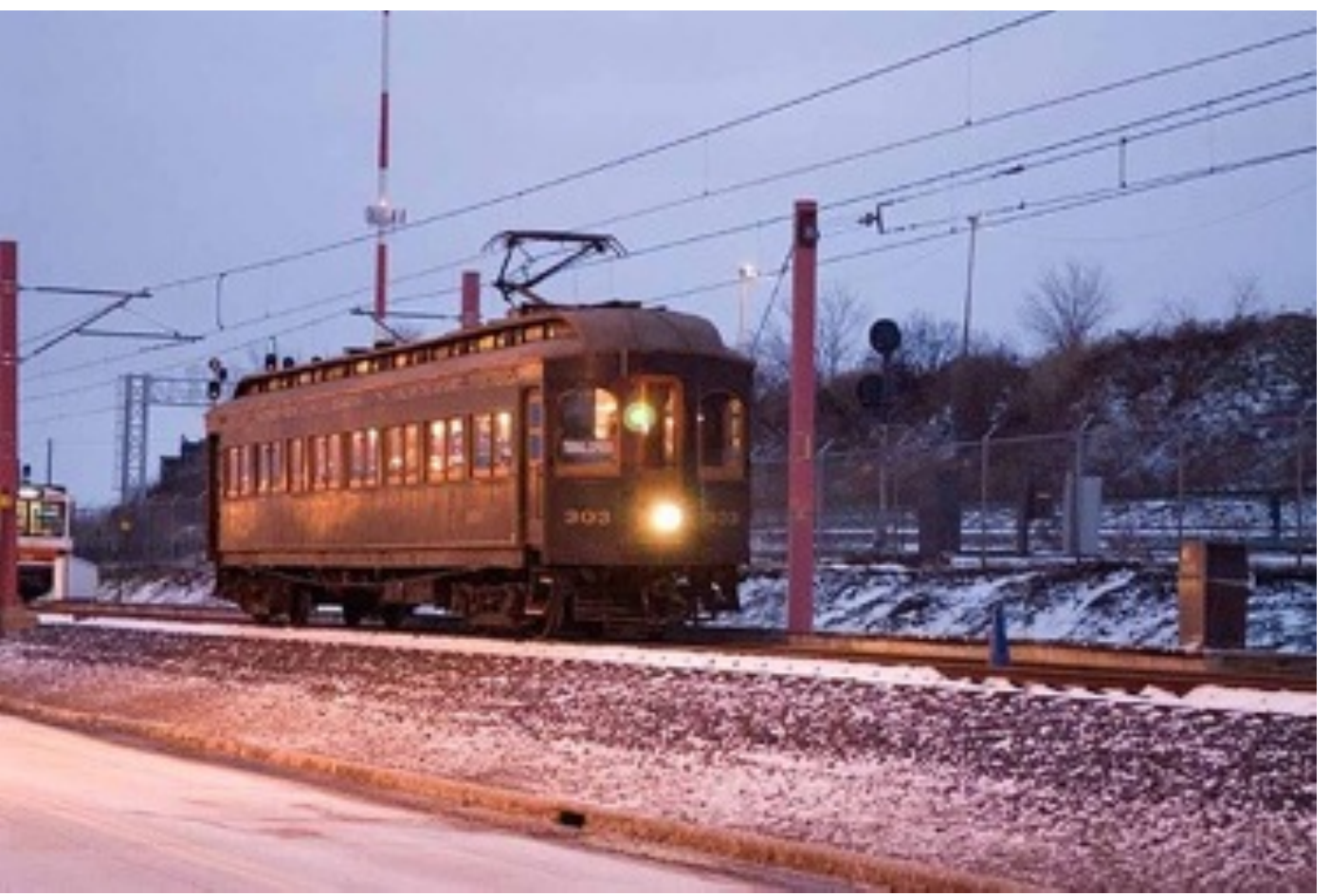}\hspace{-0.12cm}
\includegraphics[width=0.1\textwidth, height=1cm]{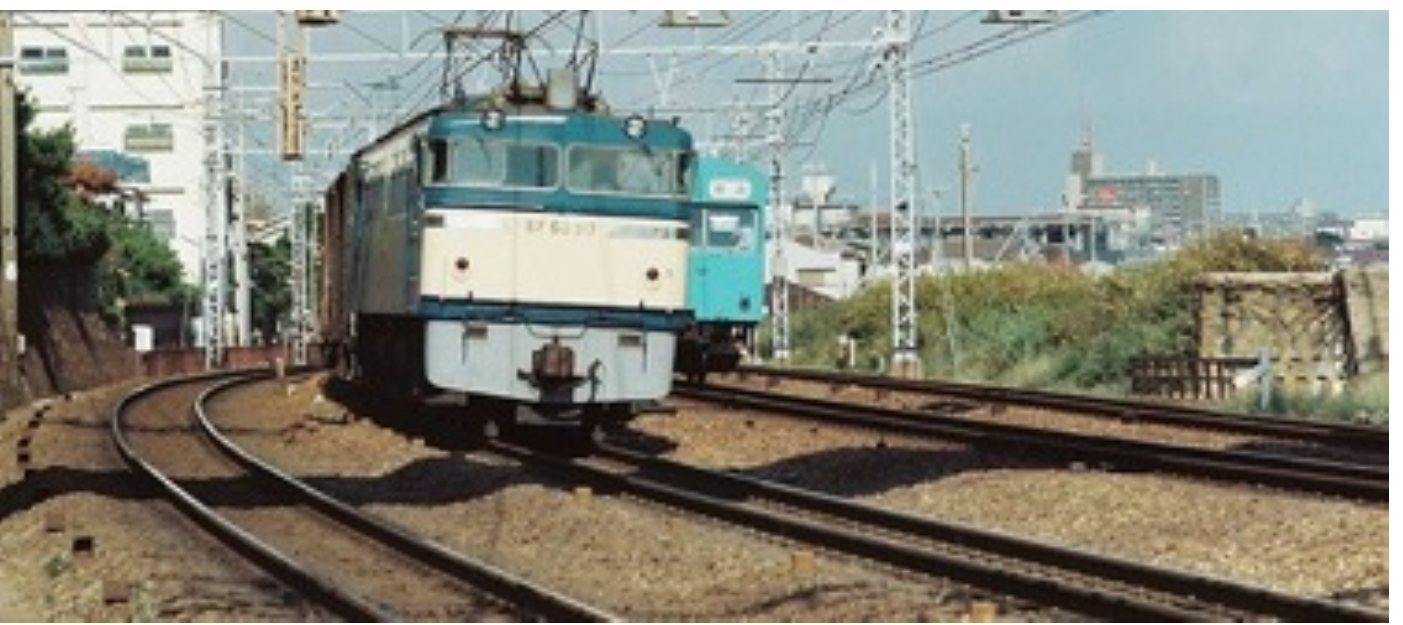}\hspace{-0.12cm}
\includegraphics[width=0.1\textwidth, height=1cm]{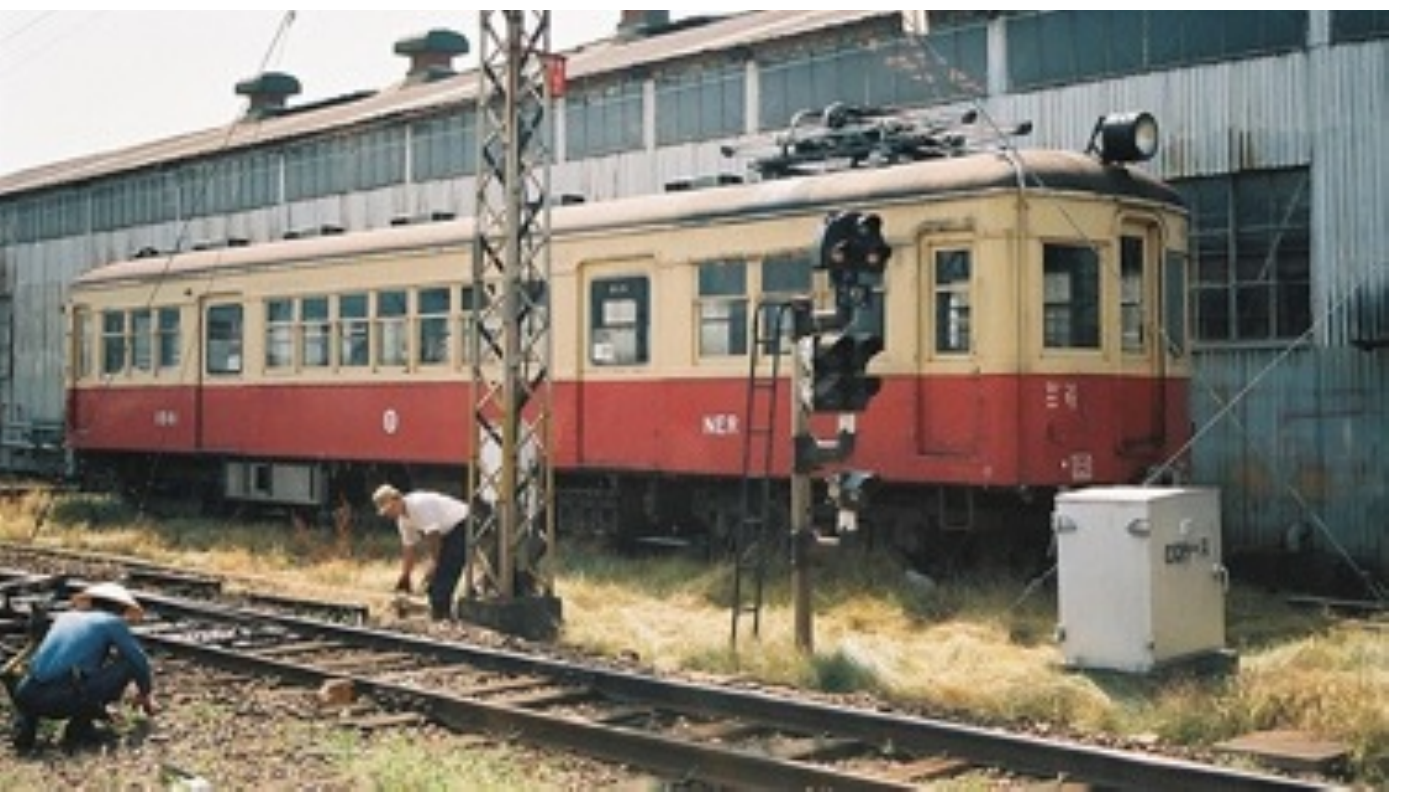}\hspace{0.2cm}
\includegraphics[width=0.1\textwidth, height=1cm]{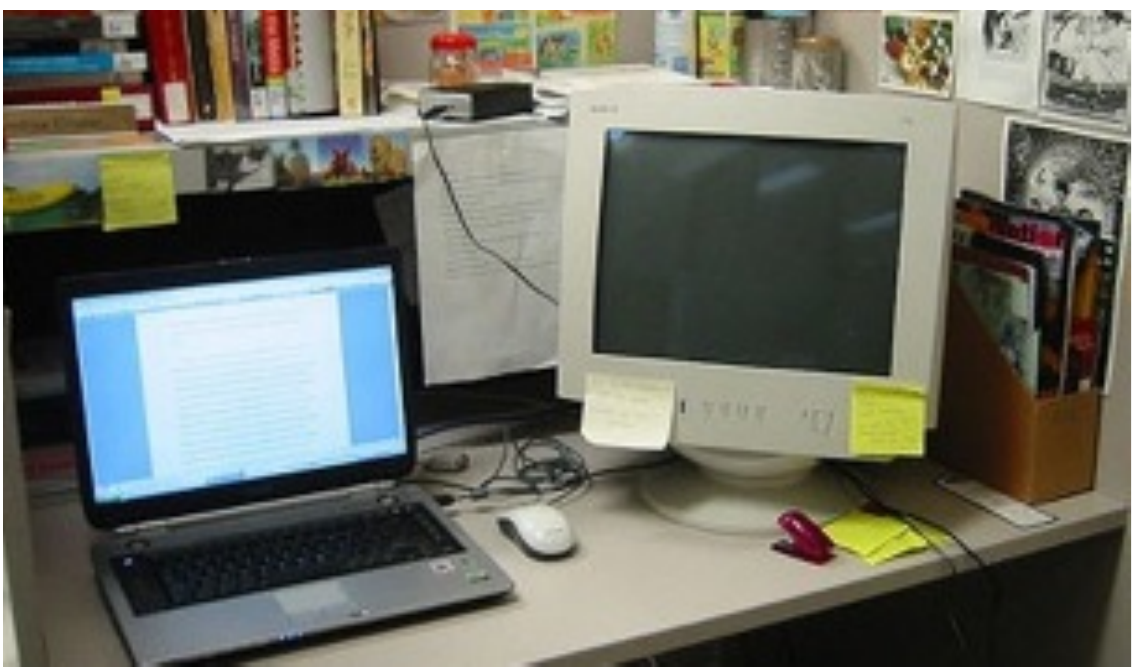}\hspace{-0.12cm}
\includegraphics[width=0.1\textwidth, height=1cm]{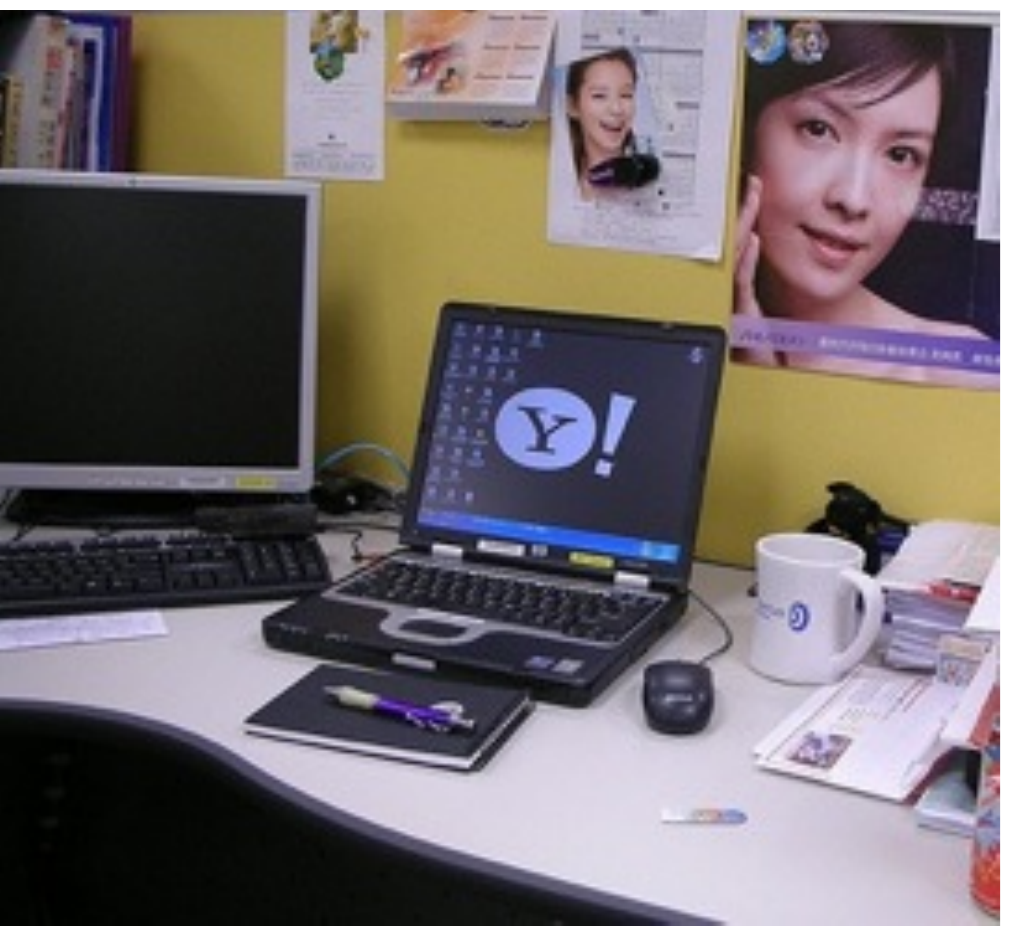}\hspace{-0.12cm}
\includegraphics[width=0.1\textwidth, height=1cm]{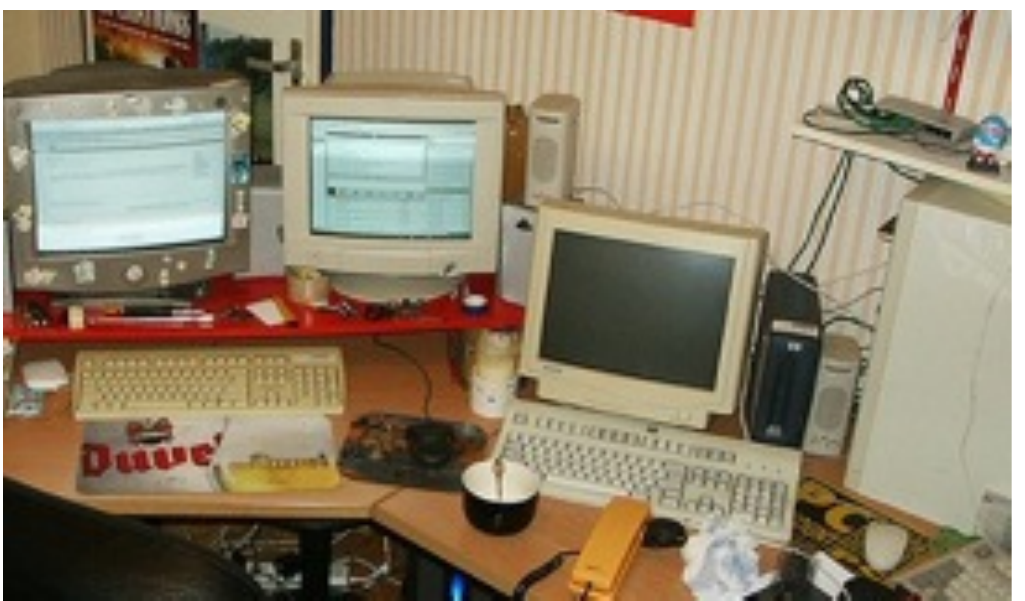}\hspace{-0.12cm}
\includegraphics[width=0.1\textwidth, height=1cm]{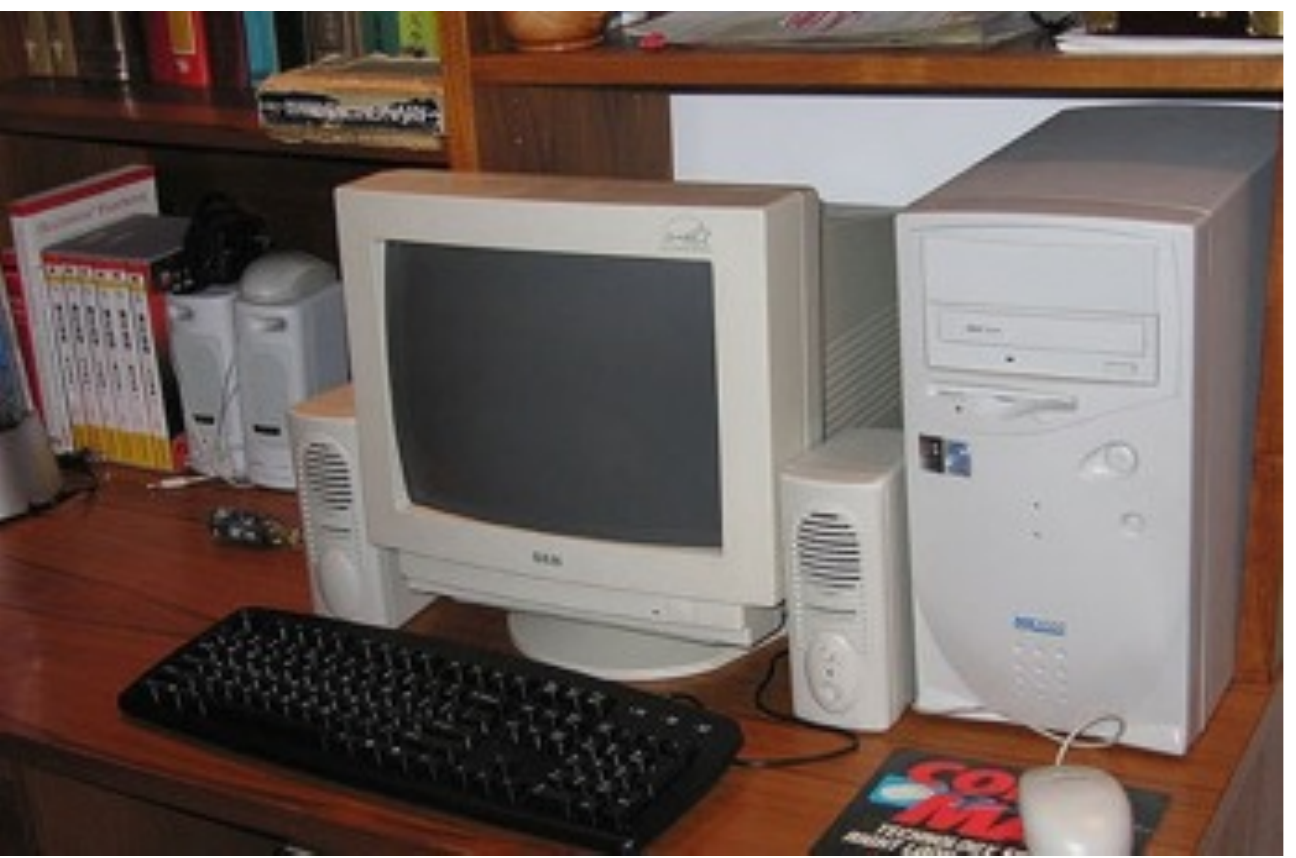}\hspace{-0.12cm}
\includegraphics[width=0.1\textwidth, height=1cm]{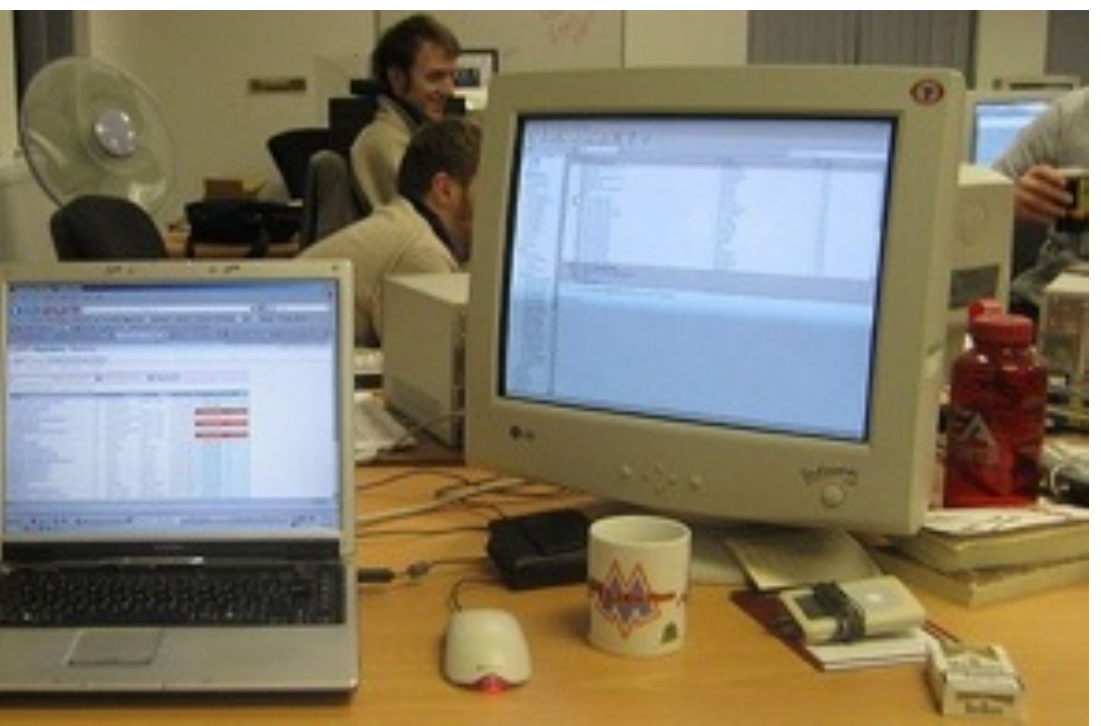}\\
\caption{Visualizations of top $5$ nearest neighbor proposal boxes with positive labels in the first cluster, $S_1$ for all $20$ classes in PASCAL VOC dataset. From left to right, aeroplane, bicycle, bird, boat, bottle, bus, car, cat, chair, cow, diningtable, dog, horse, motorbike, person, plant, sheep, sofa, train, and tvmonitor.} 
\label{fig:cluster_visualization_figure} 
\end{figure*}

In Figure \ref{fig:cluster_visualization_figure}, we visualize top $5$ nearest neighbors with positive labels in the first chosen cluster $S_1$ for all $20$ classes on the PASCAL VOC data.
Our experiments in Section~\ref{sec:exp} show the benefits of our framework. Potentially, the results might improve even further when using the complementary mode shifts of \citep{discovery3} as a pre-selection step before covering.

\section{Iterative refinement with latent variables}
\label{sec:slsvm}

\begin{figure*}[htbp]
\centering
\includegraphics[width=0.35\textwidth]{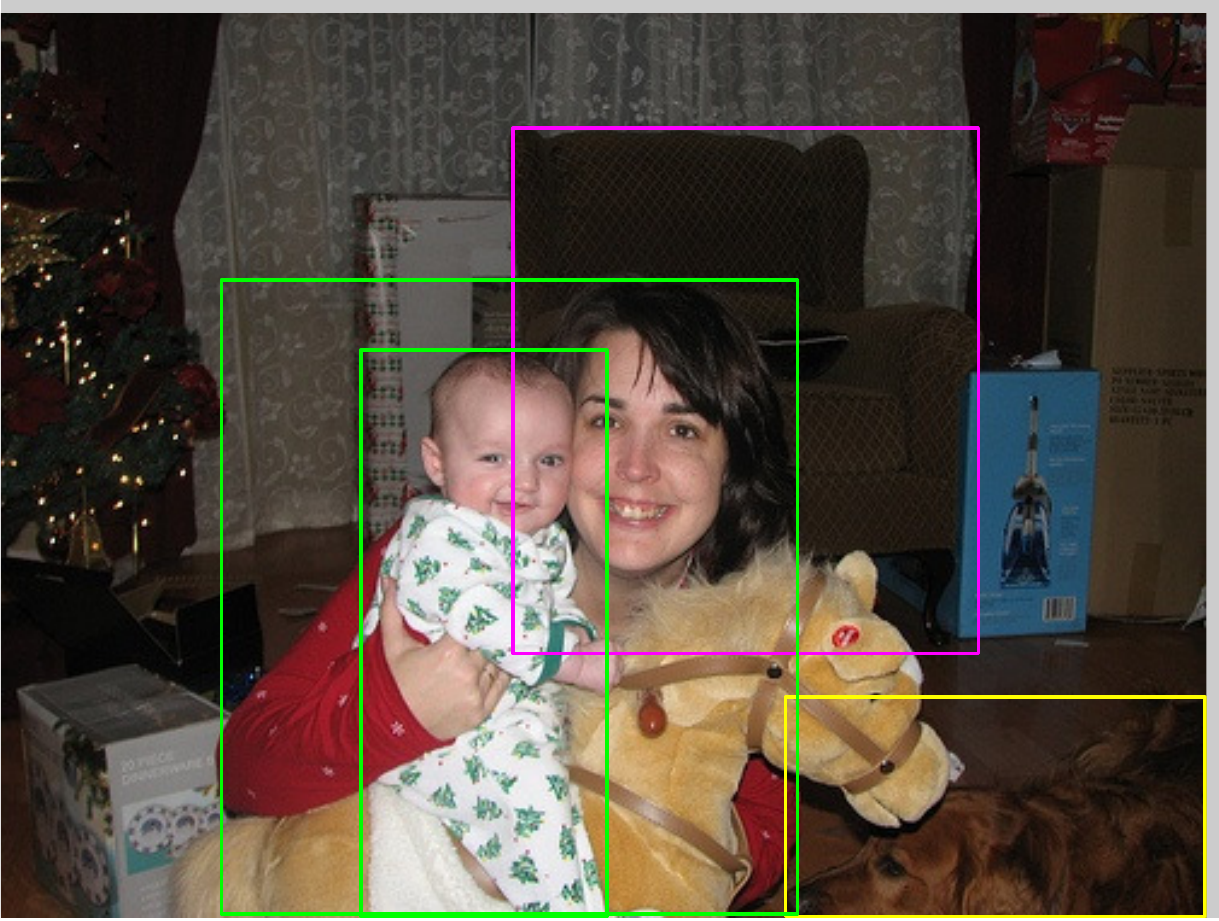} \hspace{0.1cm}
\includegraphics[width=0.35\textwidth]{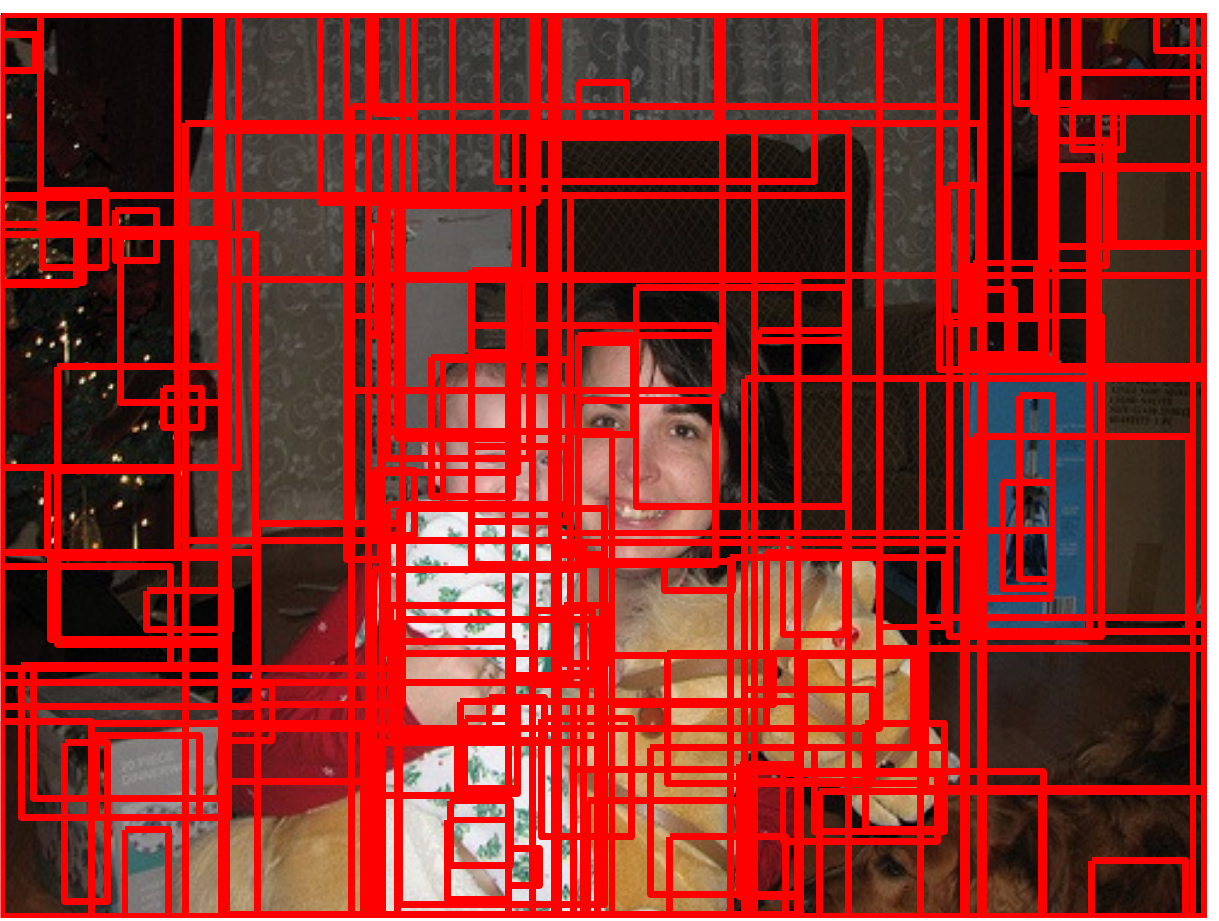}
\caption{In the refinement stage, we formulate a multiple instance learning bag per image and bag instances correspond to each window proposals from selective search.  Binary bag labels correspond to image-level annotations of whether the target object exists in the image or not. (Left) ground truth bounding boxes color coded with category labels. green: person, yellow: dog, and magenta: sofa, (Right) visualization of $100$ random subset of window proposals.} 
\label{fig:mil-figure} 
\end{figure*}

In this section, we review the latent SVM formulation, and we propose a simple
smoothing technique enabling us to use classical techniques for unconstrained
smooth optimization. Figure \ref{fig:mil-figure} illustrates our multiple
instance learning analogy for object detection with one-bit labels.

\subsection{Review of latent SVM}

For a binary classification problem, the latent SVM formulation consists of
learning a decision function involving a maximization step over a discrete set of 
configurations~$\mathcal Z$. Given a data point~$\mathbf{x}$ in~$\mathbb{R}^p$ that we
want to classify, and some learned model parameters~$\w$ in~$\Real^d$, 
we select a label~$y$ in~$\{-1,+1\}$ as follows:
\begin{equation}
   y = \sign\left(  \max_{\mathbf{z} \in \mathcal{Z}} \mathbf{w}^\intercal \phi( \x,\z) \right), \label{eq:decision}
\end{equation}
where $\z$ is called a ``latent variable'' chosen among the set~$\Z$. For
object detection, $\Z$ is typically a set of bounding boxes, and maximizing
over $\Z$ amounts to finding a bounding box containing the object.  In
deformable part models~\cite{lsvm-pami}, the set $\Z$ contains all possible
part configurations, each part being associated to a position in the image.
The resulting set~$\Z$ has exponential size, but~(\ref{eq:decision}) can be
solved efficiently with dynamic programming techniques for particular choices of~$\phi$.

Learning the model parameters~$\w$ is more involved than solving a simple SVM
problem. We are given some training data $\{(\mathbf{x}_i, y_i)\}_{i=1}^n$, where the 
vectors $\x_i$ are  in~$\Real^p$ and the scalars $y_i$ are binary labels in~$\{1,
-1\}$. Then, the latent SVM formulation becomes
\vspace{-4pt}
\begin{equation}
   \min_{\w \in \Real^d} \frac{1}{2} \|\mathbf{w}\|_2^2 + C \sum_{i = 1}^n \ell
   \left(y_i,\;  \max_{\z \in \mathcal{Z}} \mathbf{w}^\intercal
   \phi(\mathbf{x}_i, \z) \right), \label{eq:latentsvm}
 \vspace{-4pt}
\end{equation}
where $\ell : \Real \times \Real \to \Real$ is the hinge loss defined as
$\ell(y, \hat{y} ) = \max(0,1- y \hat{y})$, which encourages the decision
function for each training example to be the same as the corresponding label.
Similarly, other loss functions can be used such as the logistic or
squared hinge loss.

Problem~(\ref{eq:latentsvm}) is nonconvex and nonsmooth, making it hard to
tackle. A classical technique to obtain an approximate solution is to
use a difference of convex (DC) programming technique, called concave-convex
procedure ~\cite{yuille_cccp,Yu09}.  
We remark that the part of~(\ref{eq:latentsvm}) corresponding to negative
examples is convex with respect to~$\w$. It is indeed easy to show that each corresponding term can
be written as a pointwise maximum of convex functions, and is thus
convex~\citep[see][]{boyd}: when $y_i=-1$, $\ell \left(y_i,  \max_{\z \in \mathcal{Z}}
\mathbf{w}^\intercal \phi(\mathbf{x}_i, \z) \right) = \max_{\z \in \Z} \ell(y_i,
\w^\intercal \phi(\x_i,\x))$.  On the other hand, the part corresponding to positive
examples is concave, making the objective~(\ref{eq:latentsvm}) suitable
to DC programming. Even though such a procedure does not have any
theoretical guarantee about the quality of the optimization, it monotonically
decreases the value of the objective and performs relatively well when the
problem is well initialized~\cite{lsvm-pami}.

We propose a \emph{smooth formulation} of latent SVM, with
 two main motives. First, smoothing the objective function of
latent SVM allows the use of efficient second-order optimization algorithms such as
quasi-Newton \cite{lbfgs} that can leverage curvature information to speed up convergence.
Second, as we show later, smoothing the latent SVM boils down to considering the
top-$N$ configurations in the maximization step in place of the top-$1$
configuration in the regular latent SVM. As a result, the smooth latent SVM training
becomes more robust to unreliable configurations in the early stages, since a
larger set of plausible configurations is considered at each maximization step. 

\subsection{Smooth formulation of LSVM}





In the objective~(\ref{eq:latentsvm}), the hinge loss can be easily replaced by
a smooth alternative, e.g., squared hinge, or logistic loss. However, the
non-smooth points induced by the following functions are more difficult to handle
\begin{equation}
   f_{\x_i}(\w) := \max_{\z \in \mathcal{Z}}\mathbf{w}^\intercal \phi(\mathbf{x}_i, \z).\label{eq:fmax}
\end{equation}
We propose to use a smoothing technique studied by~~\citet{nesterov} for convex functions.
\paragraph{Nesterov's smoothing technique}
We only recall here the
simpler form of Nesterov's results that is relevant for our purpose. 
Consider a non-smooth function that can be written in the following form:
\begin{align}
   g(\w) :=  \max_{\mathbf{u} \in \Delta} \: \langle \mathbf{A}\w, \mathbf{u} \rangle\, ,
\end{align}
where $\mathbf{u} \in \mathbb{R}^m$,  $\mathbf{A}$ is in~$\Real^{m \times d}$, and $\Delta$ denotes the probability simplex, $\Delta = \{\mathbf{x}: \sum_{i=1}^m x_i = 1, x_i \geq 0\}$. Smoothing here consists of 
adding a strongly convex function $\omega$ in the maximization problem
\begin{align}
   g_{\mu}(\w) := \max_{\mathbf{u} \in \Delta} \left[ \: \langle \mathbf{A}\w, \mathbf{u} \rangle - \frac{\mu}{2} \omega(\mathbf{u}) \right] \; . \label{eq:smoothedf}
\end{align}
The resulting function $g_\mu$ is differentiable for all $\mu > 0$, and its gradient is
\begin{equation}
   \nabla g_\mu(\w) = \A^\intercal \uu^\star(\w),
\end{equation}
where $\uu^\star(\w)$ is the unique solution of~(\ref{eq:smoothedf}).
The parameter $\mu$ controls the amount of smoothing. Clearly, $g_{\mu}(\w) \to g(\w)$ for all $\w \in \mathcal{W}$
as $\mu \to 0$. As~\citet{nesterov} shows, for a given target approximation accuracy~$\epsilon$, 
there is an optimal amount of smoothing $\mu(\epsilon)$ that can be derived from a convex optimization perspective using the strong convexity parameter of $\omega(\cdot)$ on $\Delta$ and the (usually unknown) Lipschitz constant of $g$.
In the experiments, we shall simply learn
the parameter $\mu$ from data. 


\paragraph{Smoothing the latent SVM}
We now apply Nesterov's smoothing technique to the latent SVM objective
function. As we shall see, the smoothed objective takes a simple form, which can
be efficiently computed in the latent SVM framework.  Furthermore, smoothing
latent SVM implicitly models uncertainty in the selection of the best
configuration~$\z$ in~$\Z$, as shown by \citet{KumarPK12} for 
a different smoothing scheme. 

In order to smooth the functions $f_{\x_i}$ defined in~(\ref{eq:fmax}), 
we first notice that
\begin{equation}
   f_{\x_i}(\w) = \max_{{\mathbf u} \in \Delta} \: \langle \A_{\x_i} \w , {\mathbf u} \rangle,\label{eq:simplex}
\end{equation}
where $\A_{\x_i}$ is a matrix of size $|\Z| \times d$ such that the $j$-th row of~$\A_{\x_i}$ is
the feature vector $\phi(\x_i,\z_j)$ and $\z_j$ is the $j$-th element of $\Z$.
Considering any strongly convex function $\omega$ and parameter $\mu > 0$, 
the smoothed latent SVM objective is obtained by replacing in~(\ref{eq:latentsvm}) \\
~$\bullet$ the functions $f_{\x_i}$ by their smoothed counterparts $f_{\x_i,\mu}$ obtained by applying~(\ref{eq:smoothedf}) to~(\ref{eq:simplex}); \\
~$\bullet$ the non-smooth hinge-loss function $\l$ by any smooth loss.

\paragraph{Objective and gradient evaluations}
An important issue remains the computational tractability of the new
formulation in terms of objective and gradient evaluations, in order to
use quasi-Newton optimization techniques. The choice of the strongly convex
function $\omega$ is crucial in this respect. 

There are two functions known to be strongly convex on the simplex: i) the
Euclidean norm, ii) the entropy.  In the case of the Euclidean-norm
$\omega(\mathbf{\uu})= \Vert \mathbf{u} \Vert_2^2$, it turns out that the
smoothed counterpart can be efficiently computed using a projection on
the simplex, as shown below.
\begin{equation}
   \uu^\star(\w) = \argmin_{\uu \in \Delta} \left\|\frac{1}{\mu} \A \w   - \uu\right\|_2^2, \label{eq:proj}
\end{equation}
where $\uu^\star(\w)$ is the solution of~(\ref{eq:smoothedf}).  Computing
$\A\w$ requires a priori $O(|\Z|d)$ operations.  The projection can be computed
in $O(|\Z|)$~\citep[see, e.g.,][]{MAL-015}.  Once~$\uu^\star$ is obtained,
computing the gradient requires $O(d \|\uu^\star\|_0)$ operations, where
$\|\uu^\star\|_0$ is the number of non-zero entries in $\uu^\star$.

When the set $\Z$ is large, these complexities can be improved by leveraging
two properties. First, the projection on the simplex is known to produce sparse
solutions, the smoothing parameter $\mu$ controlling the sparsity of
$\uu^\star$; second, the projection preserves the order of the variables. As a
result, the following heuristic can be justified. Assume that for some $N <
|\Z|$, we can obtain the top-N entries of $\A \w$ without exhaustively
exploring~$\Z$. Then, performing the projection on these reduced set of $N$
variables yields a vector $\uu'$ which can be shown to be optimal for the
original problem~(\ref{eq:proj}) whenever $\|\uu'\|_0 < N$.  In other words,
whenever $N$ is large enough and $\mu$ small enough, computing the gradient of
$f_{\x_i,\mu}$ can be done in $O(N d)$ operations.  We use this heuristic in
all our experiments.

\begin{figure*}[htbp]
\centering
\includegraphics[width=0.19\textwidth, height=2.2cm]{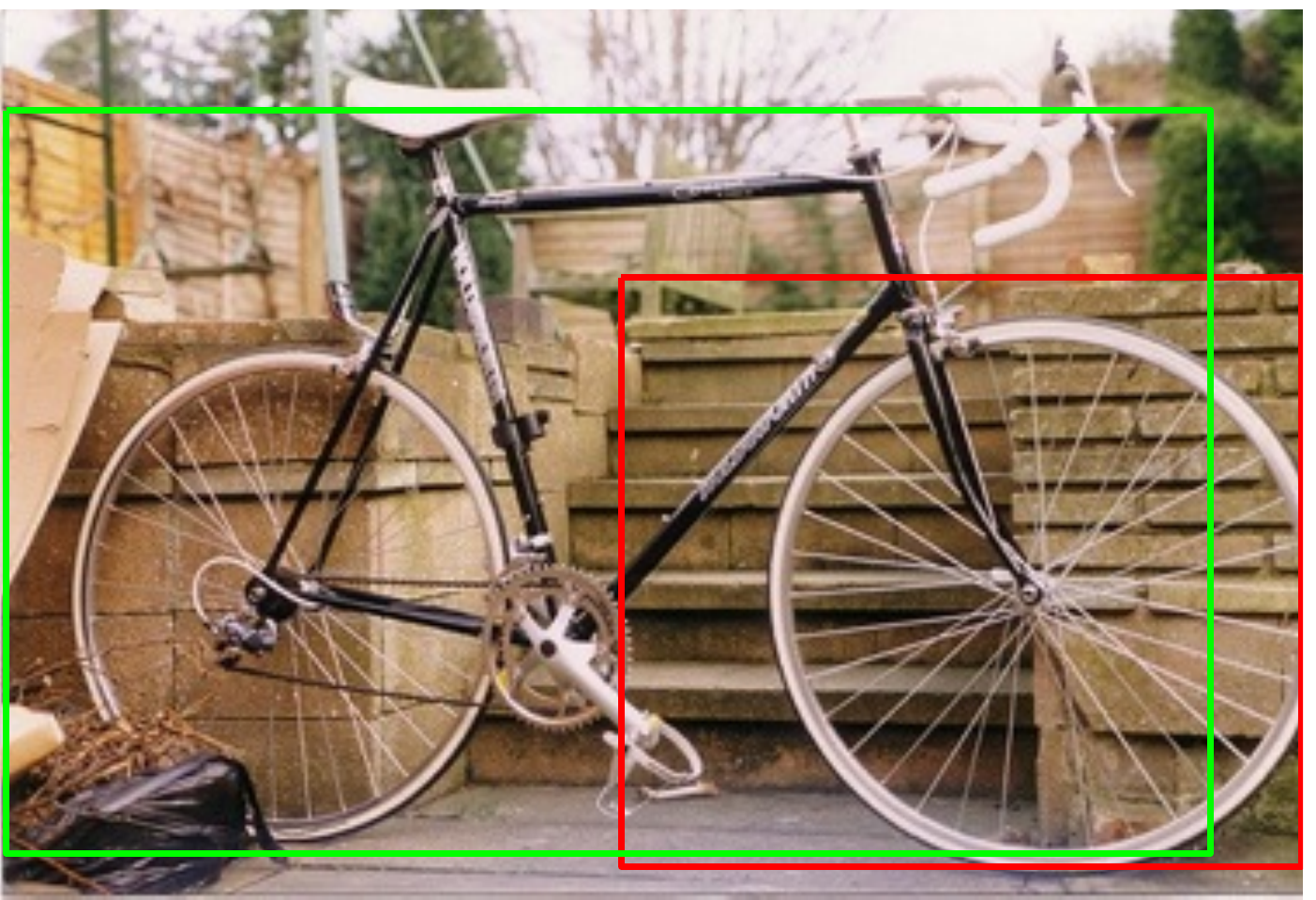} \hspace{0.04cm}
\includegraphics[width=0.19\textwidth, height=2.2cm]{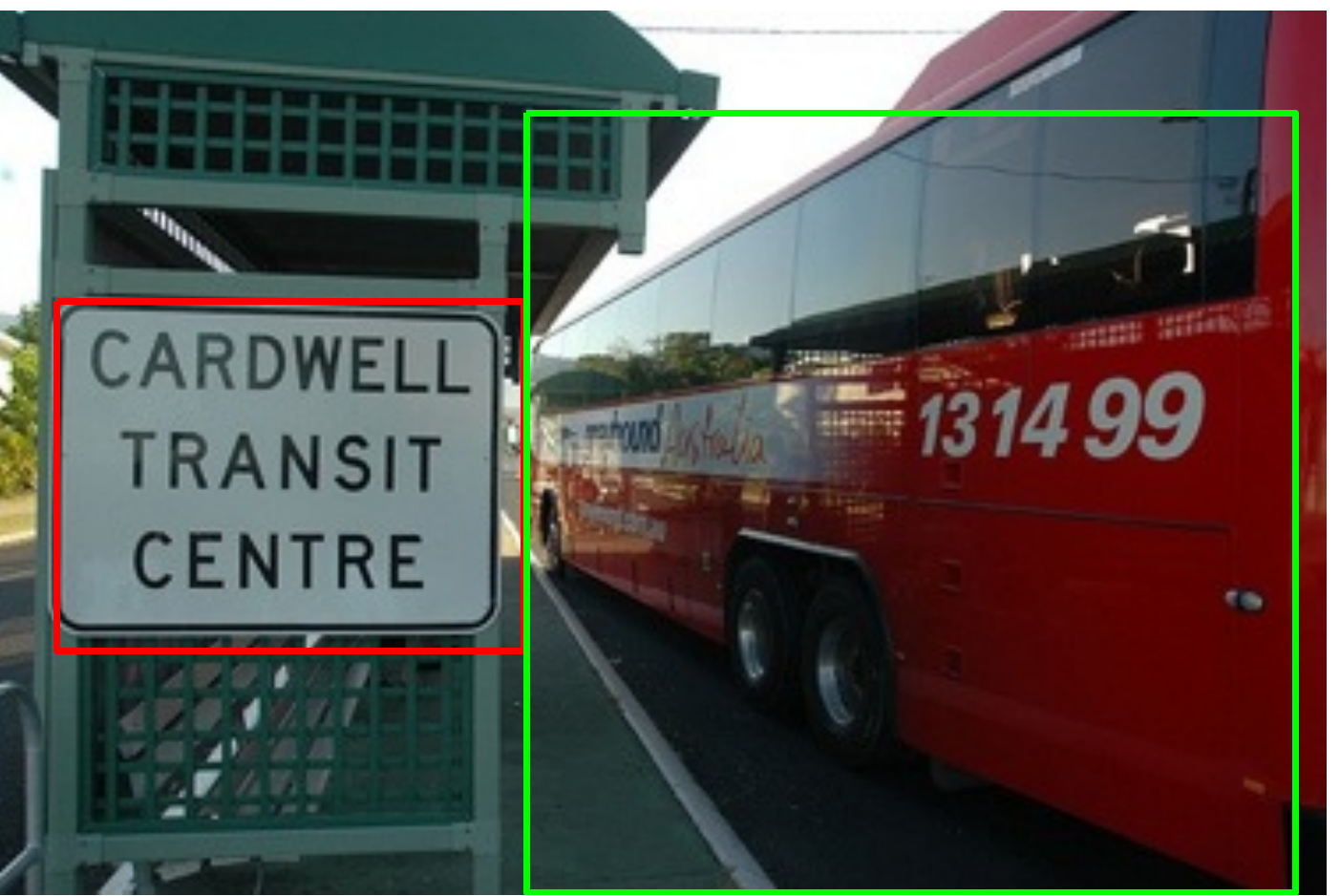}  \hspace{0.04cm} 
\includegraphics[width=0.19\textwidth, height=2.2cm]{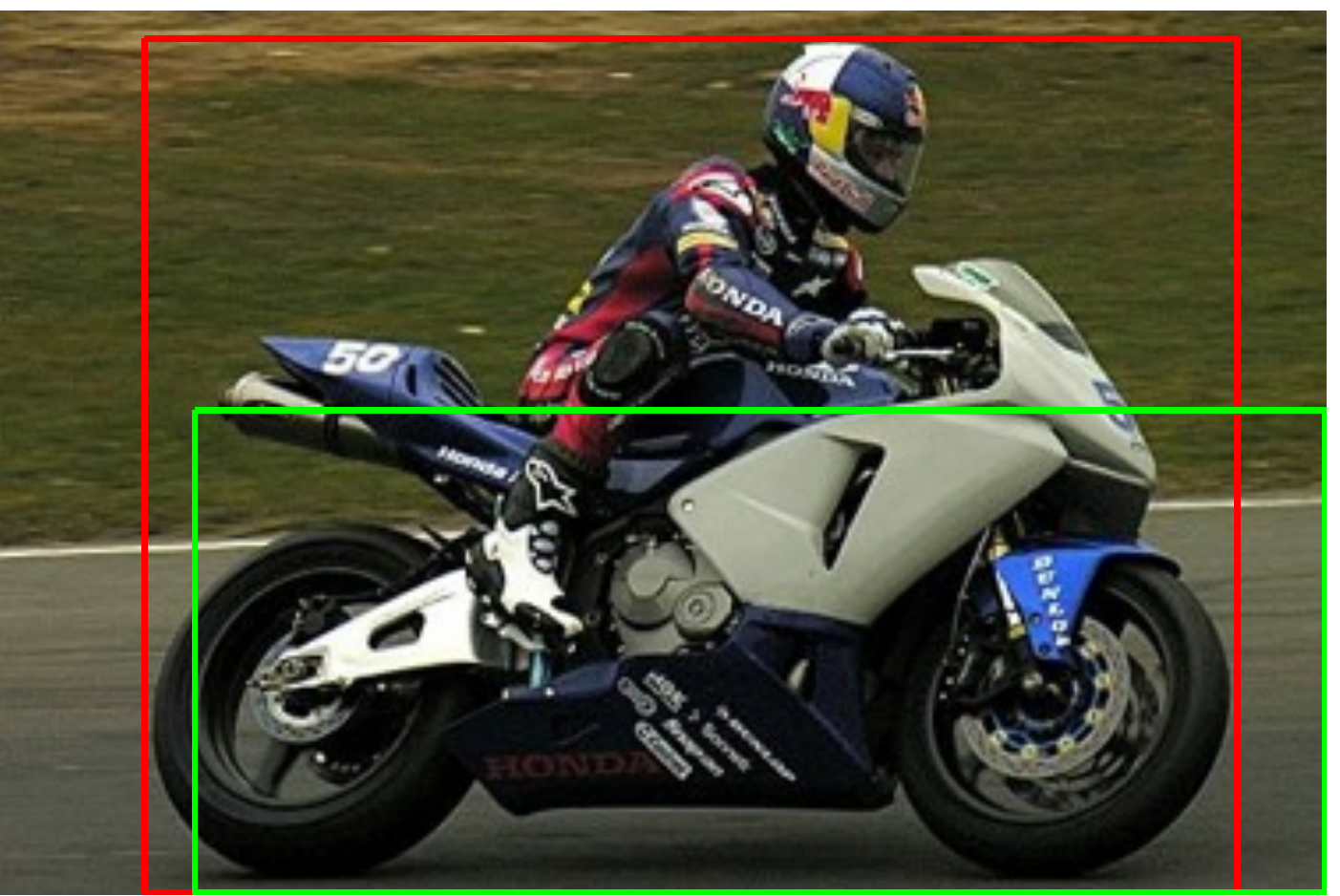}  \hspace{0.04cm}
\includegraphics[width=0.19\textwidth, height=2.2cm]{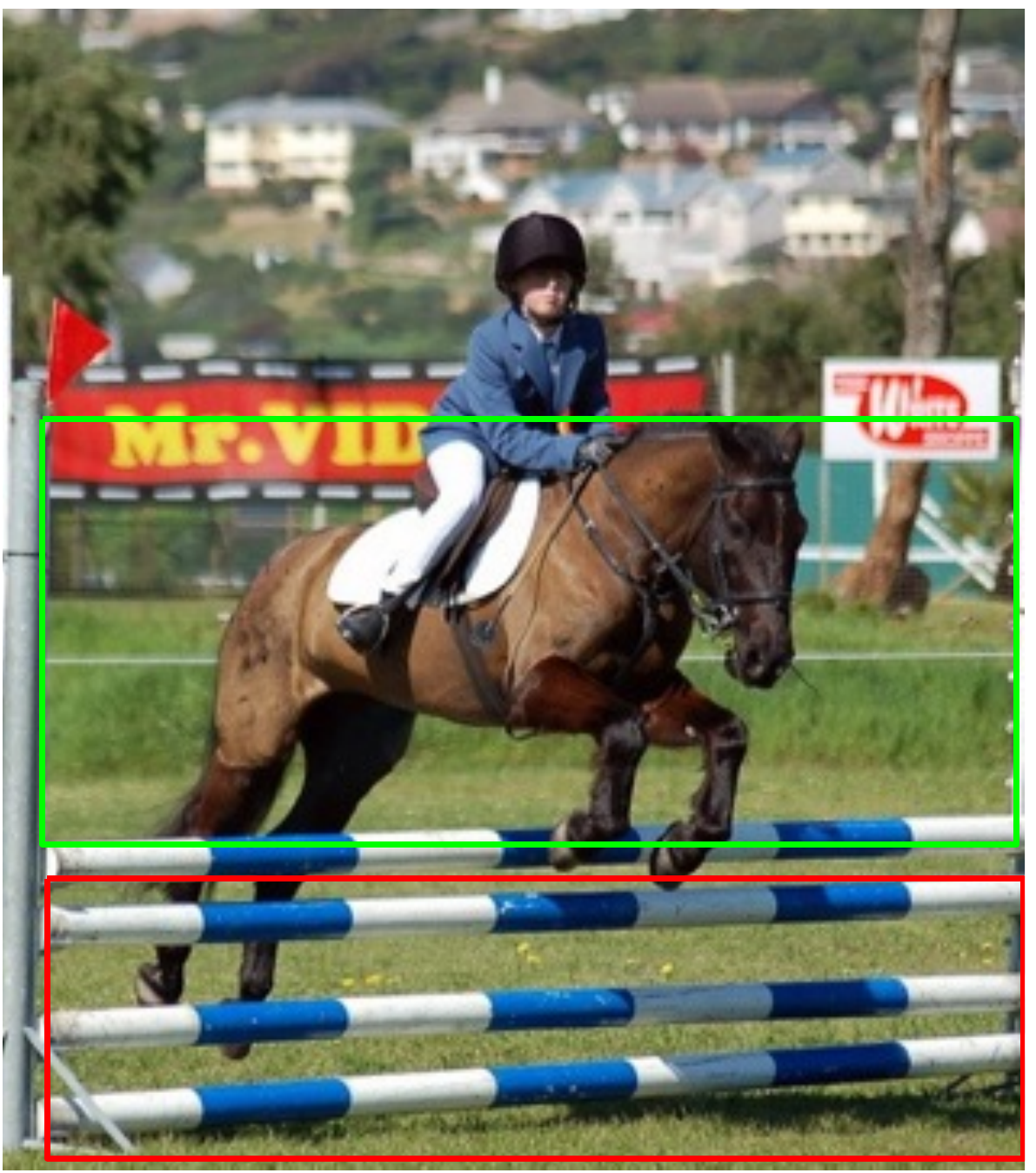}  \hspace{0.04cm}
\includegraphics[width=0.19\textwidth, height=2.2cm]{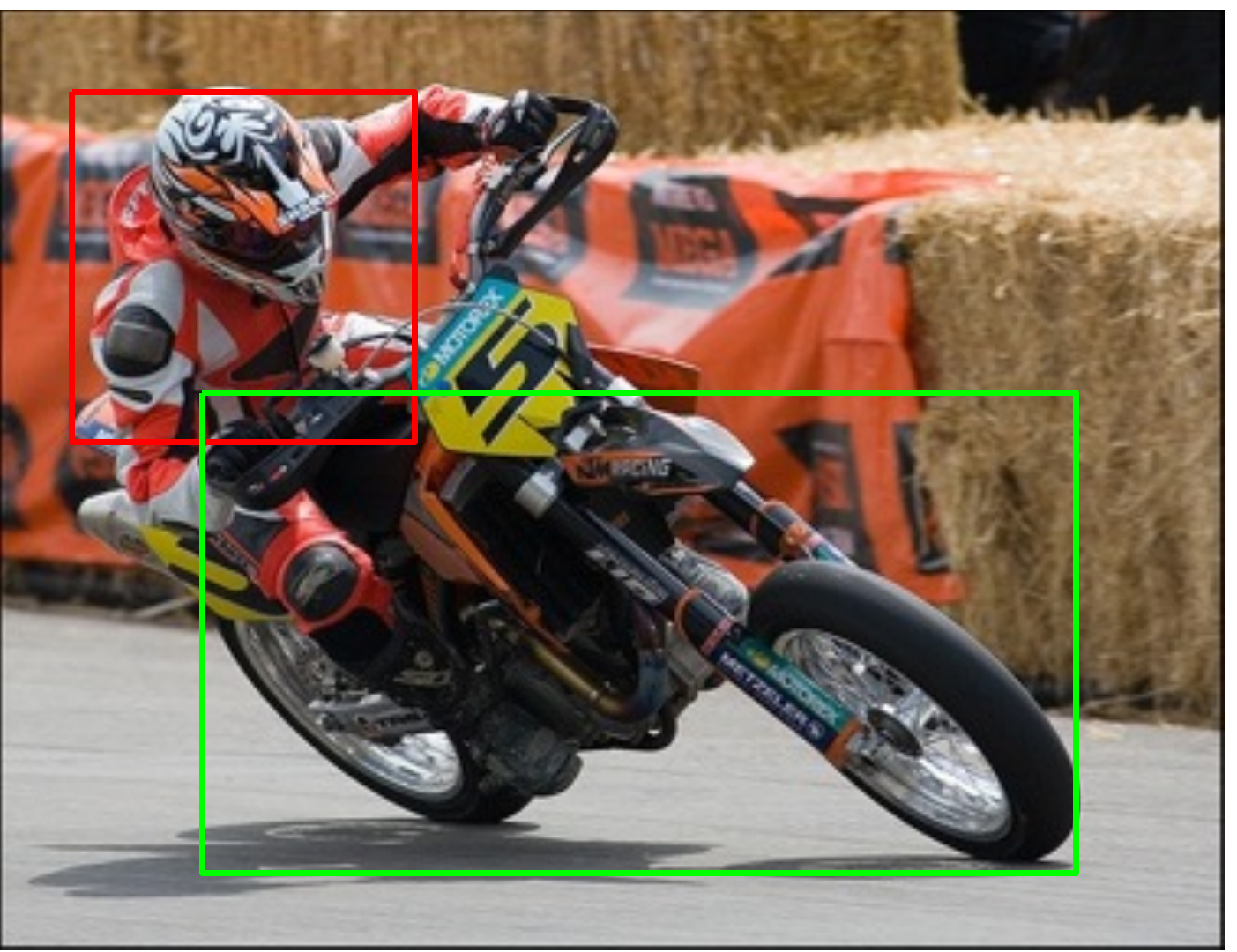} 

\includegraphics[width=0.19\textwidth, height=2.2cm]{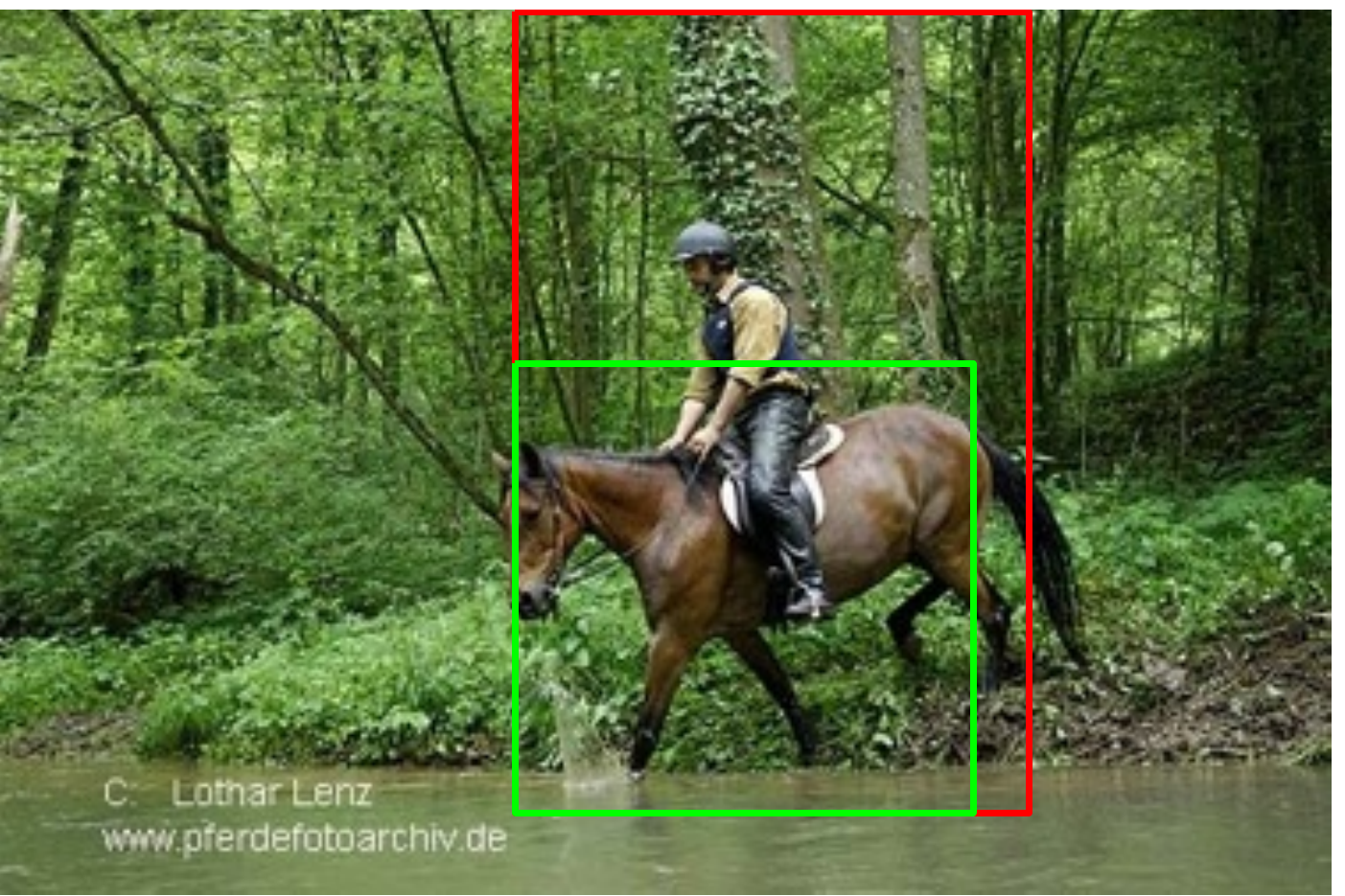} \hspace{0.04cm}
\includegraphics[width=0.19\textwidth, height=2.2cm]{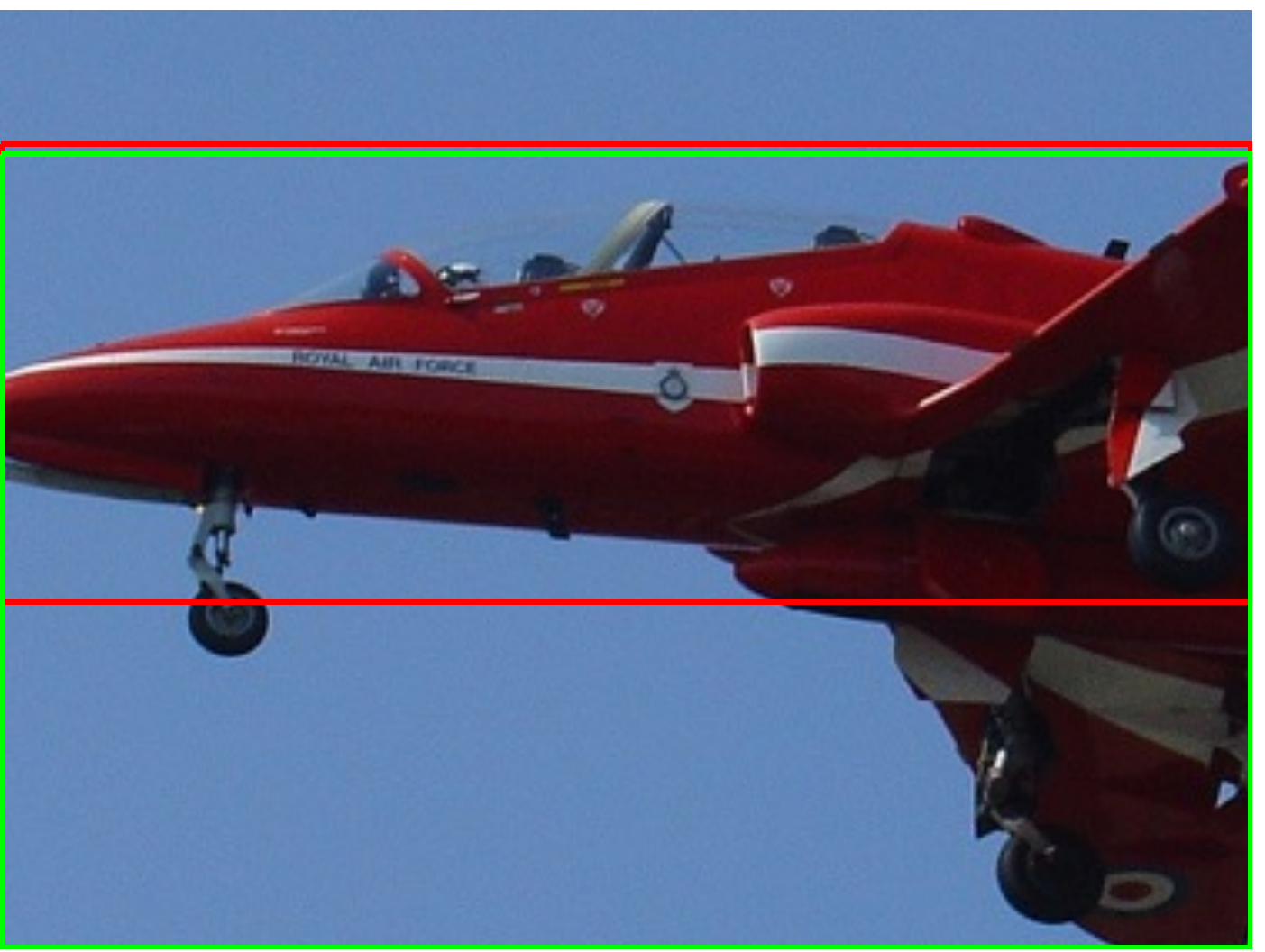} \hspace{0.04cm}
\includegraphics[width=0.19\textwidth, height=2.2cm]{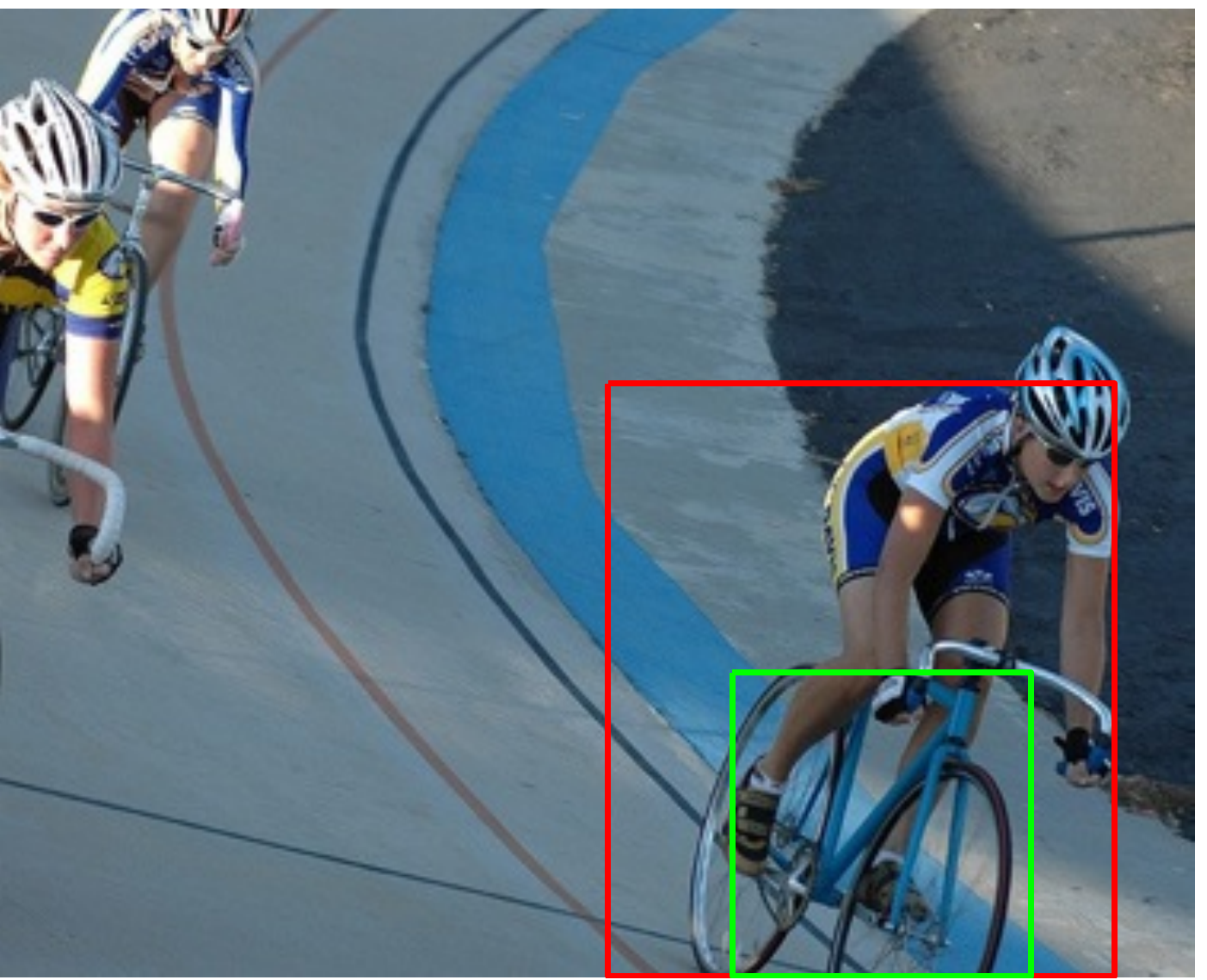} \hspace{0.04cm}
\includegraphics[width=0.19\textwidth, height=2.2cm]{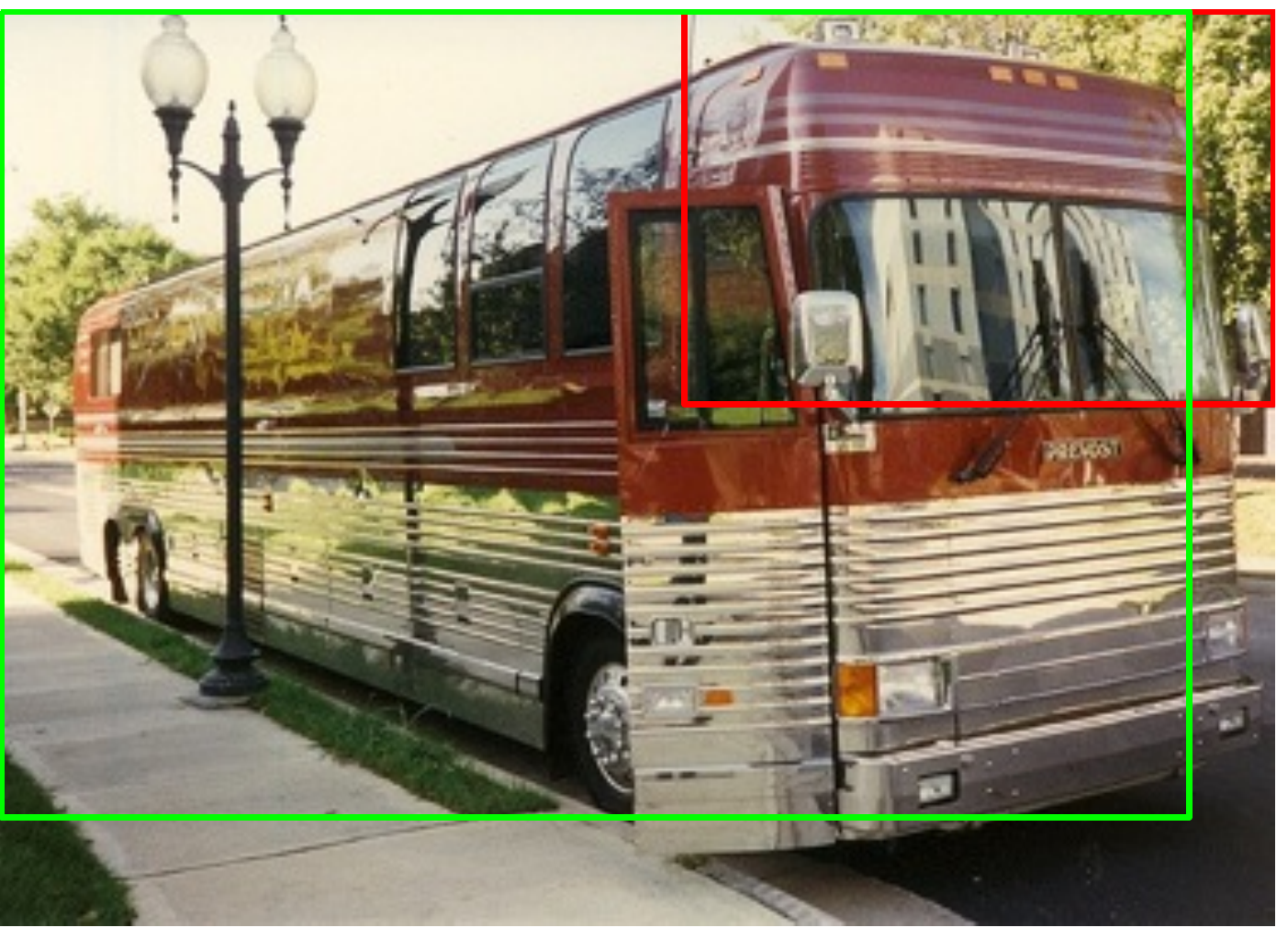} \hspace{0.04cm}
\includegraphics[width=0.19\textwidth, height=2.2cm]{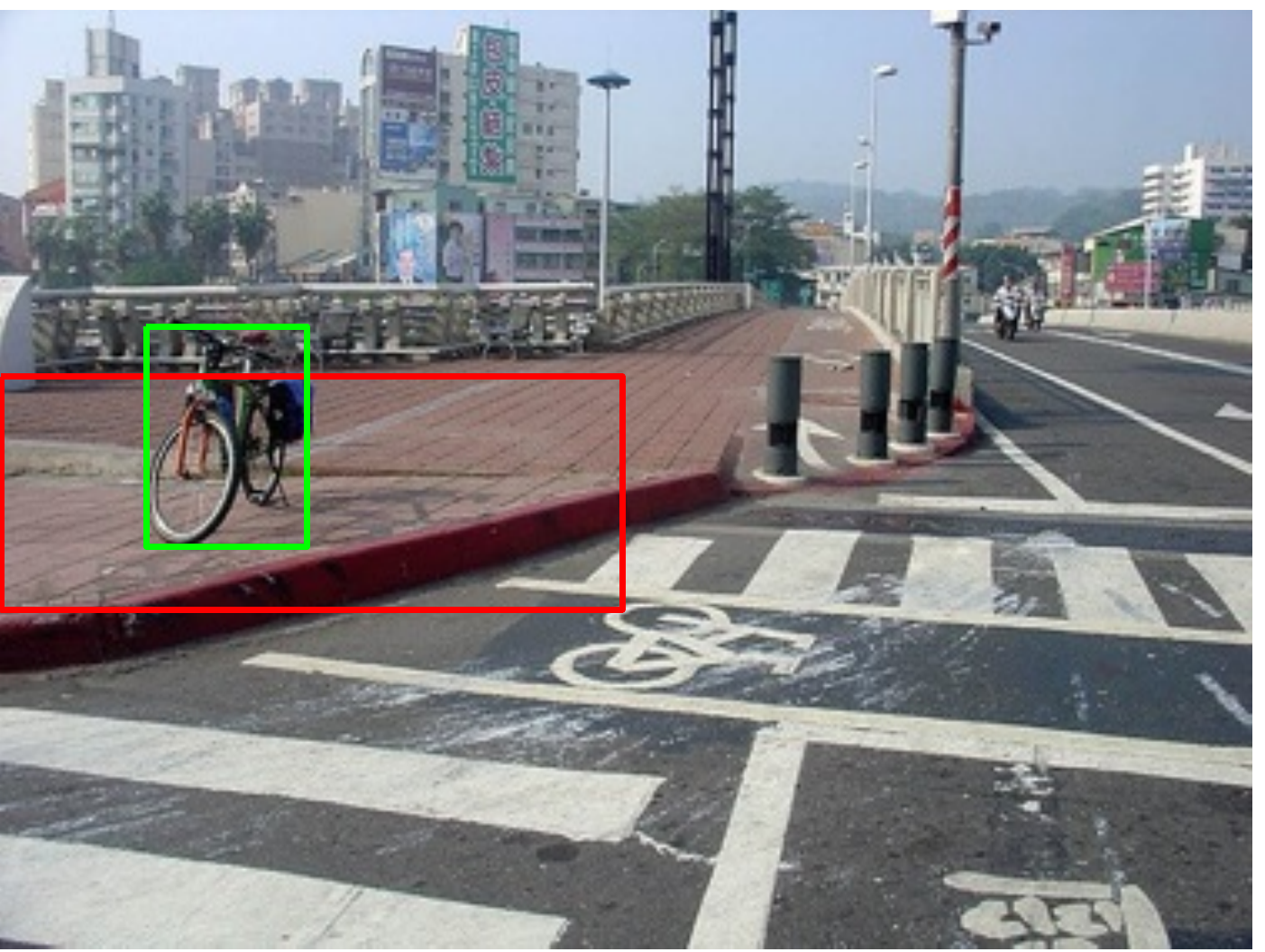}
\caption{Visualization of some common failure cases of constructed positive windows by\cite{siva2012defence} vs our method. Red bounding boxes are constructed positive windows from \cite{siva2012defence}. Green bounding boxes are constructed positive windows from our method.} 
\label{fig:us_vs_siva} 
\end{figure*}

\begin{table*}[htbp]
\footnotesize
\centering
\renewcommand{\arraystretch}{1.1}
\renewcommand{\tabcolsep}{1.5mm}
\begin{tabular}{*{5}{l}}
\toprule
Dataset& LSVM w/o bias & SLSVM w/o bias &LSVM w/ bias & SLSVM w/ bias\\
\midrule
musk1    &  70.8 $\pm$ 14.4 &  80.3 $\pm$ 10.3  &  81.7 $\pm$ 14.5  &  79.2 $\pm$ 13.4 \\
musk2    &  51.0 $\pm$ 10.9 &  79.5 $\pm$ 10.4  &  80.5 $\pm$ 9.9    &  84.3 $\pm$ 11.4 \\
\midrule
fox           &  51.5 $\pm$ 7.5   &  63.0 $\pm$ 11.8  &  57.0 $\pm$ 8.9   &  61.0 $\pm$ 12.6 \\
elephant & 81.5 $\pm$ 6.3   &  88.0 $\pm$ 6.7    &  81.5 $\pm$ 4.1    &  87.0 $\pm$ 6.3 \\
tiger        &  79.5 $\pm$ 8.6   &  85.5 $\pm$ 6.4    &  86.0 $\pm$ 9.1    &  87.5 $\pm$ 7.9 \\
\midrule
trec1       &  94.3 $\pm$ 2.9    &  95.5 $\pm$ 2.6   &  95.3 $\pm$ 3.0   &  95.3 $\pm$ 2.8 \\
trec2       &  69.0 $\pm$ 6.8    &  83.0 $\pm$ 6.5   &  86.5 $\pm$ 5.7   &  83.8 $\pm$ 7.4 \\
trec3       &  77.5 $\pm$ 5.8    &  90.0 $\pm$ 5.8   &  85.5 $\pm$ 6.3   &  86.0 $\pm$ 6.5 \\
trec4       &  77.3 $\pm$ 8.0    &  85.0 $\pm$ 5.1   &  85.3 $\pm$ 3.6   &  86.3 $\pm$ 5.2  \\
trec7       &  74.5 $\pm$ 9.8    &  83.8 $\pm$ 4.0   &  82.5 $\pm$ 7.0   & 81.5 $\pm$ 5.8  \\
trec9       &  66.8 $\pm$ 5.0    &  70.3 $\pm$ 5.7   &  68.8 $\pm$ 8.0   & 71.5 $\pm$ 6.4 \\
trec10     &  71.0 $\pm$ 9.9    &  84.3 $\pm$ 5.4   &  80.8 $\pm$ 6.6   & 82.8 $\pm$ 7.3  \\
\bottomrule
\end{tabular}
\caption{10 fold average and standard deviation of the test accuracy on MIL dataset. The two methods start from the same initialization introduced in \cite{misvm-nips}}
\label{tab:mil-experiments}
\end{table*}

\begin{table*}[htbp]
\footnotesize
\centering
\renewcommand{\arraystretch}{1.0}
\renewcommand{\tabcolsep}{1.5mm}
\begin{tabular}{l*{14}{r}}
\toprule
\multirow{2}{*}{Method} & \multicolumn{2}{c}{aeroplane} & \multicolumn{2}{c}{bicycle} & \multicolumn{2}{c}{boat} & \multicolumn{2}{c}{bus}& \multicolumn{2}{c}{horse} & \multicolumn{2}{c}{motorbike} & \multirow{2}{*}{~mAP}\\
  & left & right & left & right & left & right & left & right & left & right & left & right\\
\midrule
\cite{deselaers1}  & 9.1 & 23.6 & 33.4 & 49.4 & 0.0 & 0.0 & 0.0 & 16.4 & 9.6 & 9.1 & 20.9 & 16.1 & ~16.0\\
\midrule
\cite{pandey}  & 7.5 & 21.1 & 38.5 & 44.8 & 0.3 & 0.5 & 0.0 & 0.3 & 45.9 & 17.3 & 43.8 & 27.2 & ~20.8\\
\midrule
\cite{deselaers2}  & 5.3 & 18.1 & 48.6 & 61.6 & 0.0 & 0.0 & 0.0 & 16.4 & 29.1 & 14.1 & 47.7 & 16.2 & ~21.4\\
\midrule
\cite{russakovsky} & \multicolumn{2}{c}{30.8} & \multicolumn{2}{c}{25.0}  & \multicolumn{2}{c}{3.6} & \multicolumn{2}{c}{26.0}  & \multicolumn{2}{c}{21.3}  & \multicolumn{2}{c}{29.9} &~ 22.8\\
\midrule
\cite{siva2012defence} with our features &  \multicolumn{2}{c}{23.2} &  \multicolumn{2}{c}{15.4} &  \multicolumn{2}{c}{5.1} &  \multicolumn{2}{c}{2.0} &  \multicolumn{2}{c}{6.2} &  \multicolumn{2}{c}{17.4} & ~11.6\\
\midrule
Cover + SVM & \multicolumn{2}{c}{23.4} & \multicolumn{2}{c}{43.5}  & \multicolumn{2}{c}{8.1} & \multicolumn{2}{c}{33.9}  & \multicolumn{2}{c}{24.7}  & \multicolumn{2}{c}{40.2} &~ 29.0\\
\midrule
Cover + LSVM & \multicolumn{2}{c}{28.2} & \multicolumn{2}{c}{47.2}  & \multicolumn{2}{c}{9.6} & \multicolumn{2}{c}{34.7}  & \multicolumn{2}{c}{25.2}  & \multicolumn{2}{c}{39.8} &~ 30.8\\
\bottomrule
\end{tabular}
\caption{Detection average precision (\%) on PASCAL VOC 2007-6x2 test set. First three baseline methods report results limited to left and right subcategories of the objects.}
\label{tab:detection-6x2}
\end{table*}

\begin{table*}[htbp]
\footnotesize
\centering
\renewcommand{\arraystretch}{1.0}
\renewcommand{\tabcolsep}{0.35mm}
\begin{tabular}{l *{21}{c}}
\toprule
VOC2007 test & ~aero & bike & bird & boat & bottle & bus & car & cat & chair & cow & table & dog & horse & mbike & pson & plant & sheep & sofa & train & tv & ~mAP\\
\midrule
\cite{siva1} & ~13.4 & 44.0 & 3.1 & 3.1 & 0.0 & 31.2 & 43.9 & 7.1 & 0.1 & 9.3 & 9.9 & 1.5 & 29.4 & 38.3 & 4.6 & 0.1 & 0.4 & 3.8 & 34.2 & 0.0 & ~ 13.9\\
\midrule
Cover + SVM & ~23.4 & 43.5 & 22.4 & 8.1 & 6.2 & 33.9 & 33.8 & 30.4 & 0.1 &17.9 & 11.5 & 17.1 & 24.7 & 40.2 & 2.4 & 14.8 & 21.4 & 15.1 & 31.9 & 6.2 & ~20.3\\
\midrule
Cover + LSVM & ~28.2 & 47.2 & 17.6 & 9.6 & 6.5 & 34.7 & 35.5 & 31.5 & 0.3 & 21.7 & 13.2 & 20.7 & 25.2 & 39.8 & 12.6 & 18.6 & 21.2 & 18.6 & 31.7 & 10.2 & ~22.2\\
\midrule
Cover + SLSVM & ~27.6 & 41.9 & 19.7 & 9.1 & 10.4 & 35.8 & 39.1 & 33.6 & 0.6 & 20.9 & 10.0 & 27.7 & 29.4 & 39.2 & 9.1 & 19.3 & 20.5 & 17.1 & 35.6 & 7.1 & ~22.7\\
\bottomrule
\end{tabular}
\caption{Detection average precision (\%) on full PASCAL VOC 2007 test set.}
\label{tab:detection-full}
\end{table*}

\section{Experiments}
\label{sec:exp}

We performed two sets of experiments, one on a multiple instance learning dataset \cite{misvm-nips} and the other on the PASCAL VOC 2007 data \cite{PASCAL07}.  The first experiment was designed to compare the multiple instance learning bag classification performance of LSVM with Smooth LSVM (SLSVM). The second experiment evaluates detection accuracy (measured in average precision) of our framework  in comparison to baselines. 

\subsection{Multiple instance learning datasets}

We evaluated our method in Section 5 on standard multiple instance learning datasets \cite{misvm-nips}. For preprocessing, we centered each feature dimension and $\ell_2$ normalize the data. For fair comparison with \cite{misvm-nips}, we use the same initialization, where the initial weight vector is obtained by training an SVM with all the negative instances and bag-averaged positive instances. For this experiment, we performed 10 fold cross validation on $C$  and $\mu$.  Table \ref{tab:mil-experiments} shows the experimental results. Without the bias, our method significantly performs better than LSVM method and with the bias, our method shows modest improvement in most cases.

\subsection{Weakly-supervised object detection}

To implement our weakly-supervised detection system we need suitable image features for computing the nearest neighbors of each image window in Section \ref{sec:covering} and for learning object detectors.
We use the recently proposed R-CNN \cite{girshick2014rcnn} detection framework to compute features on image windows in both cases. Specifically, we use the convolutional neural network (CNN) distributed with DeCAF \cite{decafICML}, which is trained on the ImageNet ILSVRC 2012 dataset (using only image-level annotations).
We avoid using the better performing CNN that is fine-tuned on PASCAL data, as described in \cite{girshick2014rcnn}, because fine-tuning requires instance-level annotations. 

We report detection accuracy as average precision on the standard benchmark dataset for object detection, PASCAL VOC 2007 \emph{test} \cite{PASCAL07}. We compare to five different baseline methods that learn object detectors with limited annotations. Note that other baseline methods use additional information besides the one-bit image-level annotations. \citet{deselaers1, deselaers2} use a set of $799$ images with bounding box annotations as meta-training data. In addition to bounding box annotations, \citet{deselaers1, deselaers2, pandey} use extra instance level annotations such as \emph{pose}, \emph{difficult} and \emph{truncated}. \citet{siva2012defence,russakovsky} use \emph{difficult} instance annotations but not \emph{pose} or \emph{truncated}. First, we report the detection average precision on $6$ subsets of classes in table \ref{tab:detection-6x2} to compare with \citet{deselaers1, deselaers2, pandey}. 
 
To evaluate the efficacy of our initialization, we compare it to the state-of-the-art algorithm recently proposed by \cite{siva2012defence}. Their method constructs a set of positive windows by looping over each positive image and picking the instance that has the maximum distance to its nearest neighbor over all negative instances (and thus the name negative data \emph{mining} algorithm). For a fair comparison, we used the same window proposals, the same features \cite{girshick2014rcnn}, the same L2 distance metric, and the same PASCAL 2007 detection evaluation criteria. The class mean average precision for the mining algorithm was $11.6\%$ compared to $29.0\%$ obtained by our initialization procedure. Figure \ref{fig:us_vs_siva} visualizes some command failure modes in our implementation of  \cite{siva2012defence}.  Since the negative mining method does not take into account the similarity among positive windows (in contrast to our method) our intuition is that the method is less robust to intra-class variations and background clutter. Therefore, it often latches onto background objects (i.e. hurdle in horse images, street signs in bus images), onto parts of the full objects (i.e. wheels of bicycles), or merges two different objects (i.e. rider and motorcycle). It is worth noting that \citet{pandey, siva2012defence} use the CorLoc metric\footnote{CorLoc was proposed by \cite{deselaers1} to evaluate the detection results on PASCAL \emph{train} set} as the evaluation metric to report results on PASCAL \emph{test} set. In contrast, in our experiments, we exactly follow the PASCAL VOC evaluation protocol (and use the PASCAL VOC devkit scoring software) and report detection average precision. 
 
Table \ref{tab:detection-full} shows the detection result on the full PASCAL 2007 dataset. There are two baseline methods \cite{siva1, russakovsky} which report the result on the full dataset. Unfortunately, we were not able to obtain the per-class average precision data from the authors of \cite{russakovsky} except the class mean average precision (mAP) of 15.0\%. As shown in Table \ref{tab:detection-full}, the initial detector model trained from the constructed set of positive windows already produces good object detectors but we can provide further improvement by optimizing the MIL objective.

\vspace{-2pt}
\section{Conclusion}

We developed a framework for learning to localize objects with one-bit object presence labels. Our results show that the proposed framework can construct a set of positive windows to train initial detection models and improve the models with the refinement optimization method. We achieve state-of-the-art performance for object detection with minimal supervision on the standard benchmark object detection dataset. Source code will be available on the author's website.

\vspace{-5pt}
\section*{Acknowledgement}
\small{
We thank Yong Jae Lee for helpful insights and discussions. H. Song was supported by Samsung Scholarship Foundation. J. Mairal and Z. Harchaoui were funded by the INRIA-UC Berkeley associated team ``Hyperion", a grant from the France-Berkeley fund, the Gargantua project under program Mastodons of CNRS, and the LabEx PERSYVAL-Lab (ANR-11-LABX-0025). This work was partially supported by  ONR N00014-11-1-0688, NSF, DARPA, and Toyota.}

\bibliography{refs}
\bibliographystyle{icml2014}

\end{document}